\documentclass[acmtog]{acmart}
\usepackage{subfig}
\usepackage{colortbl}
\usepackage{xcolor}
\usepackage{booktabs}
\usepackage[ruled]{algorithm2e} % For algorithms

\SetAlFnt{\small}
\SetAlCapFnt{\small}
\SetAlCapNameFnt{\small}
\SetAlCapHSkip{0pt}
\acmJournal{TOG}
\usepackage{graphicx}
\usepackage{amsfonts}
\usepackage{amsmath}
\usepackage{multirow}
\usepackage{color}
\usepackage{tabularx}
\usepackage{soul}
 
\definecolor{bronze}{rgb}{1,1,0.6}
\definecolor{silver}{rgb}{0.969,0.796,0.600}
\definecolor{gold}{rgb}{0.941,0.592,0.600}
\definecolor{purple}{rgb}{0.913,0.753,0.941}

\usepackage{subfig}
\usepackage{colortbl}
\usepackage{xcolor}
\usepackage{booktabs}
\usepackage{stackengine}
\copyrightyear{2025}
\acmYear{2025}
\setcopyright{cc}
\setcctype{by-nc}
\acmConference[SIGGRAPH Conference Papers '25]{Special Interest Group on Computer Graphics and Interactive Techniques Conference Conference Papers }{August 10--14, 2025}{Vancouver, BC, Canada}
\acmBooktitle{Special Interest Group on Computer Graphics and Interactive Techniques Conference Conference Papers (SIGGRAPH Conference Papers '25), August 10--14, 2025, Vancouver, BC, Canada}
\acmDOI{10.1145/3721238.3730677}
\acmISBN{979-8-4007-1540-2/2025/08}
\begin{document}
%%%%%%%%%%%%%%%%%%%%%%%%%%%%%%%%%%%%%%%%%%%%%%%%%%%%%%%%%%%%%%%%%%%%%%%%%%%%%%%%%%%%%%%%%%%%%%
\title{\textbf{EVA}: \textbf{E}xpressive \textbf{V}irtual \textbf{A}vatars from Multi-view Videos}
%%%%%%%%%%%%%%%%%%%%%%%%%%%%%%%%%%%%%%%%%%%%%%%%%%%%%%%%%%%%%%%%%%%%%%%%%%%%%%%%%%%%%%%%%%%%%%
\author{Hendrik Junkawitsch}
\affiliation{
	\institution{Max Planck Institute for Informatics, Saarland Informatics Campus}
	\streetaddress{Campus E14}
	\city{Saarbr\"ucken}
	\state{Saarland}
    \country{Germany}
	\postcode{66123}
}
\author{Guoxing Sun}
\affiliation{
	\institution{Max Planck Institute for Informatics, Saarland Informatics Campus}
	\streetaddress{Campus E14}
	\city{Saarbr\"ucken}
	\state{Saarland}
    \country{Germany}
	\postcode{66123}
}
\author{Heming Zhu}
\affiliation{
	\institution{Max Planck Institute for Informatics, Saarland Informatics Campus}
	\streetaddress{Campus E14}
	\city{Saarbr\"ucken}
	\state{Saarland}
    \country{Germany}
	\postcode{66123}
}
\author{Christian Theobalt}
\affiliation{
	\institution{Max Planck Institute for Informatics, Saarland Informatics Campus and Saarbrücken Research Center for Visual Computing, Interaction and AI}
	\streetaddress{Campus E14}
	\city{Saarbr\"ucken}
	\state{Saarland}
    \country{Germany}
	\postcode{66123}
}
\author{Marc Habermann}
\affiliation{
	\institution{Max Planck Institute for Informatics, Saarland Informatics Campus and Saarbrücken Research Center for Visual Computing, Interaction and AI}
	\streetaddress{Campus E14}
	\city{Saarbr\"ucken}
	\state{Saarland}
    \country{Germany}
	\postcode{66123}
}
%%%%%%%%%%%%%%%%%%%%%%%%%%%%%%%%%%%%%%%%%%%%%%%%%%%%%%%%%%%%%%%%%%%%%%%%%%%%%%%%%%%%%%%%%%%%%%
\renewcommand{\shortauthors}{Junkawitsch et al.}
\begin{CCSXML}
<ccs2012>
   <concept>
       <concept_id>10010147.10010371.10010372</concept_id>
       <concept_desc>Computing methodologies~Rendering</concept_desc>
       <concept_significance>500</concept_significance>
       </concept>
   <concept>
       <concept_id>10010147.10010178.10010224</concept_id>
       <concept_desc>Computing methodologies~Computer vision</concept_desc>
       <concept_significance>500</concept_significance>
       </concept>
   <concept>
       <concept_id>10010520.10010570</concept_id>
       <concept_desc>Computer systems organization~Real-time systems</concept_desc>
       <concept_significance>500</concept_significance>
       </concept>
 </ccs2012>
\end{CCSXML}
\ccsdesc[500]{Computing methodologies~Rendering}
\ccsdesc[500]{Computing methodologies~Computer vision}
\ccsdesc[500]{Computer systems organization~Real-time systems}
%%%%%%%%%%%%%%%%%%%%%%%%%%%%%%%%%%%%%%%%%%%%%%%%%%%%%%%%%%%%%%%%%%%%%%%%%%%%%%%%%%%%%%%%%%%%%%
\keywords{Expressive 3D human reconstruction and rendering, human animation, multi-view reconstruction, real-time}
%%%%%%%%%%%%%%%%%%%%%%%%%%%%%%%%%%%%%%%%%%%%%%%%%%%%%%%%%%%%%%%%%%%%%%%%%%%%%%%%%%%%%%%%%%%%%%
\begin{teaserfigure}
\centering
\includegraphics[trim={0cm 0cm 0cm 0.3cm},clip,width=1\textwidth]{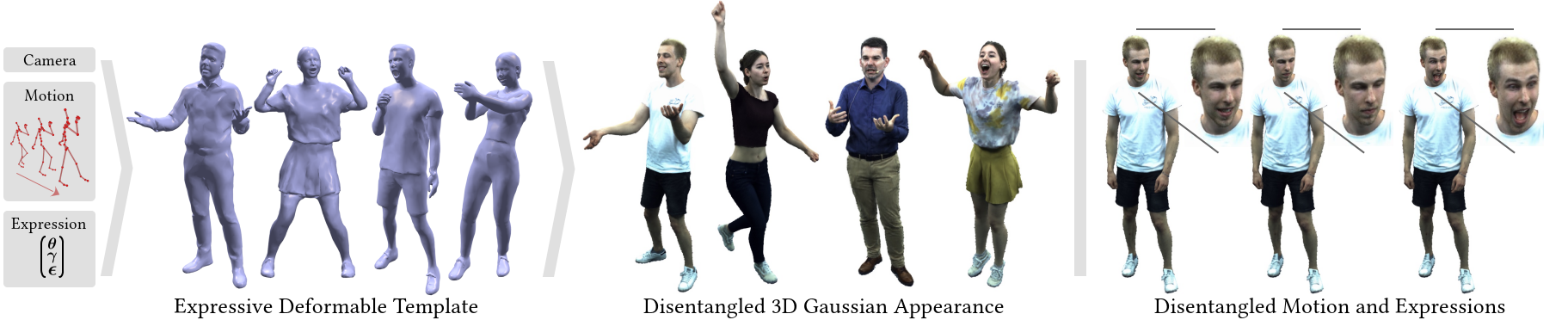}
\caption{\textbf{EVA} generates high-fidelity, real-time human renderings from arbitrary camera views, skeletal motion, and expression parameters. It leverages an expressive template geometry as a geometric proxy for its 3D Gaussian appearance model, enabling disentangled control over body, hands, and face. }
\label{fig:teaser}
\end{teaserfigure}
%%%%%%%%%%%%%%%%%%%%%%%%%%%%%%%%%%%%%%%%%%%%%%%%%%%%%%%%%%%%%%%%%%%%%%%%%%%%%%%%%%%%%%%%%%%%%%
\begin{abstract}
With recent advancements in neural rendering and motion capture algorithms, remarkable progress has been made in photorealistic human avatar modeling, unlocking immense potential for applications in virtual reality, augmented reality, remote communication, and industries such as gaming, film, and medicine.
However, existing methods fail to provide complete, faithful, and expressive control over human avatars due to their entangled representation of facial expressions and body movements.
In this work, we introduce \textbf{Expressive Virtual Avatars (EVA)}, an actor-specific, fully controllable, and expressive human avatar framework that achieves high-fidelity, lifelike renderings in real time while enabling independent control of facial expressions, body movements, and hand gestures.
Specifically, our approach designs the human avatar as a two-layer model: an \textit{expressive template geometry layer} and a \textit{3D Gaussian appearance layer}.
First, we present an expressive template tracking algorithm that leverages coarse-to-fine optimization to accurately recover body motions, facial expressions, and non-rigid deformation parameters from multi-view videos.
Next, we propose a novel decoupled 3D Gaussian appearance model designed to effectively disentangle body and facial appearance.
Unlike unified Gaussian estimation approaches, our method employs two specialized and independent modules to model the body and face separately.
Experimental results demonstrate that EVA surpasses state-of-the-art methods in terms of rendering quality and expressiveness, validating its effectiveness in creating full-body avatars.
This work represents a significant advancement towards fully drivable digital human models, enabling the creation of lifelike digital avatars that faithfully replicate human geometry and appearance.
For more details, we refer to our project page: \url{https://vcai.mpi-inf.mpg.de/projects/EVA/}.
\end{abstract}
%%%%%%%%%%%%%%%%%%%%%%%%%%%%%%%%%%%%%%%%%%%%%%%%%%%%%%%%%%%%%%%%%%%%%%%%%%%%%%%%%%%%%%%%%%%%%%
\maketitle
%%%%%%%%%%%%%%%%%%%%%%%%%%%%%%%%%%%%%%%%%%%%%%%%%%%%%%%%%%%%%%%%%%%%%%%%%%%%%%%%%%%%%%%%%%%%%%
%%%%%%%%%%%%%%%%%%%%%%%%%%%%%%%%%%%%%%%%%%%%%%%%%%%%%%%%%%%%%%%%%%%%%%%%%%%%%%%%%%%%%%%%%%%%%%
%
\section{Introduction} \label{sec:introduction}
%
%%%%%%%%%%%%%%%%%%%%%%%%%%%%%%%%%%%%%%%%%%%%%%%%%%%%%%%%%%%%%%%%%%%%%%%%%%%%%%%%%%%%%%%%%%%%%%
% Intro
%%%%%%%%%%%%%%%%%%%%%%%%%%%%%%%%%%%%%%%%%%%%%%%%%%%%%%%%%%%%%%%%%%%%%%%%%%%%%%%%%%%%%%%%%%%%%%
%
In the rapidly evolving digital landscape, the creation of realistic virtual human avatars has become increasingly important for applications in virtual reality, augmented reality, gaming, film, and remote communication. 
These avatars bridge the gap between real-world interactions and digital spaces, enabling lifelike telepresence and immersive virtual experiences.
However, existing approaches either provide only body-level control or fail to accurately capture body dynamics. 
This motivates us to develop \emph{expressive} avatars that (1) offer full and disentangled control over the body, hands, and face, (2) deliver photorealistic renderings of dynamic motions and expressions, and (3) achieve real-time inference.
%
%%%%%%%%%%%%%%%%%%%%%%%%%%%%%%%%%%%%%%%%%%%%%%%%%%%%%%%%%%%%%%%%%%%%%%%%%%%%%%%%%%%%%%%%%%%%%%
% Related work
%%%%%%%%%%%%%%%%%%%%%%%%%%%%%%%%%%%%%%%%%%%%%%%%%%%%%%%%%%%%%%%%%%%%%%%%%%%%%%%%%%%%%%%%%%%%%%
%
\par
Researchers have explored various human representations to create animatable avatars from multi-view videos \citep{starck2005video} or 3D scans \citep{li2022avatarcap}.
Early methods \citep{alldieck20183DV, xu2011video} used template meshes with texture maps for novel-view rendering but lacked photorealism.
Recent works integrate neural rendering \citep{mildenhall2021nerf, kerbl3Dgaussians} with human templates \citep{SMPL:2015, SMPL-X:2019, habermann2021DDC} for realistic renderings with control.
For example, \citet{peng2021animatable} use canonical neural radiance fields with neural blend weight fields for skeleton-driven deformations but struggle with cloth wrinkles, and their model inference is slow.
Other methods \citep{Pang_2024_CVPR, li2024animatablegaussians} predict 3D Gaussians from position or normal maps in UV or orthogonal spaces, enabling real-time photorealistic rendering, but are limited by base templates, preventing fine facial control.
To improve this, expressive templates like SMPL-X \citep{SMPL-X:2019} have been introduced.
ExAvatar \citep{moon2024exavatar} predicts pose-dependent Gaussian offsets on SMPL-X but misses dynamic clothing appearance.
Similarly, DEGAS \citep{shao2024degas} estimates Gaussian parameters from body poses and a visual expression encoder DPE \citep{pang2023dpe}, but is limited to image-based facial expression control.
%
%%%%%%%%%%%%%%%%%%%%%%%%%%%%%%%%%%%%%%%%%%%%%%%%%%%%%%%%%%%%%%%%%%%%%%%%%%%%%%%%%%%%%%%%%%%%%%
% Method high level
%%%%%%%%%%%%%%%%%%%%%%%%%%%%%%%%%%%%%%%%%%%%%%%%%%%%%%%%%%%%%%%%%%%%%%%%%%%%%%%%%%%%%%%%%%%%%%
%
\par
To address these challenges, we introduce \textbf{Expressive Virtual Avatars (EVA)}, a fully expressive and disentangled animatable avatar framework that generates real-time, photorealistic human renderings from body motion, expression parameters, and arbitrary viewpoints (see Fig.~\ref{fig:teaser}).
Unlike prior methods that either rely on mesh templates constrained to skeletal transformations and simple skinning,  or use parametric models like SMPL-X that lack accurate clothing dynamics, EVA uses a deformable and fully controllable template layer as a more expressive geometry proxy. 
Building on this foundation, we propose a disentangled Gaussian appearance layer, which provides fine-grained control over the body, hands, and face while enabling photorealistic rendering.
%
%%%%%%%%%%%%%%%%%%%%%%%%%%%%%%%%%%%%%%%%%%%%%%%%%%%%%%%%%%%%%%%%%%%%%%%%%%%%%%%%%%%%%%%%%%%%%%
% More details
%%%%%%%%%%%%%%%%%%%%%%%%%%%%%%%%%%%%%%%%%%%%%%%%%%%%%%%%%%%%%%%%%%%%%%%%%%%%%%%%%%%%%%%%%%%%%%
%
\par 
Our method leverages dense multi-view videos of an actor and corresponding motion tracking results during training. 
The approach begins with an expressive template geometry layer, constructing and tracking a fully controllable and deformable template. 
Inspired by \citet{joo2018total}, we propose a progressive deformation process to initialize a personalized head avatar and integrate it with a motion-driven deformable template \citep{habermann2021DDC} using a stitching-based method.
We then introduce a multi-stage tracking pipeline to register the template to multi-view videos. 
To enhance expression accuracy, we refine the body motion using multi-view facial landmarks and per-frame 3D reconstructions.
We then recover facial expression parameters exclusively from multi-view images.
Subsequently, we adopt the motion-based deformation learning approach of \citet{habermann2021DDC}, which recovers coarse clothing deformations.
To enable photo-realistic renderings, we construct a 3D Gaussian appearance layer in the template's 2D UV space. 
Unlike prior 3D Gaussian avatar models \citep{Pang_2024_CVPR, li2024animatablegaussians}, our disentangled Gaussian appearance layer separates body pose and facial expression, allowing independent parameter control.
Specifically, body and facial Gaussians are predicted separately using two 2D U-Nets, driven by root-centered body motions and facial expressions, respectively. 
Since no existing dataset captures diverse facial expressions, hand gestures, and body motions jointly, we introduce a new dataset featuring a wide range of body motions, detailed facial expressions, and varying hand gestures to validate our method and address limitations in available datasets. Our core contributions are summarized as:
%
%%%%%%%%%%%%%%%%%%%%%%%%%%%%%%%%%%%%%%%%%%%%%%%%%%%%%%%%%%%%%%%%%%%%%%%%%%%%%%%%%%%%%%%%%%%%%%
% Summary
%%%%%%%%%%%%%%%%%%%%%%%%%%%%%%%%%%%%%%%%%%%%%%%%%%%%%%%%%%%%%%%%%%%%%%%%%%%%%%%%%%%%%%%%%%%%%%
%
\begin{enumerate}
    \item We introduce \textbf{EVA}, a novel method enabling full-body control with real-time, photo-realistic renderings, robustly handling loose clothing dynamics and various facial expressions.
    \item We develop an \textbf{expressive deformable template} that generates a deformable human template mesh and employs a multi-stage tracking algorithm to faithfully capture  facial expressions, body motions, and non-rigid deformations from multi-view videos.
    \item We propose a \textbf{disentangled 3D Gaussian appearance module} that models the body and face independently, ensuring separated control and high-quality renderings.
\end{enumerate}
%
%%%%%%%%%%%%%%%%%%%%%%%%%%%%%%%%%%%%%%%%%%%%%%%%%%%%%%%%%%%%%%%%%%%%%%%%%%%%%%%%%%%%%%%%%%%%%%
%
EVA outperforms state-of-the-art approaches, as demonstrated by qualitative and quantitative evaluations, which highlight its superiority in achieving expressive and photorealistic human rendering.
%
%%%%%%%%%%%%%%%%%%%%%%%%%%%%%%%%%%%%%%%%%%%%%%%%%%%%%%%%%%%%%%%%%%%%%%%%%%%%%%%%%%%%%%%%%%%%%%

%%%%%%%%%%%%%%%%%%%%%%%%%%%%%%%%%%%%%%%%%%%%%%%%%%%%%%%%%%%%%%%%%%%%%%%%%%%%%%%%%%%%%%%%%%%%%%
%
\section{Related Work} \label{sec:related}
%
%%%%%%%%%%%%%%%%%%%%%%%%%%%%%%%%%%%%%%%%%%%%%%%%%%%%%%%%%%%%%%%%%%%%%%%%%%%%%%%%%%%%%%%%%%%%%%
%
Since our focus is on \textit{animatable human rendering}, human reconstruction methods \citep{peng2021neural, kwon2021neural,remelli2022drivable, xiang2023drivable,shetty2024holoported, sun2024metacap, sun2025real} that use image inputs at inference will not be discussed.
%
%%%%%%%%%%%%%%%%%%%%%%%%%%%%%%%%%%%%%%%%%%%%%%%%%%%%%%%%%%%%%%%%%%%%%%%%%%%%%%%%%%%%%%%%%%%%%%
%
\paragraph{Expressive Template Modeling.}\label{subsec:rw_exp} 
%
%%%%%%%%%%%%%%%%%%%%%%%%%%%%%%%%%%%%%%%%%%%%%%%%%%%%%%%%%%%%%%%%%%%%%%%%%%%%%%%%%%%%%%%%%%%%%%
%
Unlike earlier methods that model humans solely using body skeletons \citep{SMPL:2015, habermann2021DDC} or separately model the body, hands \citep{romero2022embodied}, and face \citep{blanz1999morphable,FLAME:SiggraphAsia2017}, expressive human templates aim to provide a fully controllable and unified human model. 
Early marker-based systems \citep{vicon}, widely used in the film industry, capture expressive motions but lack scalability. 
In the academic domain, parametric human templates \citep{SMPL-X:2019, xu2020ghum} and person-specific templates \citep{joo2018total} were introduced, modeling humans with template meshes and corresponding pose and expression parameters. 
These templates also paved the way for markerless motion capture techniques \citep{rong2020frankmocap, xiang2019monocular, zhou2021monocular, li2023hybrik, zhang2023pymaf}.
Although expressive parametric human templates can capture diverse motions of the body, hands, and face, they often neglect clothing and its associated non-rigid dynamics. 
In contrast, our method combines a motion-driven deformable template \citep{habermann2021DDC} incorporating learned pose-dependent clothing dynamics with a parametric head model \citep{FLAME:SiggraphAsia2017}.
This hybrid approach enables full controllability of human motions and expressions while capturing surface geometry and non-rigid deformations.
%
%%%%%%%%%%%%%%%%%%%%%%%%%%%%%%%%%%%%%%%%%%%%%%%%%%%%%%%%%%%%%%%%%%%%%%%%%%%%%%%%%%%%%%%%%%%%%%
%
\paragraph{Animatable Human Rendering.}\label{subsec:rw_ani}
%
%%%%%%%%%%%%%%%%%%%%%%%%%%%%%%%%%%%%%%%%%%%%%%%%%%%%%%%%%%%%%%%%%%%%%%%%%%%%%%%%%%%%%%%%%%%%%%
%
In pursuit of controllability and photorealistic rendering, researchers have explored various avatar representations, typically anchored to animatable templates. 
A wide range of prior work has focused on head-only avatars \citep{xu2023gaussianheadavatar,wang2024gaussianhead,10.1145/3687927,liao2023hhavatar,saito2024rgca,chen2024monogaussianavatar}. While these approaches have shown impressive results, they typically depend on constrained capture setups, static body poses, and specialized rigs, which limit their ability to fully capture entire humans. For this reason, we do not compare to them, as such a comparison would be unfair in both directions.
\par
A simple full-body representation utilizes a skinned mesh paired with a static texture \citep{alldieck20183DV}. 
However, it lacks photorealism, as dynamic models are needed for clothing deformations and appearance changes.
For instance, \citet{xu2011video} addressed dynamic appearances through a database-driven retrieval approach, while \citet{habermann2021DDC} proposed a learning- and template-based method, mapping motion inputs to template deformation parameters and dynamic textures.
Although mesh-based methods are practical, such as their seamless integration into classic graphics pipelines, they are fundamentally constrained by their representation.
\par 
The advent of neural rendering \citep{thies2019deferred, mildenhall2021nerf, kerbl3Dgaussians} has revolutionized high-fidelity renderings.
Techniques such as neural radiance fields (NeRFs) \citep{mildenhall2021nerf} have been incorporated into human templates for superior visual quality \citep{peng2021animatable, Habermann2023}.
Similarly, methods like \citet{Pang_2024_CVPR} and \citet{li2024animatablegaussians} leverage 3D Gaussians in UV or orthogonal projection spaces to achieve realistic renderings with real-time or near real-time performance.
%
%%%%%%%%%%%%%%%%%%%%%%%%%%%%%%%%%%%%%%%%%%%%%%%%%%%%%%%%%%%%%%%%%%%%%%%%%%%%%%%%%%%%%%%%%%%%%%
%
\begin{figure*}[bth]
    \centering
    \includegraphics[trim={0.6cm 7.03cm 1.6cm 3cm},clip,width=1\linewidth]{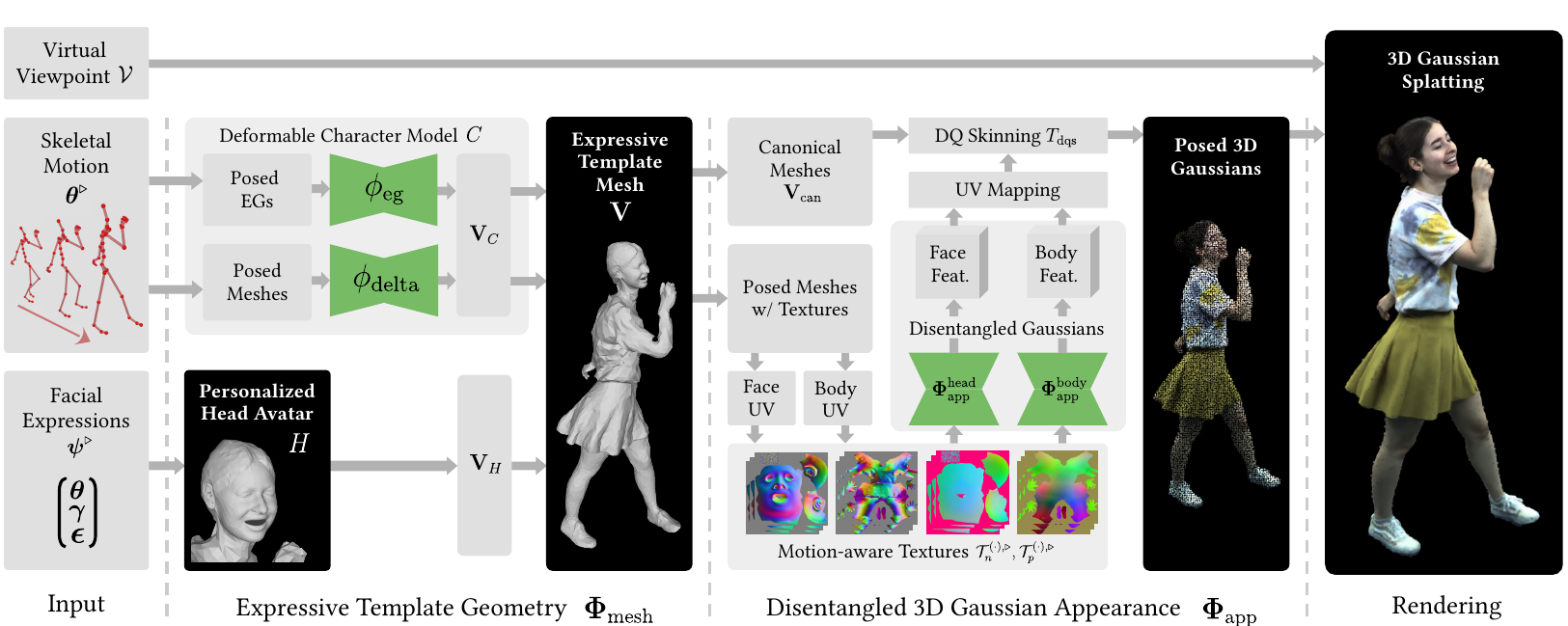}
    \caption[Method Overview: Expressive Virtual Avatars]{\textbf{Method Overview.} EVA generates high-fidelity renderings from a virtual viewpoint, skeletal motion, and expression parameters. Using a personalized head avatar and a deformable character model, we control body movements and facial expressions
    to drive an actor-specific mesh. 
    This mesh generates motion-aware textures, and separate modules independently predict the Gaussian parameters for the face and body. 
    The 3D Gaussians are combined, UV-mapped, and warped from canonical to posed space via dual quaternion skinning.
    Finally, they are splatted to render the final photorealistic image.
    }
    \label{fig:overview}
\end{figure*}
%
%%%%%%%%%%%%%%%%%%%%%%%%%%%%%%%%%%%%%%%%%%%%%%%%%%%%%%%%%%%%%%%%%%%%%%%%%%%%%%%%%%%%%%%%%%%%%%
%
While these methods deliver high-quality renderings, their controllability is constrained by the base templates.
To enable full control over the face, hands, and body, recent works \citep{shen2023x,zheng2023avatarrex, shao2024degas} replaced template models with parametric templates, such as SMPL-X \citep{SMPL-X:2019}. However, they still are not capable of accurately modeling complex clothing dynamics.
X-Avatar \citep{shen2023x} introduced the first expressive implicit avatar with part-aware initialization and sampling but suffers from slow inference due to its volumetric rendering.
ExAvatar \citep{moon2024exavatar} mounts 3D Gaussians \citep{kerbl3Dgaussians} onto SMPL-X and predicts pose-dependent offsets but fails to capture the dynamic appearance of clothing.
The concurrent work DEGAS \citep{shao2024degas} estimates Gaussian parameters based on body pose and expressions but relies on the expression encoder DPE \citep{pang2023dpe}, which requires front-facing images for facial control. 
Consequently, DEGAS is less flexible when integrating diverse driving signals. Its parametric template further limits the performance on loose cloth.
In contrast, our method takes facial expression parameters as direct inputs, supporting diverse multi-modal driving signals, such as expression control from audio \citep{VOCA2019}.
Furthermore, EVA's Gaussian parameters are predicted based on body pose and facial expression, enabling more expressive and adaptable appearances. This stands in contrast to prior approaches that rely solely on optimizing static 3D Gaussians attached to the FLAME mesh \cite{qian2024gaussianavatars}.
Unlike existing SMPL-X-based methods, which struggle to model loose clothing, our approach introduces a novel expressive and deformable human template that accounts for cloth dynamics. 
Additionally, our representation is more efficient, leveraging CNNs to learn 3D Gaussians in UV space, enabling real-time performance.
%
%%%%%%%%%%%%%%%%%%%%%%%%%%%%%%%%%%%%%%%%%%%%%%%%%%%%%%%%%%%%%%%%%%%%%%%%%%%%%%%%%%%%%%%%%%%%%%

%%%%%%%%%%%%%%%%%%%%%%%%%%%%%%%%%%%%%%%%%%%%%%%%%%%%%%%%%%%%%%%%%%%%%%%%%%%%%%%%%%%%%%%%%%%%%%
%
\section{Method} \label{sec:method}
%
%%%%%%%%%%%%%%%%%%%%%%%%%%%%%%%%%%%%%%%%%%%%%%%%%%%%%%%%%%%%%%%%%%%%%%%%%%%%%%%%%%%%%%%%%%%%%%
%
Given a virtual viewpoint $\mathcal{V}$, a window $W$ of skeletal poses (skeletal motion) $\boldsymbol{\theta}^\triangleright= \boldsymbol{\theta}_t^\triangleright = [\boldsymbol{\theta}_{t-W}, \dots,\boldsymbol{\theta}_{t}]^\top \in \mathbb{R}^{W \times D_\mathrm{body}}$, and the corresponding window of facial expression parameters $\boldsymbol{\psi}^\triangleright =\boldsymbol{\psi}_t^\triangleright =  [\boldsymbol{\psi}_{t-W}, \dots,\boldsymbol{\psi}_{t}]^\top  \in \mathbb{R}^{W \times D_\mathrm{face}}$, where $t$ denotes the current frame, and $D_\mathrm{body}$ and $D_\mathrm{face}$ the dimensionality of body pose and facial expression parameters, respectively, our method generates photorealistic real-time renderings of an actor at over $30$ frames per second (fps) at inference. 
Here, the operator $(\cdot^\triangleright)$ denotes a temporal window.
During training, EVA leverages dense-view videos capturing an actor in a controlled studio. 
Approximately $100$ cameras capture full-body views, while around $20$ focus on close-up shots of the face.
The videos are segmented~\citep{kirillov2023segany}, and skeletal poses are initially tracked~\citep{thecaptury}. 
The final skeletal poses $\boldsymbol{\theta}$ and facial expressions $\boldsymbol{\psi}$ are obtained through our dedicated human tracking pipeline, which is explained in detail later.
EVA is composed of two core modules (Fig.~\ref{fig:overview}): 
(1) an \textit{expressive template geometry layer} $\boldsymbol{\Phi}_\mathrm{mesh}$ that constructs an expressive and deformable human template. 
This layer enables full-body control using skeletal poses and facial expressions as inputs;
and (2) a \textit{3D Gaussian appearance layer} $\boldsymbol{\Phi}_\mathrm{app}$ that disentangles the body and facial appearance and generates photorealistic renderings.
In the following sections, we first cover essential preliminaries (Sec.~\ref{sec:prelim}).
We then describe the construction of the expressive template geometry layer and the methods used for accurately tracking skeletal poses $\boldsymbol{\theta}$ and facial expressions $\boldsymbol{\psi}$ (Sec.~\ref{sec:mesh}).
Finally, we present our disentangled Gaussian appearance module for photorealistic rendering (Sec.~\ref{sec:gaussian}), allowing separate control over face and body appearance.
%
%%%%%%%%%%%%%%%%%%%%%%%%%%%%%%%%%%%%%%%%%%%%%%%%%%%%%%%%%%%%%%%%%%%%%%%%%%%%%%%%%%%%%%%%%%%%%%
%
\subsection{Preliminaries}\label{sec:prelim}
%
%%%%%%%%%%%%%%%%%%%%%%%%%%%%%%%%%%%%%%%%%%%%%%%%%%%%%%%%%%%%%%%%%%%%%%%%%%%%%%%%%%%%%%%%%%%%%%
%
To capture pose-dependent deformations of the body, we leverage the \textbf{DDC} model~\citep{habermann2021DDC}:
\begin{equation}
    C({\boldsymbol{\theta}_t^\triangleright}) = T_{\text{dqs}}(\boldsymbol{\theta}_t,
    \phi_\mathrm{eg}(\hat{\boldsymbol{\theta}}^\triangleright_t,\mathbf{V}_\mathrm{T}) +\phi_\mathrm{delta}(\hat{\boldsymbol{\theta}}_t^\triangleright) ) =  \mathbf{V}_{C,t} \in \mathbb{R}^{N_C \times 3},
\end{equation}
which takes the skeletal motion as input and outputs $N_C$ posed and deformed template vertices $\mathbf{V}_{C,t}$. Here, $\mathbf{V}_\mathrm{T}$ denotes the vertices in the unposed space, and our skeleton definition also includes hand joints.
Given a window of normalized (subtracted root translation) skeletal poses $\hat{\boldsymbol{\theta}}^\triangleright_t$, an embedded graph layer $\phi_{\mathrm{eg}}(\cdot)$ first coarsely deforms the unposed base template $\mathbf{V}_\mathrm{T}$, and then a displacement network $\phi_{\mathrm{delta}}(\cdot)$ estimates additional finer per-vertex displacements.
Finally, we skin the deformed vertices using dual quaternion skinning~\citep{Kavan2007} $T_\text{dqs}(\cdot)$ to obtain the final posed vertices $\mathbf{V}_{C,t}$.
%
%%%%%%%%%%%%%%%%%%%%%%%%%%%%%%%%%%%%%%%%%%%%%%%%%%%%%%%%%%%%%%%%%%%%%%%%%%%%%%%%%%%%%%%%%%%%%%
%
\par
%
%%%%%%%%%%%%%%%%%%%%%%%%%%%%%%%%%%%%%%%%%%%%%%%%%%%%%%%%%%%%%%%%%%%%%%%%%%%%%%%%%%%%%%%%%%%%%%
%
To represent human head geometry and expressions, we build upon the \textbf{FLAME} model~\citep{FLAME:SiggraphAsia2017}, a learned parametric head model defined as $H_\mathrm{FLAME}(\boldsymbol{\beta}, \boldsymbol{\theta}_H, \boldsymbol{\gamma}, \boldsymbol{\epsilon})= \mathbf{V}_H \in \mathbb{R}^{N_H\times 3}$.
This function takes shape parameters $\boldsymbol{\beta} \in \mathbb{R}^{300}$, pose parameters $\boldsymbol{\theta}_H \in \mathbb{R}^{18}$, expression parameters $\boldsymbol{\gamma} \in \mathbb{R}^{100}$, and eyelid parameters $\boldsymbol{\epsilon} \in \mathbb{R}^2$ \citep{zielonka2022mica,Zielonka2022InstantVH} as inputs and outputs $N_H$ deformed head vertices $\mathbf{V}_H$.
The pose parameters include global translation $\boldsymbol{\theta}_{\mathrm{trans}}$, global rotation $\boldsymbol{\theta}_{\mathrm{rot}}$, and joint articulations for the neck joint $\boldsymbol{\theta}_{\mathrm{neck}}$, jaw joint $\boldsymbol{\theta}_{\mathrm{jaw}}$, and the two eyeball joints $\boldsymbol{\theta}_{\mathrm{eyes}}$.
%
%%%%%%%%%%%%%%%%%%%%%%%%%%%%%%%%%%%%%%%%%%%%%%%%%%%%%%%%%%%%%%%%%%%%%%%%%%%%%%%%%%%%%%%%%%%%%%
%
\par
%
%%%%%%%%%%%%%%%%%%%%%%%%%%%%%%%%%%%%%%%%%%%%%%%%%%%%%%%%%%%%%%%%%%%%%%%%%%%%%%%%%%%%%%%%%%%%%%
%
We model appearance using \textbf{3D Gaussians}~\citep{kerbl3Dgaussians} parameterized as $\mathcal{G} = (\boldsymbol{\mu}, \mathbf{s}, \mathbf{q}, \alpha, \boldsymbol{\eta})$, where $\boldsymbol{\mu} \in \mathbb{R}^3$ represents its 3D position, $\mathbf{s} \in \mathbb{R}^3$ the anisotropic scale, $\mathbf{q} \in \mathbb{R}^4$ the dual quaternion rotation, $\alpha \in \mathbb{R}$ the opacity, and $\boldsymbol{\eta} \in \mathbb{R}^{48}$ the spherical harmonics coefficients.
The 3D Gaussians are projected into 2D \citep{Zwicker2001} by updating their covariance matrix as $\boldsymbol{\Sigma}^\mathcal{V} = \mathbf{J}\mathbf{P}\mathbf{R}\mathbf{S}\mathbf{S}^T\mathbf{R}^T\mathbf{P}^T\mathbf{J}^T$, where $\mathbf{J}$ is the Jacobian of the affine approximation of the projective transformation $\mathbf{P}$.
Next, we compute colors of each pixel through blending $N_\mathrm{b}$ sorted Gaussians, $\mathbf{c} = \sum_{n\in N_\mathrm{b}}{SH}(\boldsymbol{\eta}_\mathrm{n},\mathcal{V},\boldsymbol{\mu}_{\mathrm{n}})\boldsymbol{\alpha}_{n}^{'}\prod_{m=1}^{n-1}(1-\boldsymbol{\alpha}_{m}^{'})$, where $SH(\cdot)$ transfers spherical harmonics $\boldsymbol{\eta}$ to RGB colors based on their positions $\boldsymbol{\mu}$ and camera views $\mathcal{V}$. $\alpha^{'}$ are blending weights.
%
%%%%%%%%%%%%%%%%%%%%%%%%%%%%%%%%%%%%%%%%%%%%%%%%%%%%%%%%%%%%%%%%%%%%%%%%%%%%%%%%%%%%%%%%%%%%%%
%
\par
%
%%%%%%%%%%%%%%%%%%%%%%%%%%%%%%%%%%%%%%%%%%%%%%%%%%%%%%%%%%%%%%%%%%%%%%%%%%%%%%%%%%%%%%%%%%%%%%
%
To enable efficient and effective learning of \textbf{Animatable Gaussians}, we follow ASH~\citep{Pang_2024_CVPR} and frame the learning process of Gaussians as an image-to-image translation task.
Specifically, we embed the appearance models into 2D texture space, where each texel corresponds to a Gaussian splat in the canonical pose.
To transform the estimated Gaussian parameters into world space, we render the deformed canonical mesh $V_\mathrm{can}$, introduced in Sec.~\ref{sec:mesh}, into two separate texture spaces for the body and head. 
The final position of a single Gaussian is computed by taking the canonical vertices in 2D texture space as the base position, adding the learned offset $\mathbf{d}_\mathrm{uv}$, and applying dual quaternion skinning \citep{Kavan2007} as $\boldsymbol{\mu}_\mathrm{uv} = T_\mathrm{dqs}((w_a \cdot v_{\mathrm{can},a} + w_b \cdot v_{\mathrm{can},b}+ w_c \cdot v_{\mathrm{can},c})+\mathbf{d}_\mathrm{uv})$
%
%%%%%%%%%%%%%%%%%%%%%%%%%%%%%%%%%%%%%%%%%%%%%%%%%%%%%%%%%%%%%%%%%%%%%%%%%%%%%%%%%%%%%%%%%%%%%%
%
\subsection{Expressive Human Template Modeling}\label{sec:mesh}
%
%%%%%%%%%%%%%%%%%%%%%%%%%%%%%%%%%%%%%%%%%%%%%%%%%%%%%%%%%%%%%%%%%%%%%%%%%%%%%%%%%%%%%%%%%%%%%%
%
As illustrated in recent approaches~\citep{Pang_2024_CVPR,li2024animatablegaussians,shao2024degas}, a controllable geometric proxy is a key component in the control of 3D Gaussian avatars.
However, the templates used in prior methods are either not expressive or non-deformable, leading to limited expressivity and rendering performance.
Thus, we first introduce an expressive \emph{and} deformable human template.
For each subject, we define the human template as the function $\boldsymbol{\Phi}_{\mathrm{mesh}} ({\boldsymbol{\theta}^\triangleright}, \boldsymbol{\psi}) = \boldsymbol{\Phi}_{\mathrm{mesh},t} ({\boldsymbol{\theta}^\triangleright_t}, \boldsymbol{\psi}_t) = \mathbf{V} \in \mathbb{R}^{N\times3}$, which, given a window of skeletal poses ${\boldsymbol{\theta}^\triangleright}$ and facial expression parameters $\boldsymbol{\psi}$, deforms and poses the template mesh to vertices $\mathbf{V}$, where $N$ denotes the number of vertices. By using a temporal window of skeletal poses, the template implicitly captures motion dynamics through velocity and acceleration.
Setting the skeletal pose parameters to the canonical pose while retaining body deformations results in a deformed canonical mesh with vertices $V_\mathrm{can}$, which is used later in our method.
Specifically, we construct $\boldsymbol{\Phi}_{\mathrm{mesh}}$ through a three-stage process:
(1) learning a character model $C({\boldsymbol{\theta}^\triangleright})$ that captures body motion and hand gestures, while also modeling motion-dependent deformations leveraging the window of skeletal poses ${\boldsymbol{\theta}^\triangleright}$; 
(2) constructing the head model to generate a {personalized head avatar} $H(\boldsymbol{\psi})$; 
and (3) seamlessly stitching ($\bowtie$) the body and head together to form a complete and cohesive full-body mesh $\boldsymbol{\Phi}_{\text{mesh}} ({\boldsymbol{\theta}^\triangleright}, \boldsymbol{\psi}) = C({\boldsymbol{\theta}^\triangleright}) \bowtie H(\boldsymbol{\psi})$. 
For details regarding the stitching, refer to our supplemental document.
%
%%%%%%%%%%%%%%%%%%%%%%%%%%%%%%%%%%%%%%%%%%%%%%%%%%%%%%%%%%%%%%%%%%%%%%%%%%%%%%%%%%%%%%%%%%%%%%
%
\paragraph{Deformable Character Model}
%
%%%%%%%%%%%%%%%%%%%%%%%%%%%%%%%%%%%%%%%%%%%%%%%%%%%%%%%%%%%%%%%%%%%%%%%%%%%%%%%%%%%%%%%%%%%%%%
%
The foundation of our deformable character model is a rigged and skinned template mesh $\mathbf{V}_\mathrm{T}$, derived from a 3D scan.
To account for non-rigid, motion-dependent deformations, we utilize the GNN-based deformation model $C({\boldsymbol{\theta}_t^\triangleright})$ introduced by \citet{habermann2021DDC} (Sec.~\ref{sec:prelim}).
%
%%%%%%%%%%%%%%%%%%%%%%%%%%%%%%%%%%%%%%%%%%%%%%%%%%%%%%%%%%%%%%%%%%%%%%%%%%%%%%%%%%%%%%%%%%%%%%
%
For training both GNNs, we follow \citet{habermann2021DDC} and refer to the original work and our supplemental document for details.
However, we introduce two important changes.
First, to improve deformation estimation, we follow~\citep{shetty2024holoported} by providing additional geometric supervision from 3D scans~\citep{neus2} and removing the deformations applied to the hands.
Second, we improve motion tracking as described in the remainder of this subsection and leverage this motion for training.
So far, we have built a controllable template with hand gestures and body deformations, but it does not yet offer control over facial expressions.
%
%%%%%%%%%%%%%%%%%%%%%%%%%%%%%%%%%%%%%%%%%%%%%%%%%%%%%%%%%%%%%%%%%%%%%%%%%%%%%%%%%%%%%%%%%%%%%%
%
\paragraph{Personalized Head Avatar}
%
%%%%%%%%%%%%%%%%%%%%%%%%%%%%%%%%%%%%%%%%%%%%%%%%%%%%%%%%%%%%%%%%%%%%%%%%%%%%%%%%%%%%%%%%%%%%%%
%
To make the character model expressive, we also need to accurately capture the head geometry and facial expressions. 
Based on the observation that most deformations of the head come from facial expressions~\citep{qian2024gaussianavatars}, we decompose a personalized head avatar into initial displacement optimization and facial expression tracking. 
Note that, in contrast to previous work, our approach does not require the use of specialized face rigs. Furthermore, we deliberately refrain from capturing the head in a separate setup, instead learning the head model directly from the full-body capture. This allows us to significantly reduce the overall capture time, approximately halving it.
Specifically, we use the 3D morphable head model FLAME~\citep{FLAME:SiggraphAsia2017} as a head template (Sec.~\ref{sec:prelim}).
We extend the original FLAME model to the function
%
%%%%%%%%%%%%%%%%%%%%%%%%%%%%%%%%%%%%%%%%%%%%%%%%%%%%%%%%%%%%%%%%%%%%%%%%%%%%%%%%%%%%%%%%%%%%%%
%
\begin{equation}
    H\left(\boldsymbol{\beta}, \boldsymbol{\theta}_{\mathrm{trans}}, \boldsymbol{\theta}_{\mathrm{rot}}, \boldsymbol{\theta}_{\mathrm{neck}}, \boldsymbol{\theta}_{\mathrm{jaw}}, \boldsymbol{\theta}_{\mathrm{eyes}}, \boldsymbol{\gamma}, \boldsymbol{\epsilon}, \mathbf{D}\right) =  \mathbf{V}_H
\end{equation}
%
%%%%%%%%%%%%%%%%%%%%%%%%%%%%%%%%%%%%%%%%%%%%%%%%%%%%%%%%%%%%%%%%%%%%%%%%%%%%%%%%%%%%%%%%%%%%%%
%
with an additional per-vertex displacement $\mathbf{D} \in \mathbb{R}^{N_H\times 3}$, since the shape parameters of FLAME are insufficient for capturing fine, subject-specific details such as hair.
Next, we fit this model to the head region of our full-body scan to obtain a personalized head model.
To do this, we introduce a coarse-to-fine optimization strategy using gradient descent, consisting of three distinct stages.
%
%%%%%%%%%%%%%%%%%%%%%%%%%%%%%%%%%%%%%%%%%%%%%%%%%%%%%%%%%%%%%%%%%%%%%%%%%%%%%%%%%%%%%%%%%%%%%%
%
\par \textit{1) Coarse Alignment.} 
%
%%%%%%%%%%%%%%%%%%%%%%%%%%%%%%%%%%%%%%%%%%%%%%%%%%%%%%%%%%%%%%%%%%%%%%%%%%%%%%%%%%%%%%%%%%%%%%
%
Since the head scan and the FLAME model may be arbitrarily positioned and oriented with respect to each other, we first optimize the global translation $\boldsymbol{\theta}_\mathrm{trans} \in \mathbb{R}^3$ and rotation $\boldsymbol{\theta}_\mathrm{rot} \in \mathbb{R}^3$ by fitting the facial landmarks~\citep{Sagonas2013a, Sagonas2013b, SAGONAS20163} to their manually annotated counterparts on the 3D scan.
%
%%%%%%%%%%%%%%%%%%%%%%%%%%%%%%%%%%%%%%%%%%%%%%%%%%%%%%%%%%%%%%%%%%%%%%%%%%%%%%%%%%%%%%%%%%%%%%
%
\par
\begin{figure}[t]
    \centering
    \includegraphics[width=\linewidth]{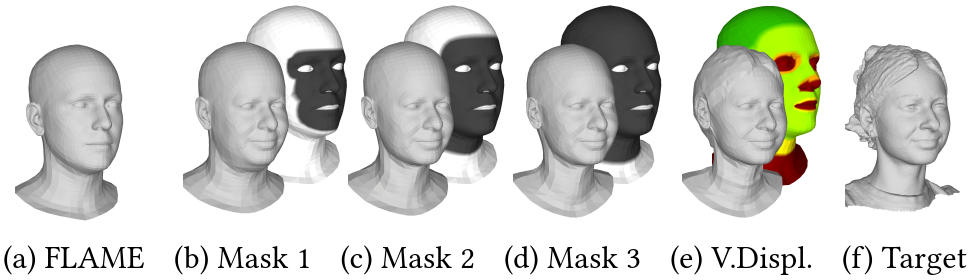}
    \caption{Illustration of the head fitting: (b-d) results after each iteration of our \textit{freeze and refine} optimization strategy, with (d) the optimized FLAME model in a neutral expression. Black regions indicate the optimized areas for each iteration. (e) displays added vertex displacements with corresponding fitting weights, where green indicates high and red low weights.}
    \label{fig:paramfit}
\end{figure}

 \textit{2) Shape Fitting.} 
%
%%%%%%%%%%%%%%%%%%%%%%%%%%%%%%%%%%%%%%%%%%%%%%%%%%%%%%%%%%%%%%%%%%%%%%%%%%%%%%%%%%%%%%%%%%%%%%
%
Next, we optimize the shape parameters $\boldsymbol{\beta} \in \mathbb{R}^{300}$, along with global translation $\boldsymbol{\theta}_{\mathrm{trans}} \in \mathbb{R}^3$ and rotation $\boldsymbol{\theta}_{\mathrm{rot}} \in \mathbb{R}^3$.
Unconstrained optimization using a bidirectional Chamfer distance often leads to misalignment, especially around facial landmarks, as all regions are weighted equally.
This would introduce significant distortions as FLAME does not model details at the back of the head such as hair.
To address this, we implement a \textit{freeze and refine} pipeline that incrementally refines larger head regions while freezing previously optimized vertices via strong regularization.
This is achieved using a masked Chamfer loss.
The pipeline involves three refinement steps, with masks shown in Fig.~\ref{fig:paramfit}.
Furthermore, using a one-sided Chamfer loss $\mathcal{L}^\rightharpoonup_\mathrm{Chamf}$ avoids minimizing distances for surface points on the FLAME mesh without corresponding vertices on the target scan.
Additionally, a 3D landmark loss $\mathcal{L}_{\mathrm{Lmk3D}}$ provides coarse regularization for facial feature alignment.
%
%%%%%%%%%%%%%%%%%%%%%%%%%%%%%%%%%%%%%%%%%%%%%%%%%%%%%%%%%%%%%%%%%%%%%%%%%%%%%%%%%%%%%%%%%%%%%%
%
\par \textit{3) Per-vertex Displacements.} 
%
%%%%%%%%%%%%%%%%%%%%%%%%%%%%%%%%%%%%%%%%%%%%%%%%%%%%%%%%%%%%%%%%%%%%%%%%%%%%%%%%%%%%%%%%%%%%%%
%
Now, we fit the per-vertex displacements $\mathbf{D}$ to the scan to recover person-specific details such as hair, which are not captured by FLAME's shape parameters.
We use a one-sided Chamfer loss $\mathcal{L}^\rightharpoonup_{\mathrm{Chamf}}$ to optimize these displacements.
To prevent deformations in well-modeled regions, we segment the FLAME mesh and assign region-specific fitting weights that scale the gradients.
Critical facial regions, such as the eyes and nose, are given lower weights, while hair (not modeled by FLAME) is given higher weights (Fig.~\ref{fig:paramfit}).
To compute the loss, we sample points uniformly from, both, the scan and the FLAME model.
%
%%%%%%%%%%%%%%%%%%%%%%%%%%%%%%%%%%%%%%%%%%%%%%%%%%%%%%%%%%%%%%%%%%%%%%%%%%%%%%%%%%%%%%%%%%%%%%
%
\par 
%
%%%%%%%%%%%%%%%%%%%%%%%%%%%%%%%%%%%%%%%%%%%%%%%%%%%%%%%%%%%%%%%%%%%%%%%%%%%%%%%%%%%%%%%%%%%%%%
%
Once we have optimized the shape $\boldsymbol{\beta}$, translation $\boldsymbol{\theta}_{\mathrm{trans}}$, rotation $\boldsymbol{\theta}_{\mathrm{rot}}$, and per-vertex displacements $\mathbf{D}$, we keep them constant. 
Additionally, the neck articulation is set to zero, as we will use the body motion to handle neck articulations. 
We also set the eyeball rotations $\boldsymbol{\theta}_\mathrm{eyes}$ to zero, as our tracking with landmarks and scans cannot capture eye gaze.
Thus, after optimizing the head model, we can remove unnecessary variables from the head avatar function, simplifying it to $H\left(\boldsymbol{\psi} \right)=H\left(\boldsymbol{\theta}_{\mathrm{jaw}}, \boldsymbol{\gamma}, \boldsymbol{\epsilon} \right) = \mathbf{V}_H$.
Next, we refine the skeletal motion tracking and track facial expressions from multi-view images.
%
%%%%%%%%%%%%%%%%%%%%%%%%%%%%%%%%%%%%%%%%%%%%%%%%%%%%%%%%%%%%%%%%%%%%%%%%%%%%%%%%%%%%%%%%%%%%%%
%
%
%%%%%%%%%%%%%%%%%%%%%%%%%%%%%%%%%%%%%%%%%%%%%%%%%%%%%%%%%%%%%%%%%%%%%%%%%%%%%%%%%%%%%%%%%%%%%%
%
%%%%%%%%%%%%%%%%%%%%%%%%%%%%%%%%%%%%%%%%%%%%%%%%%%%%%%%%%%%%%%%%%%%%%%%%%%%%%%%%%%%%%%%%%%%%%%
%
\paragraph{Motion Optimization \& Expression Tracking}
%
%%%%%%%%%%%%%%%%%%%%%%%%%%%%%%%%%%%%%%%%%%%%%%%%%%%%%%%%%%%%%%%%%%%%%%%%%%%%%%%%%%%%%%%%%%%%%%
%
We observed that markerless motion capture \citep{thecaptury}, used in recent state-of-the-art avatar approaches \citep{Habermann2023,Pang_2024_CVPR, habermann2021DDC}, often suffers from inaccurate head pose tracking and overall imprecise body poses. 
These inaccuracies can result in suboptimal expression tracking and negatively affect the training of our appearance model, which is introduced later. 
To address this issue, we refine the initial skeletal poses $\boldsymbol{\theta}_\mathrm{init}$ to better align the facial region and overall posed geometry of the virtual avatar mesh with multi-view images. 
We begin by selecting camera views where the face is clearly visible, ensuring that the angular difference between the inverse camera viewing direction and the head orientation is less than $70$ degrees.
For the selected cameras, we use Mediapipe (MP) \citep{mediapipe} and FAN \citep{bulat2017far} to predict 2D facial landmarks for all selected views, with MP providing dense coverage and FAN offering confidence-scored landmarks.
Next, we refine the skeletal poses $\boldsymbol{\theta}_\mathrm{init}$ by minimizing the bidirectional Chamfer loss $\mathcal{L}^\rightleftharpoons_{\mathrm{Chamf}}$, the 2D landmark loss $\mathcal{L}_{\mathrm{Lmk2D}}$, and a total variation loss $\mathcal{L}_{\mathrm{TV}}$ for temporal smoothing. We refer to our supplementary document for loss function definitions and weightings.
This strategy enhances facial tracking while also refining the entire human pose, leading to more precise motion capture.
%
%%%%%%%%%%%%%%%%%%%%%%%%%%%%%%%%%%%%%%%%%%%%%%%%%%%%%%%%%%%%%%%%%%%%%%%%%%%%%%%%%%%%%%%%%%%%%%
%
\par
Once we have obtained the optimized skeletal poses $\boldsymbol{\theta}$ and the personalized head avatar $H$, we leverage 2D landmarks to track the facial expression parameters $\boldsymbol{\psi}$, including $\boldsymbol{\theta}_\mathrm{jaw}$, $\boldsymbol{\gamma}$, and $\boldsymbol{\epsilon}$. 
By combining MP and FAN landmarks, we take advantage of each of their strengths. 
Specifically, MP landmarks are used to track the eyelids, as FAN struggles to accurately detect them when the eyes are nearly closed. 
We also employ a specialized computation to enhance the eyelid parameters $\boldsymbol{\epsilon}$ for precise eye blink capture. 
For additional details, we refer to our supplementary document.
%
%%%%%%%%%%%%%%%%%%%%%%%%%%%%%%%%%%%%%%%%%%%%%%%%%%%%%%%%%%%%%%%%%%%%%%%%%%%%%%%%%%%%%%%%%%%%%%
%
\subsection{Disentangled Gaussian Prediction}\label{sec:gaussian}
%
%%%%%%%%%%%%%%%%%%%%%%%%%%%%%%%%%%%%%%%%%%%%%%%%%%%%%%%%%%%%%%%%%%%%%%%%%%%%%%%%%%%%%%%%%%%%%%
%
From Sec.~\ref{sec:mesh}, we built an expressive, deformable character mesh $\boldsymbol{\Phi}_{\mathrm{mesh}}$, enabling disentangled control of skeletal motion $\boldsymbol{\theta}^\triangleright$ and facial expressions $\boldsymbol{\psi}$.
However, it currently lacks photorealistic renderings. 
Following previous work~\citep{pang2023dpe}, we model the appearance as 3D Gaussians in the texture space.
In this subsection, we introduce our dynamic appearance module for clothed humans through disentangled animatable 3D Gaussians.
%
%%%%%%%%%%%%%%%%%%%%%%%%%%%%%%%%%%%%%%%%%%%%%%%%%%%%%%%%%%%%%%%%%%%%%%%%%%%%%%%%%%%%%%%%%%%%%%
%
\paragraph{Disentangled Animatable Gaussians.} 
%
%%%%%%%%%%%%%%%%%%%%%%%%%%%%%%%%%%%%%%%%%%%%%%%%%%%%%%%%%%%%%%%%%%%%%%%%%%%%%%%%%%%%%%%%%%%%%%
%
To represent appearance, we utilize Gaussian splatting~\citep{kerbl3Dgaussians} for its high quality and fast rendering performance (Sec.~\ref{sec:prelim}).
Similar to previous methods~\citep{Pang_2024_CVPR,li2024animatablegaussians}, we aim to map skeletal poses to a set of 3D Gaussians $\{\mathcal{G}_i\}_{N_\mathrm{G}}$ to model pose-dependent appearance changes. 
However, we also extend this approach by mapping expression parameters to Gaussian parameters, enabling the modeling of expression-dependent appearance effects, which prior works often neglect.
DEGAS~\citep{shao2024degas}, for example, maps visual features to Gaussian parameters but lacks explicit control via facial expression parameters. 
The main challenge lies in decoupling body motion- and expression-dependent appearance effects to enable independent control.
To address this, we propose two disentangled Gaussian models that form our expressive rendering module:
%
%%%%%%%%%%%%%%%%%%%%%%%%%%%%%%%%%%%%%%%%%%%%%%%%%%%%%%%%%%%%%%%%%%%%%%%%%%%%%%%%%%%%%%%%%%%%%%
%
\begin{align}\label{eqn:gaussians}
    \Phi_\mathrm{app}({(\boldsymbol{\theta}^\triangleright)^\triangleright}, \boldsymbol{\psi}^\triangleright) &= \Phi_\mathrm{app}^\mathrm{body}(C^\triangleright({\hat{\boldsymbol{\theta}}^\triangleright})) \cup  \Phi_\mathrm{app}^\mathrm{head}(H^\triangleright(\boldsymbol{\psi} ),\hat{\boldsymbol{\vartheta}}^\triangleright) \\
    &= 
    \{\mathcal{G}_i\}_{i=0}^{{N_\mathrm{G}^\mathrm{body}}}
    \cup
    \{\mathcal{G}_j\}_{j=0}^{{N_\mathrm{G}^\mathrm{head}}}.
\end{align}
%
%%%%%%%%%%%%%%%%%%%%%%%%%%%%%%%%%%%%%%%%%%%%%%%%%%%%%%%%%%%%%%%%%%%%%%%%%%%%%%%%%%%%%%%%%%%%%%
%
Here, $\Phi_\mathrm{app}^\mathrm{body}(\cdot)$ predicts the Gaussian parameters for the body based on the motion-dependent, non-rigidly deformable character model. 
$C^\triangleright({\hat{\boldsymbol{\theta}^\triangleright}})$ denotes a window of normalized, posed, and deformed character meshes, each requiring a normalized motion window $\hat{\boldsymbol{\theta}^\triangleright}$. Therefore, the rendering module takes a concatenation of motion windows, which in practice corresponds to the motion window of the current frame $t$ and two previous frames $t-1$ and $t-2$. Consequently, $(\boldsymbol{\theta}^\triangleright)^\triangleright = (\boldsymbol{\theta}_{t}^\triangleright, \boldsymbol{\theta}_{t-1}^\triangleright, \boldsymbol{\theta}_{t-2}^\triangleright)$.
In contrast, $\Phi_\mathrm{app}^\mathrm{head}( \cdot)$ predicts Gaussian parameters for the head using the expressive head model, where $H^\triangleright(\boldsymbol{\psi} )$ represents a window of expressive head meshes. Notably, $\Phi_\mathrm{app}^\mathrm{head}( \cdot)$ also incorporates $\hat{\boldsymbol{\vartheta}}^\triangleright$, a normalized window of the head's global rotation and translation, derived from the skeletal body pose, to account for variations in facial appearance due to changing lighting conditions within the capturing studio.
Importantly, employing windows of body and head meshes for the two previous frames implicitly encodes surface dynamics (velocity and acceleration).
$N_\mathrm{G}^\mathrm{body}$ and $N_\mathrm{G}^\mathrm{head}$ denote the number of Gaussians for the body and the head, respectively.
Importantly, the entire appearance is controlled solely by the skeletal poses ${\boldsymbol{\theta}}$ and facial expressions $\boldsymbol{\psi}$.
%
%%%%%%%%%%%%%%%%%%%%%%%%%%%%%%%%%%%%%%%%%%%%%%%%%%%%%%%%%%%%%%%%%%%%%%%%%%%%%%%%%%%%%%%%%%%%%%
%
We use the body texture and head texture of our expressive mesh template $\boldsymbol{\Phi}_\text{mesh}$ to represent the respective Gaussians:
%
%%%%%%%%%%%%%%%%%%%%%%%%%%%%%%%%%%%%%%%%%%%%%%%%%%%%%%%%%%%%%%%%%%%%%%%%%%%%%%%%%%%%%%%%%%%%%%
%
\begin{equation}
    \{{\mathcal{G}_i}\}_{N_\mathrm{G}^{(\cdot)}} = \{(\boldsymbol{}{\boldsymbol{\mu}}_\mathrm{uv}, \mathbf{s}_\mathrm{uv}, \mathbf{q}_\mathrm{uv}, \alpha_{\mathrm{uv}}, \boldsymbol{\eta}_\mathrm{uv})_i\}_{i=0}^{{N_\mathrm{G}^{(\cdot)}}} \in \mathbb{R}^{N_\mathrm{G}^{(\cdot)} \times 59}.
\end{equation}
%
%%%%%%%%%%%%%%%%%%%%%%%%%%%%%%%%%%%%%%%%%%%%%%%%%%%%%%%%%%%%%%%%%%%%%%%%%%%%%%%%%%%%%%%%%%%%%%
%
\paragraph{Disentangled Gaussian Predictor}
%
%%%%%%%%%%%%%%%%%%%%%%%%%%%%%%%%%%%%%%%%%%%%%%%%%%%%%%%%%%%%%%%%%%%%%%%%%%%%%%%%%%%%%%%%%%%%%%
%
We employ two separate U-Nets \citep{Ronneberger2015} to predict appearance parameters $\phi_\mathrm{app}^\mathrm{body}$, $\phi_\mathrm{app}^\mathrm{head}$, and geometry parameters $\phi_\mathrm{geo}^\mathrm{body}$, $\phi_\mathrm{geo}^\mathrm{head}$ for the body and head, respectively.
Inspired by prior works \cite{habermann2021DDC, Habermann2023, Pang_2024_CVPR}, we render $C^\triangleright$ and $H^\triangleright$ (Eq.~\ref{eqn:gaussians}) into UV space using inverse texture mapping, generating windows of normal maps $\mathcal{T}_n^{\mathrm{body},\triangleright}$, $\mathcal{T}_n^{\mathrm{head},\triangleright}$ and position maps $\mathcal{T}_p^{\mathrm{body},\triangleright}$,$\mathcal{T}_p^{\mathrm{head},\triangleright}$. The meshes $H^\triangleright$ are first translated and rotated according to $\hat{\boldsymbol{\vartheta}}^\triangleright$.
To account for varying lighting conditions in large capture spaces, we additionally incorporate a global position latent $\Psi$ as an additional input to the appearance networks $\mathcal{E}_\mathrm{app}^\mathrm{body}$, $\mathcal{E}_\mathrm{app}^\mathrm{head}$ \citep{Pang_2024_CVPR}. 
This latent $\Psi$ is computed by passing the character's normalized root translation through a shallow MLP.
The predictors are formulated as follows:
%
%%%%%%%%%%%%%%%%%%%%%%%%%%%%%%%%%%%%%%%%%%%%%%%%%%%%%%%%%%%%%%%%%%%%%%%%%%%%%%%%%%%%%%%%%%%%%%
%
\begin{align}
    \phi_\mathrm{geo}^{(\cdot)} (\mathcal{T}_{n}^{(\cdot),\triangleright}, \mathcal{T}_{p}^{(\cdot),\triangleright}) &= \left(\mathbf{d}_{\mathrm{uv}}, \mathbf{s}_\mathrm{uv}, \mathbf{q}_\mathrm{uv}, \alpha_{\mathrm{uv}}\right)_{i},\\
    \phi_\mathrm{app}^{(\cdot)} (\mathcal{T}^{(\cdot),\triangleright}_{n}, \mathcal{T}^{(\cdot),\triangleright}_{p}, \Psi) &= (\boldsymbol{\eta}_\mathrm{uv})_{i}.
\end{align}
%
%%%%%%%%%%%%%%%%%%%%%%%%%%%%%%%%%%%%%%%%%%%%%%%%%%%%%%%%%%%%%%%%%%%%%%%%%%%%%%%%%%%%%%%%%%%%%%
%
\paragraph{Training Strategy} 
%
%%%%%%%%%%%%%%%%%%%%%%%%%%%%%%%%%%%%%%%%%%%%%%%%%%%%%%%%%%%%%%%%%%%%%%%%%%%%%%%%%%%%%%%%%%%%%%
%
Following ASH \citep{Pang_2024_CVPR}, we adopt an enhanced training approach. 
A warmup stage initializes the motion-aware texture decoders $\phi^{(\cdot)}_\text{geo}$ and $\phi^{(\cdot)}_\text{app}$ by learning pseudo-ground-truth 3D Gaussian parameters from evenly sampled training frames through minimizing an $\mathcal{L}_2$ loss.
In the training, we implement several enhancements:
(1) The sequence is preprocessed to maximize pose diversity by analyzing character motion and facial expression changes, prioritizing frames with significant variations instead of sampling every $10$th frame; 
(2) Training on image crops rather than full-frame images enables effective use of high resolutions;
(3) A randomized background color replaces ASH’s fixed black background, enhancing the 3D Gaussians' robustness to varying backgrounds; 
And (4) in addition to $\mathcal{L}_1$ and $\mathcal{L}_\mathrm{SSIM}$, we incorporate an IDMRF loss \citep{wang2018image} for improved rendering quality and apply positional regularization to prevent significant deviation of Gaussians from their original positions.
%
%%%%%%%%%%%%%%%%%%%%%%%%%%%%%%%%%%%%%%%%%%%%%%%%%%%%%%%%%%%%%%%%%%%%%%%%%%%%%%%%%%%%%%%%%%%%%%
%
\section{Results} \label{sec:experiments}
%
%%%%%%%%%%%%%%%%%%%%%%%%%%%%%%%%%%%%%%%%%%%%%%%%%%%%%%%%%%%%%%%%%%%%%%%%%%%%%%%%%%%%%%%%%%%%%%
%
\begin{figure*}[h]
    \centering
    \includegraphics[width=\textwidth]{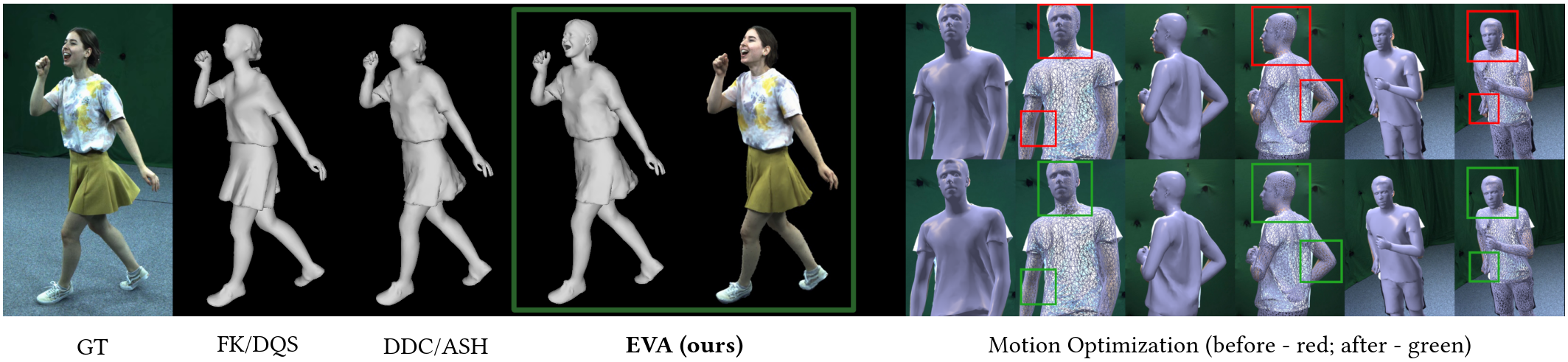}
    \caption{Qualitative result and comparison of our expressive mesh template and appearance (green box) with the ground truth image (GT), alongside results from a character animated using only an underlying skeleton with simple forward kinematics (FK) and dual quaternion skinning (DQS) \cite{Kavan2007}, as well as the template mesh employed by ASH \citep{Pang_2024_CVPR}, which closely resembles the standard DDC \citep{habermann2021DDC} mesh but includes control over the hands. Our approach not only controls body pose and hand gestures but also defines facial expressions. Additionally, we showcase the outcomes of our motion optimization strategy, which significantly enhances the alignment between our expressive template mesh and the underlying multi-view images.}
\label{fig:mesh_results}
\end{figure*}
%
%%%%%%%%%%%%%%%%%%%%%%%%%%%%%%%%%%%%%%%%%%%%%%%%%%%%%%%%%%%%%%%%%%%%%%%%%%%%%%%%%%%%%%%%%%%%%%
%
\begin{figure*}[h]
    \centering
    \includegraphics[width=\textwidth]{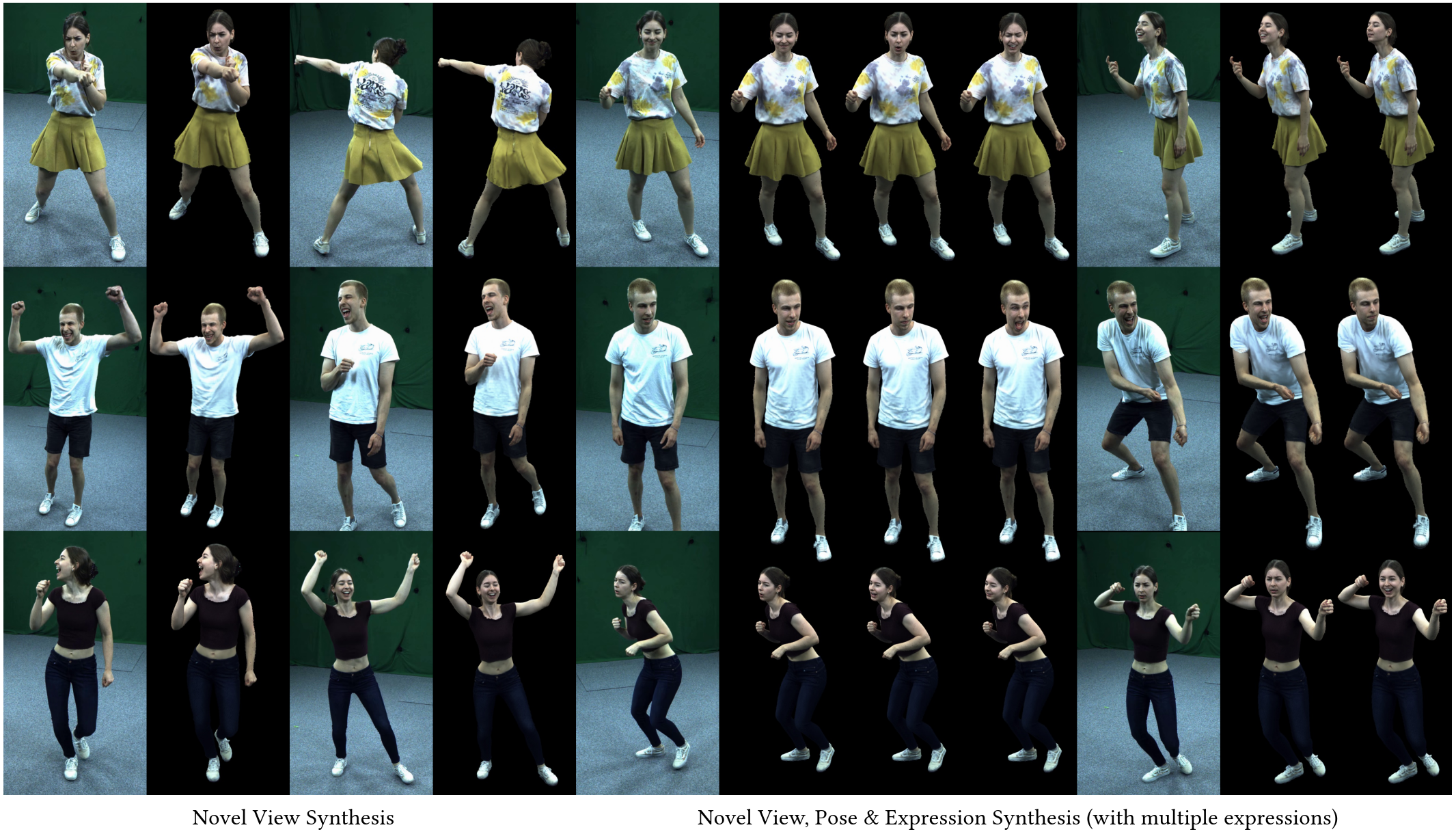}
    \caption{Qualitative results of the complete EVA model. The first set of results demonstrates EVA’s ability to render images of a character with previously observed motions and expressions from novel viewpoints. The second set showcases EVA’s performance in scenarios combining novel viewpoints, unseen motions, and new expressions. Notably, we illustrate EVA's capability to modify expressions independently of body appearance, enabled by our explicitly modeled disentanglement between body and head. This disentanglement is a key contribution of our appearance model. EVA demonstrates the capability to render high-fidelity images of humans, effectively capturing motion-dependent details while robustly handling unseen expression parameters.}
    \label{fig:qualitative_results}
\end{figure*}
%
%%%%%%%%%%%%%%%%%%%%%%%%%%%%%%%%%%%%%%%%%%%%%%%%%%%%%%%%%%%%%%%%%%%%%%%%%%%%%%%%%%%%%%%%%%%%%%
%
\paragraph{Implementation and Runtime} 
%
%%%%%%%%%%%%%%%%%%%%%%%%%%%%%%%%%%%%%%%%%%%%%%%%%%%%%%%%%%%%%%%%%%%%%%%%%%%%%%%%%%%%%%%%%%%%%%
%
The expressive template mesh is implemented in TensorFlow \citep{tensorflow2015-whitepaper} using the Adam optimizer \citep{KingBa15}. 
Given a high-resolution full-body target scan and the DDC template mesh \citep{habermann2021DDC}, the fitting process takes approximately $45$ minutes on an NVIDIA GeForce RTX $3090$, with vertex displacement optimization being the most time-intensive step due to random sampling of points on the mesh surfaces.
Our 3D Gaussian appearance model is implemented in PyTorch \citep{NEURIPS2019_9015} and trained using the Adam optimizer \citep{KingBa15} on an NVIDIA A$100$/A$40$ GPU. 
Training involves extracting $600\times800$ crops based on foreground segmentation masks, effectively enabling training on 2K resolution. 
Motion-aware face and body Gaussian texture resolutions are set to $128\times128$ and $256\times256$, respectively, with $15{,}000$ warmup iterations for each appearance model similar to ASH~\citep{Pang_2024_CVPR}.
At test time, the appearance layer of EVA achieves $\approx 35$ fps while the geometry layer runs at $\approx 41$ fps, which allows us to render 2K images in real-time on two A$40$ GPUs with one frame delay.
%
%%%%%%%%%%%%%%%%%%%%%%%%%%%%%%%%%%%%%%%%%%%%%%%%%%%%%%%%%%%%%%%%%%%%%%%%%%%%%%%%%%%%%%%%%%%%%%
%
\paragraph{Dataset and Evaluation Protocol}
%
%%%%%%%%%%%%%%%%%%%%%%%%%%%%%%%%%%%%%%%%%%%%%%%%%%%%%%%%%%%%%%%%%%%%%%%%%%%%%%%%%%%%%%%%%%%%%%
%
We leveraged a subject from \citet{shetty2024holoported} and captured multi-view videos of three subjects wearing loose or tight cloth. 
We do not evaluate our model on the DynaCap dataset \citep{habermann2021DDC}, as the actors exhibit limited facial expressions and clenched fists.
The sequences were recorded using a calibrated multi-camera system comprising $100$ full-frame cameras for whole-body capture and $20$ close-up cameras for facial details, operating at 4K resolution and $25$ fps.
The dataset includes $27{,}000$ training frames and $8{,}000$ testing frames, with each frame annotated with 3D skeletal poses \citep{thecaptury} that we further refine, FLAME parameters \citep{FLAME:SiggraphAsia2017}, foreground segmentations \citep{BGMv2, kirillov2023segany}, and 3D reconstructions \citep{neus2}.
Importantly, this dataset faithfully samples facial expressions and also records them with 20 dedicated close-up cameras, which is in contrast to prior datasets~\citep{Pang_2024_CVPR,habermann2021DDC} that mostly contain far-away views with limited resolution on the face and limited facial expressions.
We use PSNR, LPIPS \citep{zhang2018unreasonable}, and SSIM \citep{wang2004image} as quantitative evaluation metrics and compute them at 2K resolution over every $10$th frame for all four subjects. 
Testing utilized four unseen camera views excluded from the training views.
%
%%%%%%%%%%%%%%%%%%%%%%%%%%%%%%%%%%%%%%%%%%%%%%%%%%%%%%%%%%%%%%%%%%%%%%%%%%%%%%%%%%%%%%%%%%%%%%
%
\subsection{Qualitative Results}
%
%%%%%%%%%%%%%%%%%%%%%%%%%%%%%%%%%%%%%%%%%%%%%%%%%%%%%%%%%%%%%%%%%%%%%%%%%%%%%%%%%%%%%%%%%%%%%%
%
We demonstrate EVA's qualitative capabilities through expressive human template tracking and detailed renderings.
In \textbf{expressive human template tracking} (Fig.~\ref{fig:mesh_results}), our expressive template mesh and motion optimization strategy achieve superior alignment with multi-view images\textemdash crucial for accurate expression tracking. EVA’s fully drivable mesh significantly improves ASH’s proxy and animation via standard forward kinematics and dual quaternion skinning \citep{Kavan2007}.
In \textbf{novel view synthesis} (Fig.~\ref{fig:qualitative_results}, left), EVA renders high-quality facial expressions and full-body appearances\textemdash including motion-dependent details like loose clothing\textemdash in real time from unseen camera views.
For \textbf{novel view, pose \& expression synthesis} (Fig.~\ref{fig:qualitative_results}, right), EVA renders characters in unseen poses and expressions from novel viewpoints, accurately capturing the full human appearance while enabling independent control over expressions and body\textemdash addressing limitations of prior works \citep{Pang_2024_CVPR, habermann2021DDC}.
%
%%%%%%%%%%%%%%%%%%%%%%%%%%%%%%%%%%%%%%%%%%%%%%%%%%%%%%%%%%%%%%%%%%%%%%%%%%%%%%%%%%%%%%%%%%%%%%
%
\subsection{Comparison}
%
%%%%%%%%%%%%%%%%%%%%%%%%%%%%%%%%%%%%%%%%%%%%%%%%%%%%%%%%%%%%%%%%%%%%%%%%%%%%%%%%%%%%%%%%%%%%%%
%
We evaluate EVA against two real-time human rendering methods that are trained on multi-view data and rely solely on motion parameters for character animation:
(1) DDC \citep{habermann2021DDC}, a mesh-based approach where geometry is modeled using learned graph transformations and vertex displacements, while appearance is represented through learned dynamic textures; 
(2) ASH \citep{Pang_2024_CVPR}, a hybrid approach that utilizes animatable template meshes generated with DDC to model geometry and Gaussian splats to model the dynamic appearances. 
However, it lacks explicit control over facial expressions.
The code of the concurrent work DEGAS \citep{shao2024degas} is not publicly available, preventing a comparison.
Nonetheless, they require facial images at inference in order to drive facial expressions, which is in stark contrast to our method allowing to drive the face solely from facial expression parameters.
We also refer to our supplementary document for a comparison against AnimatableGaussians \citep{li2024animatablegaussians}.
%
%%%%%%%%%%%%%%%%%%%%%%%%%%%%%%%%%%%%%%%%%%%%%%%%%%%%%%%%%%%%%%%%%%%%%%%%%%%%%%%%%%%%%%%%%%%%%%
%
\par
%
%%%%%%%%%%%%%%%%%%%%%%%%%%%%%%%%%%%%%%%%%%%%%%%%%%%%%%%%%%%%%%%%%%%%%%%%%%%%%%%%%%%%%%%%%%%%%%
%
As shown in Tab.~\ref{tab:comparisons}, EVA \textbf{quantitatively} outperforms competing methods in both novel view synthesis (training sequences from unseen viewpoints) and novel view, pose, and expression synthesis (testing sequences from unseen viewpoints), except that DDC achieves a slightly higher PSNR for novel views\textemdash likely due to training on all frames rather than sampled ones.
Despite the face appearing in only a subset of evaluation images and occupying a small portion of the 2K frames, our explicit use of facial expression information, disentangled model structure, and training strategy together yield significantly better rendering quality and generalization to unseen poses and expressions.
%
%%%%%%%%%%%%%%%%%%%%%%%%%%%%%%%%%%%%%%%%%%%%%%%%%%%%%%%%%%%%%%%%%%%%%%%%%%%%%%%%%%%%%%%%%%%%%%
%
\begin{table}[t]
    \centering
    \small
    \setlength{\tabcolsep}{4pt} 
    \renewcommand{\arraystretch}{0.75} 
     \caption{Comparison results for four subjects of EVA against ASH and DDC. We highlight the \textbf{best} and \underline{second-best} results.}
    \label{tab:comparisons}
    \begin{tabular}{lcccccc}
        \toprule
        \textbf{} & \multicolumn{3}{c}{\textbf{Novel Pose/Expr.}} & \multicolumn{3}{c}{\textbf{Novel View}} \\
        \cmidrule(lr){2-4} \cmidrule(lr){5-7}
        & LPIPS $\downarrow$ & PSNR $\uparrow$ & SSIM $\uparrow$ & LPIPS $\downarrow$ & PSNR $\uparrow$ & SSIM $\uparrow$ \\
        \midrule
        ASH & 22.15 & \underline{45.78}  & \underline{98.75}  & 19.30  & 46.31  & \underline{99.09} \\
        DDC & \underline{20.70} & 45.36  & 98.62 & \underline{18.14}  & \textbf{46.87}  & 99.05 \\
        \midrule
        \textbf{EVA} & \textbf{15.55}  & \textbf{46.00}  & \textbf{98.79}  & \textbf{12.55}  & \underline{46.39}  & \textbf{99.22} \\
        \bottomrule
    \end{tabular}
\end{table}
%
%%%%%%%%%%%%%%%%%%%%%%%%%%%%%%%%%%%%%%%%%%%%%%%%%%%%%%%%%%%%%%%%%%%%%%%%%%%%%%%%%%%%%%%%%%%%%%
%
\par
%
%%%%%%%%%%%%%%%%%%%%%%%%%%%%%%%%%%%%%%%%%%%%%%%%%%%%%%%%%%%%%%%%%%%%%%%%%%%%%%%%%%%%%%%%%%%%%%
%
%%%%
\begin{figure*}[bth]
    \centering
    \includegraphics[width=\textwidth]{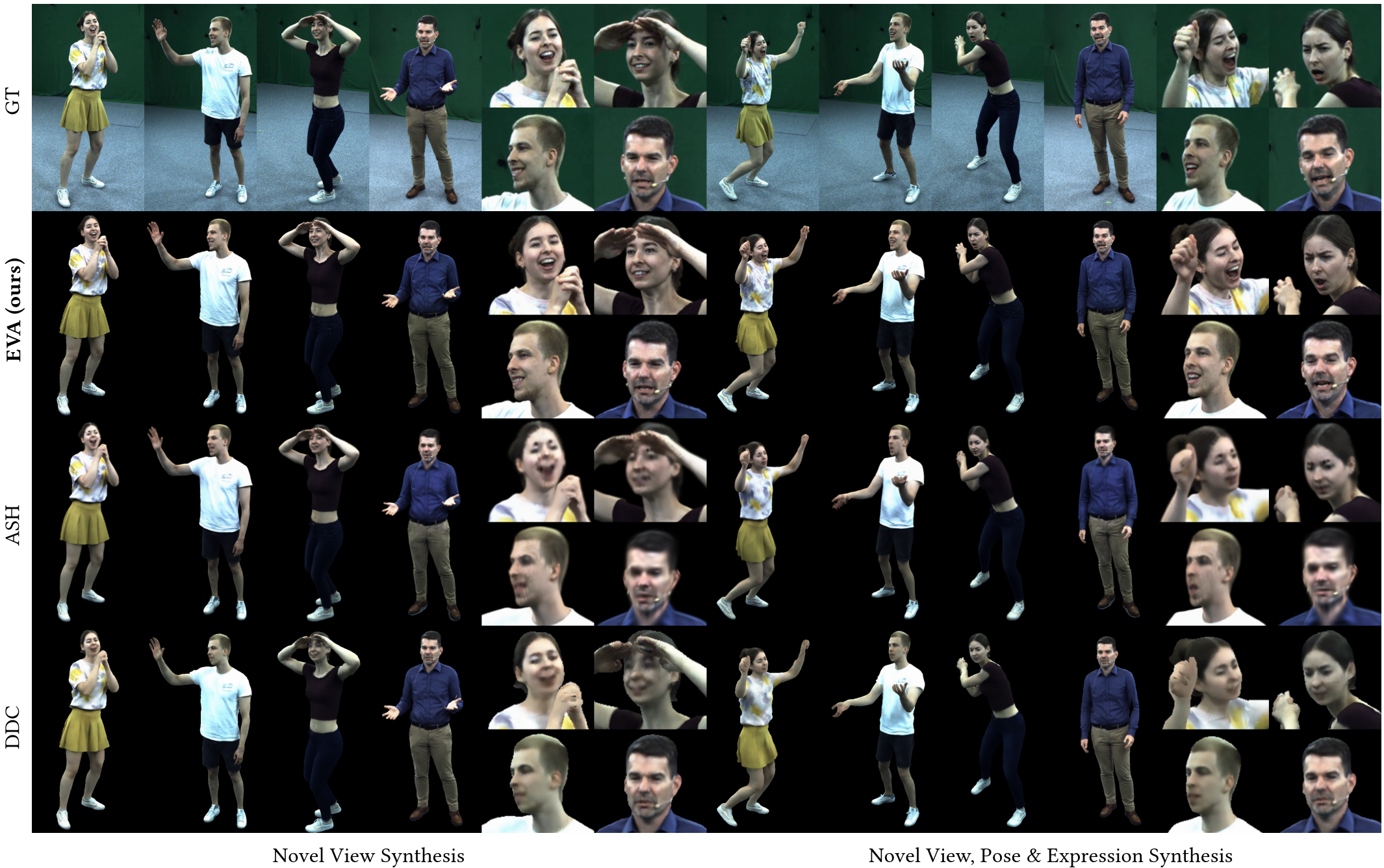}
    \caption{Qualitative comparison of EVA with two real-time human rendering approaches: ASH \citep{Pang_2024_CVPR} and DDC \citep{habermann2021DDC}. By conditioning our appearance model on the character's facial expression, EVA successfully predicts high-fidelity facial appearance in novel viewpoints, as well as for unseen motions and expressions. Additionally, our enhanced training strategy, supplementary regularization techniques, and the introduction of the IDMRF loss significantly improve both the quality of facial rendering and the overall accuracy of the generated images, outperforming ASH and DDC in terms of rendering fidelity.}
    \label{fig:quali_comparison}
\end{figure*}
%
%%%%
\begin{figure*}[bth]
    \centering
    \includegraphics[width=\textwidth]{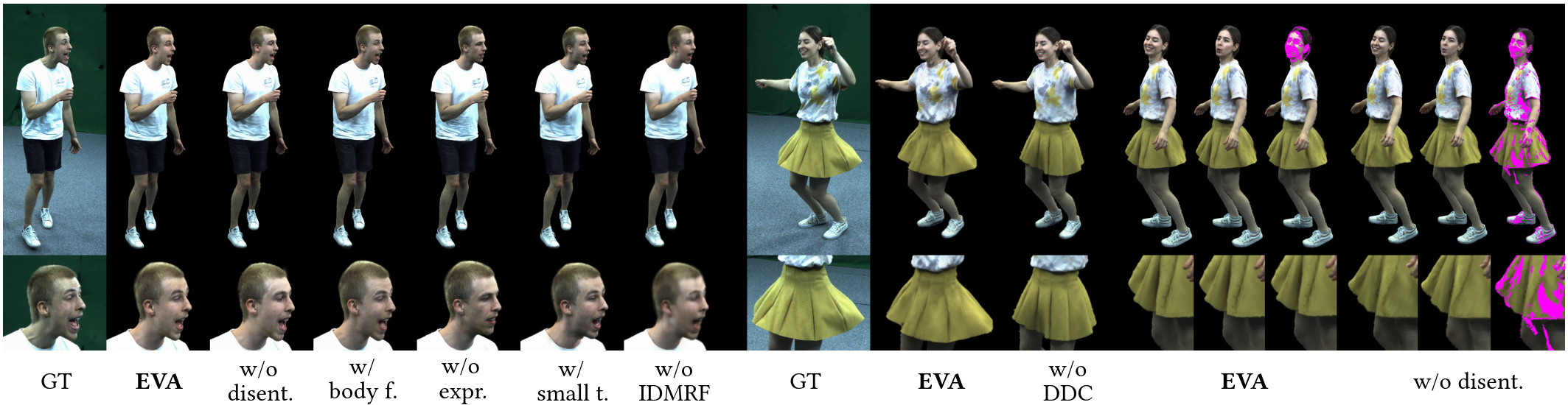}
    \caption{Qualitative ablation results highlighting the impact of our design and training strategies. The fully disentangled modeling of EVA's appearance delivers superior results compared to alternatives: without disentanglement (w/o disent.), with disentanglement but incorporating additional body features for the head (w/ body f.), without expressions (w/o expr.), with smaller texture embeddings (w/ small t.), without the IDMRF loss (w/o IDMRF), and without DDC deformations (w/o DDC). Furthermore, our disentangled modeling enables complete and independent control over the head and body, allowing changes to expression without altering body appearance (highlighted in pink) and vice versa.}
\label{fig:quali_ablations}
\end{figure*}
%%%%
%
%%%%
%
In Fig.~\ref{fig:quali_comparison}, we \textbf{qualitatively} compare EVA with ASH and DDC for novel view synthesis and novel view, pose, and expression synthesis. 
EVA captures detailed facial expressions, while ASH and DDC fail to reproduce them accurately. 
ASH infers expressions from body pose, resulting in incorrect facial appearances. 
DDC, lacking 3D Gaussians and relying on dynamic textures, produces less realistic results and misses dynamic details.
Both methods produce blurry results with artifacts due to the lack of facial expression modeling. 
Finally, ASH is susceptible to overfitting, a limitation our approach overcomes through disentanglement and effective regularization. 
%
%%%%%%%%%%%%%%%%%%%%%%%%%%%%%%%%%%%%%%%%%%%%%%%%%%%%%%%%%%%%%%%%%%%%%%%%%%%%%%%%%%%%%%%%%%%%%%
%
\subsection{Ablations}
%
%%%%%%%%%%%%%%%%%%%%%%%%%%%%%%%%%%%%%%%%%%%%%%%%%%%%%%%%%%%%%%%%%%%%%%%%%%%%%%%%%%%%%%%%%%%%%%
%
We conduct several ablation studies to evaluate the contributions of our key design choices and training components. 
The ablation setups include:
(1) a model without the IDMRF loss;
(2) a model without the motion-dependent DDC deformations that is solely driven by a skinned template mesh;
(3) a version with smaller texture embedding sizes ($64 \times 64$ for the face and $128 \times 128$ for the body);
(4) a model without explicit disentanglement of head and body;
(5) a variant without expression parameters as input; 
(6) a variant with additional body features as inputs for head appearance prediction; Specifically, we additionally condition the head Gaussian prediction on the motion-aware textures of the body;
and (7) the version without separate geometry and appearance U-Nets.
%
%%%%%%%%%%%%%%%%%%%%%%%%%%%%%%%%%%%%%%%%%%%%%%%%%%%%%%%%%%%%%%%%%%%%%%%%%%%%%%%%%%%%%%%%%%%%%%
%
\par
%
%%%%%%%%%%%%%%%%%%%%%%%%%%%%%%%%%%%%%%%%%%%%%%%%%%%%%%%%%%%%%%%%%%%%%%%%%%%%%%%%%%%%%%%%%%%%%%
%
Tab.~\ref{tab:ablations} presents a \textbf{quantitative} comparison of these ablation setups using LPIPS and PSNR metrics for both novel view synthesis and novel view, pose, and expression synthesis. 
EVA outperforms all ablated variants. 
Notably, the variants without disentanglement and those with additional body features show a decline in quantitative performance, as the network often learns incorrect correlations between facial expressions and body motion instead of the correct disentangled behavior.
Furthermore, the absence of the IDMRF leads to a noticeable degradation in perceptual quality.
%
%%%%%%%%%%%%%%%%%%%%%%%%%%%%%%%%%%%%%%%%%%%%%%%%%%%%%%%%%%%%%%%%%%%%%%%%%%%%%%%%%%%%%%%%%%%%%%
%
\begin{table}[t]
    \centering
    \small
    \setlength{\tabcolsep}{5pt} 
    \renewcommand{\arraystretch}{0.75} 
     \caption{Ablation study results for three subjects. We highlight the \textbf{best} and \underline{second-best} results. The table is sorted by LPIPS in descending order.}
    \label{tab:ablations}
    \begin{tabular}{lcccc}
        \toprule
        \textbf{Ablation Setup} & \multicolumn{2}{c}{\textbf{Novel Pose/Expr.}} & \multicolumn{2}{c}{\textbf{Novel View}} \\
        \cmidrule(lr){2-3} \cmidrule(lr){4-5}
        & \underline{LPIPS} $\downarrow$ & PSNR $\uparrow$ & \underline{LPIPS} $\downarrow$ & PSNR $\uparrow$ \\
        \midrule
        w/o IDMRF loss      & 19.27  & 45.02  & 15.27  & 45.57  \\
        w/o DDC deformations & 17.31  & 44.52  & 13.95  & 45.01  \\
        w/ small textures   & 17.07  & 45.02  & 13.27  & 45.46  \\
        w/o disentanglement & 16.74  & 44.92  & 12.20  & 45.55  \\
        w/o expressions     & 16.27  & \underline{45.05}  & \underline{11.73}  & 45.59  \\
        w/ body features    & 16.07  & 45.02  & \underline{11.73}  & \underline{45.61}  \\
        w/o geo/app netw.   & \underline{15.73}  & \underline{45.05}  & \textbf{11.67}  & 45.55  \\
         \midrule
        \textbf{EVA} & \textbf{15.27}  & \textbf{45.21}  & \textbf{11.67}  & \textbf{45.73}  \\
        \bottomrule
    \end{tabular}
\end{table}
%
%%%%%%%%%%%%%%%%%%%%%%%%%%%%%%%%%%%%%%%%%%%%%%%%%%%%%%%%%%%%%%%%%%%%%%%%%%%%%%%%%%%%%%%%%%%%%%
%
\par
%
%%%%%%%%%%%%%%%%%%%%%%%%%%%%%%%%%%%%%%%%%%%%%%%%%%%%%%%%%%%%%%%%%%%%%%%%%%%%%%%%%%%%%%%%%%%%%%
%
Fig.~\ref{fig:quali_ablations} illustrates \textbf{qualitative} results from the ablation studies. 
The full EVA model produces higher-quality results than variants lacking disentanglement or incorporating extra body features for head appearance prediction. 
Additionally, our model’s ability to independently alter facial expressions and body pose without affecting the other highlights the expressive power of our approach. 
In contrast, the ablation setup without facial expression inputs fails to reconstruct facial appearance accurately, and the variants with smaller texture embeddings or without the IDMRF loss produce noticeably blurrier results, further emphasizing the importance of these design choices. 
Importantly, a variant without the learned DDC deformations is unable to reconstruct loose clothing.
%
%
%%%%%%%%%%%%%%%%%%%%%%%%%%%%%%%%%%%%%%%%%%%%%%%%%%%%%%%%%%%%%%%%%%%%%%%%%%%%%%%%%%%%%%%%%%%%%%
%
\section{Limitations.}
%
%%%%%%%%%%%%%%%%%%%%%%%%%%%%%%%%%%%%%%%%%%%%%%%%%%%%%%%%%%%%%%%%%%%%%%%%%%%%%%%%%%%%%%%%%%%%%%
%
EVA marks a major step forward in photorealistic rendering of expressive humans, but several limitations remain.
First, its mesh-based representation prevents handling topological changes. Future work could explore layered representations, such as learnable garment layers, for greater flexibility.
EVA also fixes the number of Gaussians describing the body, limiting multi-scale rendering. A hierarchical Gaussian model, like that of \citep{kerbl2024hierarchical}, could offer both global structure and fine detail.
The use of separate Gaussian textures for the head and body can cause minor visual inconsistencies at the neck. A consistency loss across this boundary may help. 
Furthermore, the current independent predictions prevent modeling cross-region lighting or shadows (e.g., a hand shadow on the face).
Lighting is modeled via a latent representation, constrained by the studio's lighting conditions during training. Future work may benefit from combining physically based rendering with expressive full-body avatars.
%
%%%%%%%%%%%%%%%%%%%%%%%%%%%%%%%%%%%%%%%%%%%%%%%%%%%%%%%%%%%%%%%%%%%%%%%%%%%%%%%%%%%%%%%%%%%%%%
%
%
%%%%%%%%%%%%%%%%%%%%%%%%%%%%%%%%%%%%%%%%%%%%%%%%%%%%%%%%%%%%%%%%%%%%%%%%%%%%%%%%%%%%%%%%%%%%%%
%
\section{Conclusion} \label{sec:conclusion}
%
%%%%%%%%%%%%%%%%%%%%%%%%%%%%%%%%%%%%%%%%%%%%%%%%%%%%%%%%%%%%%%%%%%%%%%%%%%%%%%%%%%%%%%%%%%%%%%
%
We introduced Expressive Virtual Avatars (EVA), a novel method for real-time, high-quality, and expressive human avatar rendering.
EVA excels in achieving superior realism while providing full control over the avatar's expression and skeletal motion.
At its core, EVA leverages an expressive human template that integrates face, hand, and body control, combined with 3D Gaussians distributed across the avatar's surface. 
To ensure precise and separate control over facial expressions and body poses, we introduce an explicit disentanglement of the 3D Gaussian predictions of the body and the head. 
We believe that EVA establishes a robust foundation for future research and innovation in XR environments, virtual production, and social telepresence applications.
%
%%%%%%%%%%%%%%%%%%%%%%%%%%%%%%%%%%%%%%%%%%%%%%%%%%%%%%%%%%%%%%%%%%%%%%%%%%%%%%%%%%%%%%%%%%%%%%
%
%%%%%%%%%%%%%%%%%%%%%%%%%%%%%%%%%%%%%%%%%%%%%%%%%%%%%%%%%%%%%%%%%%%%%%%%%%%%%%%%%%%%%%%%%%%%%%
%
\begin{acks}
This project was supported by the Saarbr\"ucken Research Center for Visual Computing, Interaction, and AI. We would like to sincerely thank Pranay Raj Kamuni and Janine Haberkorn for their invaluable help during the data capture in the multi-camera studio.
\end{acks}
%
%%%%%%%%%%%%%%%%%%%%%%%%%%%%%%%%%%%%%%%%%%%%%%%%%%%%%%%%%%%%%%%%%%%%%%%%%%%%%%%%%%%%%%%%%%%%%%
\bibliographystyle{ACM-Reference-Format}
\bibliography{reference}
%%%%%%%%%%%%%%%%%%%%%%%%%%%%%%%%%%%%%%%%%%%%%%%%%%%%%%%%%%%%%%%%%%%%%%%%%%%%%%%%%%%%%%%%%%%%%%
\clearpage
%%%%%%%%%%%%%%%%%%%%%%%%%%%%%%%%%%%%%%%%%%%%%%%%%%%%%%%%%%%%%%%%%%%%%%%%%%%%%%%%%%%%%%%%%%%%%%
%
\appendix
\section{Overview}
%
%%%%%%%%%%%%%%%%%%%%%%%%%%%%%%%%%%%%%%%%%%%%%%%%%%%%%%%%%%%%%%%%%%%%%%%%%%%%%%%%%%%%%%%%%%%%%%
%
In the supplementary document, we provide detailed explanations of the implementation of our competing methods, DDC~\citep{habermann2021DDC} (Sec.~\ref{sec:ddc_impl}) and ASH \citep{Pang_2024_CVPR} (Sec.~\ref{sec:ash_impl}). 
We then delve into the technical aspects of our own method, EVA (Sec.~\ref{sec:eva_impl}), with a focus on three key components: the construction of the expressive deformable template $\boldsymbol{\Phi}_\mathrm{mesh}$, the pipelines for motion optimization and expression tracking, and the disentangled 3D Gaussian appearance model $\boldsymbol{\Phi}_\mathrm{app}$. 
Following this, we describe the head stitching operator $(\bowtie)$, which seamlessly integrates the personalized head avatar with the deformable character model (Sec.~\ref{sec:supp_head_stitching}).
We further elaborate on our eye blink postprocessing strategy that enhances the accuracy of facial expression tracking (Sec.~\ref{sec:supp_eyeblink}). 
A comprehensive runtime analysis is also included, detailing the performance of individual components of our system (Sec.~\ref{sec:supp_runtime}).
Additionally, we provide information about the dataset used for training and evaluation (Sec.~\ref{sec:dataset_supp}), followed by more qualitative results (Sec.~\ref{sec:results_supp}). 
These include personalized head avatars for all four subjects, insights into the effects of the learned deformable character for loose clothing, and demonstrations of an audio-driven avatar. 
We also present a comparison with the recent human avatar approach AnimatableGaussians~\citep{li2024animatablegaussians} (Sec.~\ref{sec:ag_comp}). 
To further support our findings, we include several additional ablation studies (Sec.~\ref{sec:ablation_supp}), such as the impact of random background augmentation, visualizations highlighting the role of our global position latent, and a variant of our method using constant head Gaussians, resembling a reimplementation of \citet{qian2023gaussianavatars}. 
Finally, we address potential ethical considerations (Sec.~\ref{sec:ethical}).
%
%%%%%%%%%%%%%%%%%%%%%%%%%%%%%%%%%%%%%%%%%%%%%%%%%%%%%%%%%%%%%%%%%%%%%%%%%%%%%%%%%%%%%%%%%%%%%%
%
\section{DDC - Technical Details}\label{sec:ddc_impl}
\begin{figure}[t]
    \centering
    \subfloat[3D Scan]
    {
        \includegraphics[width=.32\linewidth]{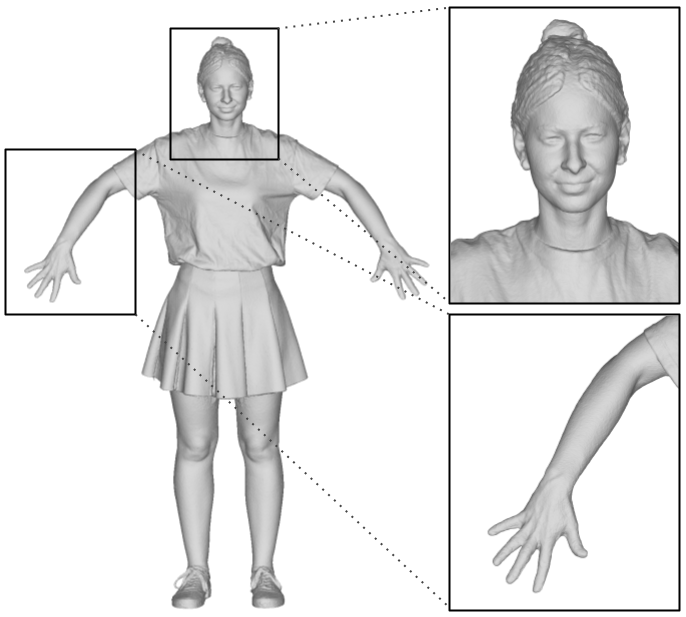}
    } 
    \hspace*{-0.7em}
    \subfloat[Uniform]
    {
        \includegraphics[width=.32\linewidth]{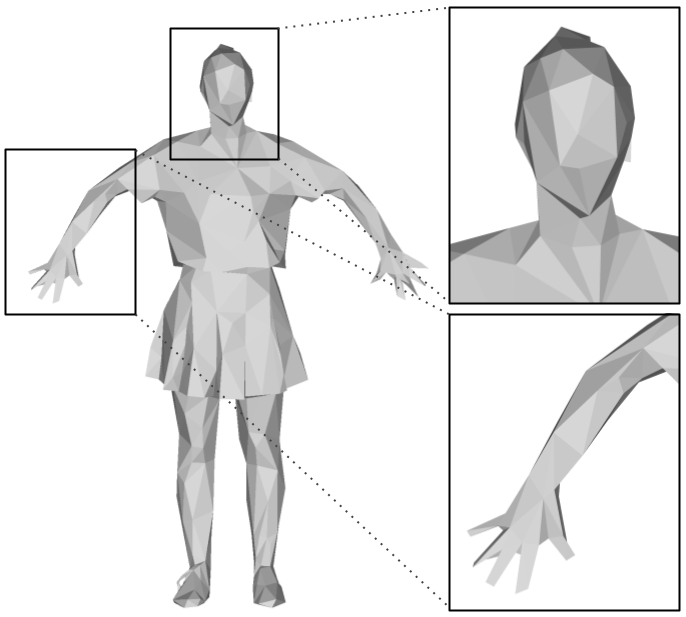}
    }
    \hspace*{-0.7em}
    \subfloat[\textbf{Adaptive (ours)}]
    {
        \includegraphics[width=.32\linewidth]{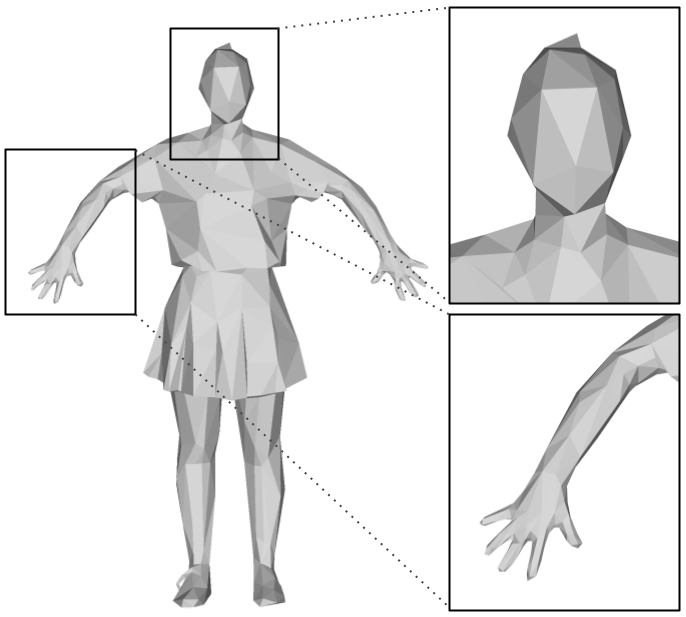}
    }
    \vspace{-8pt}
    \caption{Comparison between the high-resolution scan, the uniformly downsampled embedded graph, and the adaptively simplified embedded graph, highlighting their differences.}
    \label{fig:meshescompare}
\end{figure}

%
%%%%%%%%%%%%%%%%%%%%%%%%%%%%%%%%%%%%%%%%%%%%%%%%%%%%%%%%%%%%%%%%%%%%%%%%%%%%%%%%%%%%%%%%%%%%%%
%
We follow the implementation of \citet{habermann2021DDC} to train DDC, utilizing TensorFlow \citep{tensorflow2015-whitepaper} and the Adam optimizer \citep{KingBa15}. 
The training process involves four stages: (1) training the embedded graph module $\phi_\mathrm{eg}$, (2) optimizing global lighting, (3) training the per-vertex displacement module $\phi_\mathrm{delta}$, and (4) training the network for dynamic texture prediction. 
Detailed descriptions of the loss functions and training procedures can be found in \citet{habermann2021DDC} and \citep{shetty2024holoported}. All stages are trained on four concurrently utilized NVIDIA A$100$/A$40$ GPUs. 
Unlike the original DDC and ASH \citep{Pang_2024_CVPR} implementation, we incorporate our optimized skeletal pose parameters $\boldsymbol{\theta}$ to improve training.
The following paragraphs will cover technical details of the four DDC training stages.
%
%%%%%%%%%%%%%%%%%%%%%%%%%%%%%%%%%%%%%%%%%%%%%%%%%%%%%%%%%%%%%%%%%%%%%%%%%%%%%%%%%%%%%%%%%%%%%%
%
\paragraph{Embedded Graph Layer}
%
%%%%%%%%%%%%%%%%%%%%%%%%%%%%%%%%%%%%%%%%%%%%%%%%%%%%%%%%%%%%%%%%%%%%%%%%%%%%%%%%%%%%%%%%%%%%%%
%
A learned embedded graph~\citep{Sumner2007} layer $\phi_\mathrm{eg}$ is adopted to model coarse, motion-aware surface deformations within the canonical pose space
%
%%%%%%%%%%%%%%%%%%%%%%%%%%%%%%%%%%%%%%%%%%%%%%%%%%%%%%%%%%%%%%%%%%%%%%%%%%%%%%%%%%%%%%%%%%%%%%
%
\begin{equation} 
\label{eq:ddc_deform}
\phi_\mathrm{eg}(\hat{\boldsymbol{\theta}}^\triangleright_t,\mathbf{V}_\mathrm{T}) = \mathbf{V}_{C,i}=\mathbf{D}_i +\sum_{j \in \mathcal{A}_{\mathrm{n},i}}w_{i,j}(\mathbf{R}_j(\mathbf{V}_{\mathrm{T},i}-\mathbf{V}_{\mathrm{G},j}) + \mathbf{V}_{\mathrm{G},j}+T_{j}),
\end{equation}
%
%%%%%%%%%%%%%%%%%%%%%%%%%%%%%%%%%%%%%%%%%%%%%%%%%%%%%%%%%%%%%%%%%%%%%%%%%%%%%%%%%%%%%%%%%%%%%%
%
where $\mathbf{V}_{T,i}$ represents the vertices of the template mesh, $\mathbf{V}_{\mathrm{G},j}$ denotes the embedded graph nodes, and $\mathcal{A}_{\mathrm{n},i}$ specifies the set of embedded graph nodes corresponding to the template mesh vertices $\mathbf{V}_{T,i}$. 
The embedded graph nodes $\mathbf{V}_{\mathrm{G},j}$ are generated by simplifying the canonical template mesh vertices $\mathbf{V}_{T,i}$ using quadric edge collapse decimation~\cite{Garland1998}.
The correspondence between the embedded graph nodes $\mathbf{V}_{\mathrm{G},j}$ and the template mesh vertices $\mathbf{V}_{\mathrm{T},i}$ is established following the method proposed in ~\citep{Sumner2007}. 
The transformations of the embedded graph nodes, namely rotations $\mathbf{R}_j$ and translations $\mathbf{T}_{j}$, are predicted by a graph neural network (GNN) trained on full-frame multi-view images resized to $643 \times 470$ pixels.
Training is conducted with a batch size of $32$, an initial learning rate of $0.0001$, and for $360{,}000$ iterations. 
We employ a silhouette loss $\mathcal{L}_\mathrm{sil}$ with a weight of $\alpha_\mathrm{sil} = 100.0$, and an as-rigid-as-possible regularization $\mathcal{L}_\mathrm{arap}$ with a weight of $\alpha_\mathrm{arap} = 1500.0$ \citep{ARAP_modeling:2007,Sumner2007}. 
Drawing from ideas of \citet{shetty2024holoported}, additional geometric supervision from per-frame surface reconstructions \citep{neus2} is provided via a bidirectional Chamfer loss $\mathcal{L}_\mathrm{Chamf}^\rightleftharpoons$ weighted at $\alpha_\mathrm{Chamf} = 1500.0$.
In contrast to \citet{habermann2021DDC,Pang_2024_CVPR,Habermann2023}, who uniformly simplify template meshes to generate the embedded graph, we perform adaptive downsampling. Uniform downsampling often misallocates vertices, resulting in oversampling static regions while undersampling highly articulated areas such as joints. Our adaptive method simplifies the mesh based on manually annotated body parts, preserving vertex density in regions like the hands, armpits, shoulders, and knees, while more aggressively simplifying less dynamic areas such as the head, chest, and legs (see Fig.~\ref{fig:meshescompare}). Additionally, hand deformations are discarded after training.
%
%%%%%%%%%%%%%%%%%%%%%%%%%%%%%%%%%%%%%%%%%%%%%%%%%%%%%%%%%%%%%%%%%%%%%%%%%%%%%%%%%%%%%%%%%%%%%%
%
%
%%%%%%%%%%%%%%%%%%%%%%%%%%%%%%%%%%%%%%%%%%%%%%%%%%%%%%%%%%%%%%%%%%%%%%%%%%%%%%%%%%%%%%%%%%%%%%
%
\paragraph{Lighting Optimization}
%
%%%%%%%%%%%%%%%%%%%%%%%%%%%%%%%%%%%%%%%%%%%%%%%%%%%%%%%%%%%%%%%%%%%%%%%%%%%%%%%%%%%%%%%%%%%%%%
%
Global lighting within the capture studio is optimized using a rendering loss $\mathcal{L}_\mathrm{ren}$ on full-frame images resized to $1285\times 940$ pixels. Training is conducted with a batch size of $32$, an initial learning rate of $0.01$, and for $8{,}000$ iterations. This stage builds upon the embedded graph deformations and optimizes the global lighting coefficients to match the studio lighting. We refer to \citet{habermann2021DDC} for more details.
%
%%%%%%%%%%%%%%%%%%%%%%%%%%%%%%%%%%%%%%%%%%%%%%%%%%%%%%%%%%%%%%%%%%%%%%%%%%%%%%%%%%%%%%%%%%%%%%
%
\paragraph{Vertex Displacement Layer}
%
%%%%%%%%%%%%%%%%%%%%%%%%%%%%%%%%%%%%%%%%%%%%%%%%%%%%%%%%%%%%%%%%%%%%%%%%%%%%%%%%%%%%%%%%%%%%%%
%
The per-vertex displacements $\mathbf{D}_i$ in Eq.~\ref{eq:ddc_deform} are predicted by a GNN trained using full-frame images resized to $643\times 470$ pixels and their corresponding surface reconstructions \citep{neus2,shetty2024holoported}. We train with a batch size of $16$, a learning rate of $0.0001$, and for $360{,}000$ iterations. Loss functions include a silhouette loss $\mathcal{L}_\mathrm{sil}$ weighted at  $\alpha_\mathrm{sil} = 50.0$, a rendering loss $\mathcal{L}_\mathrm{ren}$ with $\alpha_\mathrm{ren} = 0.001$, Laplacian regularization  $\mathcal{L}_\mathrm{lap}$ weighted at $\alpha_\mathrm{lap} = 4000.0$, and bidirectional Chamfer loss $\mathcal{L}_\mathrm{Chamf}^\rightleftharpoons$ weighted at $\alpha_\mathrm{Chamf} = 5000.0$. Similar to the embedded graph layer, we exclude hand displacements after training.
%
%%%%%%%%%%%%%%%%%%%%%%%%%%%%%%%%%%%%%%%%%%%%%%%%%%%%%%%%%%%%%%%%%%%%%%%%%%%%%%%%%%%%%%%%%%%%%%
%
\paragraph{Dynamic Texture Layer}
%
%%%%%%%%%%%%%%%%%%%%%%%%%%%%%%%%%%%%%%%%%%%%%%%%%%%%%%%%%%%%%%%%%%%%%%%%%%%%%%%%%%%%%%%%%%%%%%
%
The texture prediction network is trained exclusively with a rendering loss $\mathcal{L}_\mathrm{ren}$ weighted at $\alpha_\mathrm{ren} = 1.0$. Full-frame images are resized to $1285\times 940$ pixels, and training is conducted with a batch size of $16$, an initial learning rate of $0.0001$, and for $360{,}000$ iterations. The resolution of the dynamic texture is set to $1024\times 1024$. To ensure unbiased evaluation, four camera views used for testing are excluded from the training process.
%
%%%%%%%%%%%%%%%%%%%%%%%%%%%%%%%%%%%%%%%%%%%%%%%%%%%%%%%%%%%%%%%%%%%%%%%%%%%%%%%%%%%%%%%%%%%%%%
%
\section{ASH - Technical Details}\label{sec:ash_impl}
%
%%%%%%%%%%%%%%%%%%%%%%%%%%%%%%%%%%%%%%%%%%%%%%%%%%%%%%%%%%%%%%%%%%%%%%%%%%%%%%%%%%%%%%%%%%%%%%
%
To train ASH, we follow the implementation described by \citet{Pang_2024_CVPR}. Training is conducted on a single NVIDIA A$100$/A$40$ GPU using PyTorch \citep{NEURIPS2019_9015} and the Adam optimizer \citep{KingBa15}. ASH utilizes the learned deformable character described above (see Sec.~\ref{sec:ddc_impl}) as its template. Training is performed on every $10$th full-frame image from the multi-view training videos, with the images downsized to a resolution of $1285\times 940$ pixels. The motion-aware textures are set to a resolution of $256\times256$.
ASH is trained in batches of size $1$, with an initial learning rate of $0.0001$, and optimized for $4{,}000{,}000$ iterations. Four cameras are excluded from the training process to reserve them for evaluation. Unlike the original ASH implementation, we utilize our optimized skeletal pose parameters $\boldsymbol{\theta}$ during training.
For supervision, we apply a pixel-wise L$1$ loss $\mathcal{L}_1$ with a weight of $\alpha_1 = 0.9$ and a structural similarity index measure (SSIM) loss $\mathcal{L}_\mathrm{ssim}$ with a weight of $\alpha_\mathrm{ssim} = 0.1$. Detailed definitions of the loss functions can be found in \citet{Pang_2024_CVPR}. Additionally, the training process includes $15{,}000$ warmup iterations to initialize the motion-aware texture decoders before the main optimization begins.
%
%%%%%%%%%%%%%%%%%%%%%%%%%%%%%%%%%%%%%%%%%%%%%%%%%%%%%%%%%%%%%%%%%%%%%%%%%%%%%%%%%%%%%%%%%%%%%%
%
\section{EVA - Technical Details}\label{sec:eva_impl}
%
%%%%%%%%%%%%%%%%%%%%%%%%%%%%%%%%%%%%%%%%%%%%%%%%%%%%%%%%%%%%%%%%%%%%%%%%%%%%%%%%%%%%%%%%%%%%%%
%
EVA comprises several components, and we provide detailed technical insights into its individual layers and pipelines below, namely: (1) the expressive deformable template, (2) the motion optimization and expression tracking pipelines, and (3) the disentangled 3D Gaussian appearance layer.
%
%%%%%%%%%%%%%%%%%%%%%%%%%%%%%%%%%%%%%%%%%%%%%%%%%%%%%%%%%%%%%%%%%%%%%%%%%%%%%%%%%%%%%%%%%%%%%%
%
\subsection{Expressive Deformable Template}
%
%%%%%%%%%%%%%%%%%%%%%%%%%%%%%%%%%%%%%%%%%%%%%%%%%%%%%%%%%%%%%%%%%%%%%%%%%%%%%%%%%%%%%%%%%%%%%%
%
The expressive human template $\boldsymbol{\Phi}_\mathrm{mesh}$ is constructed by training a learned character deformation model $C$ and fitting a personalized head avatar $H$. Finally, we utilize the stitching operator $\bowtie$ (detailed in Sec.~\ref{sec:supp_head_stitching}) to seamlessly combine the character model $C$ with the personalized head avatar $H$.
For $C$, we follow the implementation outlined in Sec.~\ref{sec:ddc_impl}.
%
%%%%%%%%%%%%%%%%%%%%%%%%%%%%%%%%%%%%%%%%%%%%%%%%%%%%%%%%%%%%%%%%%%%%%%%%%%%%%%%%%%%%%%%%%%%%%%
%
\par
%
%%%%%%%%%%%%%%%%%%%%%%%%%%%%%%%%%%%%%%%%%%%%%%%%%%%%%%%%%%%%%%%%%%%%%%%%%%%%%%%%%%%%%%%%%%%%%%
%
The personalized head avatar fitting process follows the multi-stage optimization strategy described in the main paper. We developed a standalone, fully automated pipeline for creating the head avatar. This pipeline takes a target scan of the actor with manually annotated 3D landmarks as input and generates the personalized head avatar as output. Implemented in TensorFlow \citep{tensorflow2015-whitepaper}, the pipeline employs a gradient descent approach using the Adam optimizer \citep{KingBa15}. When executed on a single NVIDIA GeForce RTX $3090$ GPU, the entire optimization process for $H$ takes approximately $45$ minutes. The following sections provide details on each fitting stage.
%
%%%%%%%%%%%%%%%%%%%%%%%%%%%%%%%%%%%%%%%%%%%%%%%%%%%%%%%%%%%%%%%%%%%%%%%%%%%%%%%%%%%%%%%%%%%%%%
%
\paragraph{Coarse Alignment} 
%
%%%%%%%%%%%%%%%%%%%%%%%%%%%%%%%%%%%%%%%%%%%%%%%%%%%%%%%%%%%%%%%%%%%%%%%%%%%%%%%%%%%%%%%%%%%%%%
%
The coarse alignment stage employs a 3D landmark loss $\mathcal{L}_\mathrm{Lmk3D}$ to align the FLAME mesh \citep{FLAME:SiggraphAsia2017} with the target scan. The loss function is defined as:
%
%%%%%%%%%%%%%%%%%%%%%%%%%%%%%%%%%%%%%%%%%%%%%%%%%%%%%%%%%%%%%%%%%%%%%%%%%%%%%%%%%%%%%%%%%%%%%%
%
\begin{equation}\label{eqn:lmk3d}
    \mathcal{L}_\mathrm{Lmk3D}(\mathbf{V}_H, \mathbf{P}_\mathrm{lmk}) = \sqrt{\sum_{i=1}^{N_\mathrm{lmk}} \lvert\lvert g_i(\mathbf{V}_H)-\mathbf{p}_i\rvert\rvert_2^2}
\end{equation}
%
%%%%%%%%%%%%%%%%%%%%%%%%%%%%%%%%%%%%%%%%%%%%%%%%%%%%%%%%%%%%%%%%%%%%%%%%%%%%%%%%%%%%%%%%%%%%%%
%
where $\mathbf{V}_H$ represents the FLAME vertices,  $\mathbf{P}_\mathrm{lmk}$ denotes the $N_\mathrm{lmk}$ 3D landmark positions on the target mesh, and $g(\cdot)$ maps FLAME vertices to 3D landmarks using FLAME's landmark embeddings.
Specifically, we use $51$ manually annotated 3D landmarks on the target scan, following the $300$-W facial landmark definition \citep{Sagonas2013a, Sagonas2013b, SAGONAS20163}, and corresponding landmarks on the FLAME mesh derived from FLAME's landmark embeddings.
We initialize the learning rate at $0.02$ and optimize global translation $\boldsymbol{\theta}_\mathrm{trans}$ and rotation $\boldsymbol{\theta}_\mathrm{rot}$ for $3{,}000$ iterations. An early stopping mechanism halts optimization if no significant improvement is observed.
%
%%%%%%%%%%%%%%%%%%%%%%%%%%%%%%%%%%%%%%%%%%%%%%%%%%%%%%%%%%%%%%%%%%%%%%%%%%%%%%%%%%%%%%%%%%%%%%
%
\begin{figure}[t]
    \centering
    \subfloat
    {
        \includegraphics[trim={0cm 0cm 0cm 0cm},clip,width=.06\linewidth]{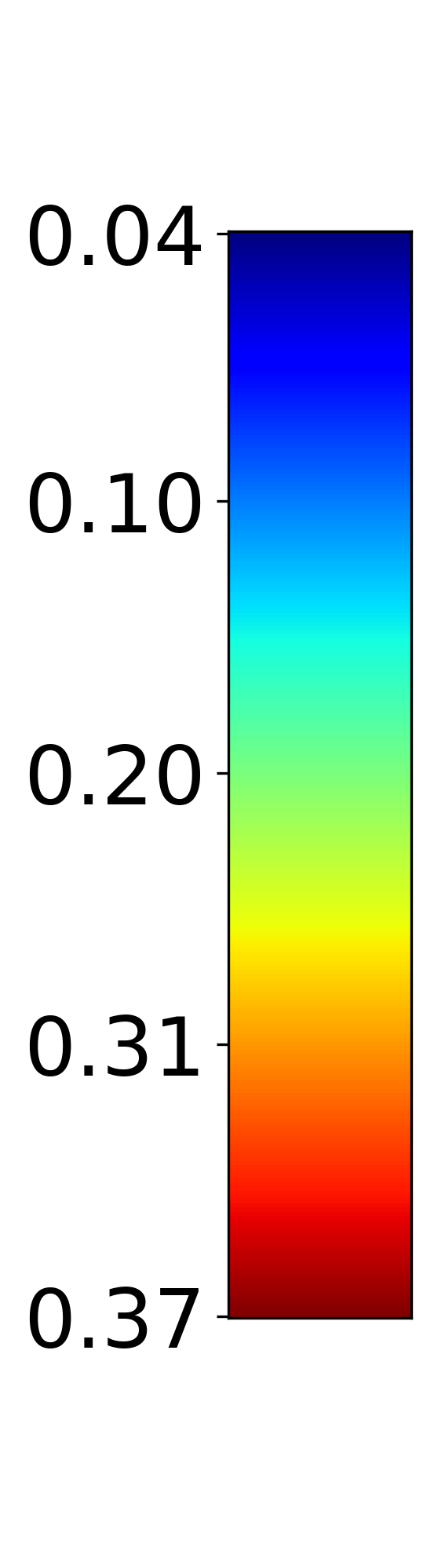}
    }
    \subfloat
    {
        \includegraphics[trim={5cm 0.833cm 6.333cm 0.666cm},clip,width=.17\linewidth]{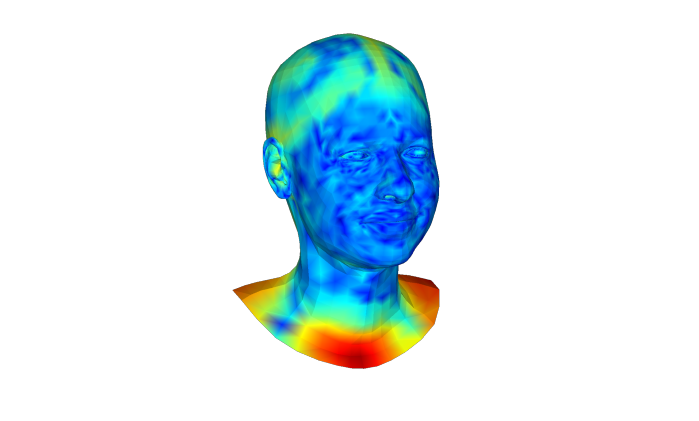}
    } 
    \subfloat
    {
        \includegraphics[trim={5.333cm 0.666cm 5.333cm 0.333cm},clip,width=.17\linewidth]{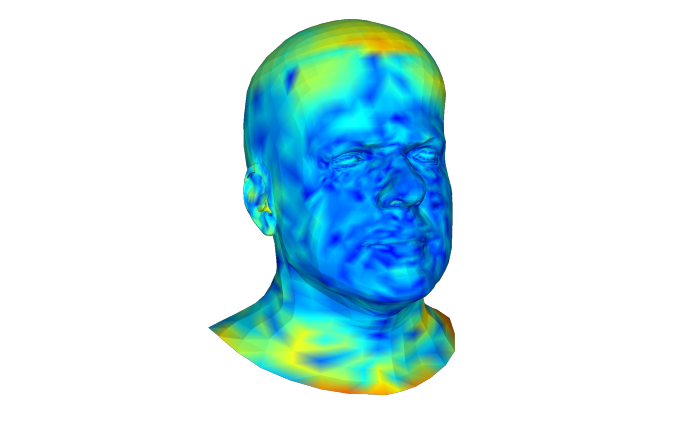}
    } 
    \subfloat
    {
        \includegraphics[trim={5cm 0.833cm 6.333cm 0.666cm},clip,width=.17\linewidth]{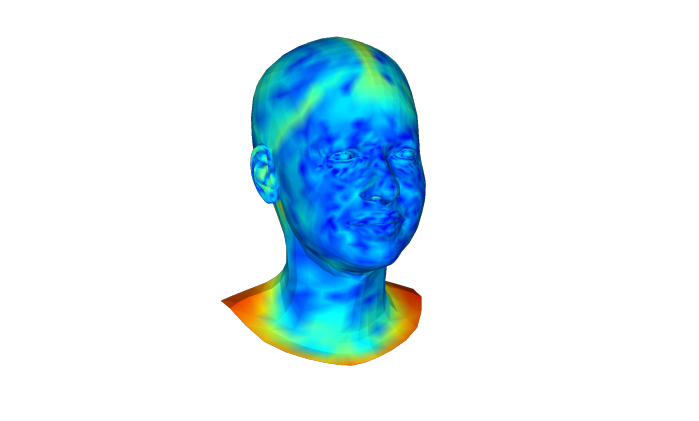}
    } 
    \subfloat
    {
        \includegraphics[trim={5.333cm 0.666cm 5.333cm 0.333cm},clip,width=.17\linewidth]{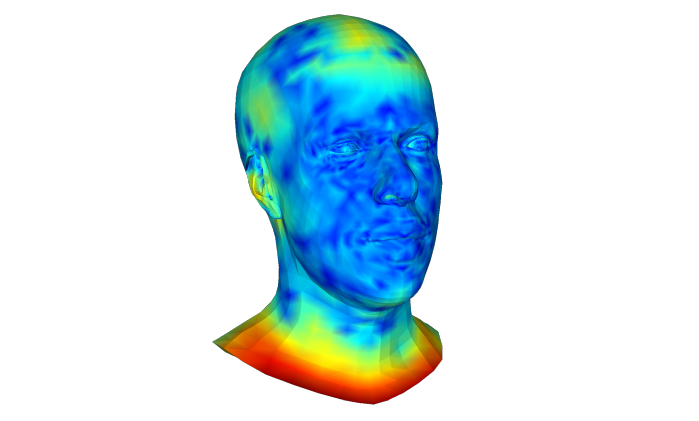}
    }
    \\
    \vspace*{-0.92em}
    \subfloat
    {
        \includegraphics[trim={0cm 0cm 0cm 0cm},clip,width=.06\linewidth]{fig_supp/color_legend_25_new.png}
    }
    \subfloat
    {
        \includegraphics[trim={5cm 0.833cm 6.333cm 0.666cm},clip,width=.17\linewidth]{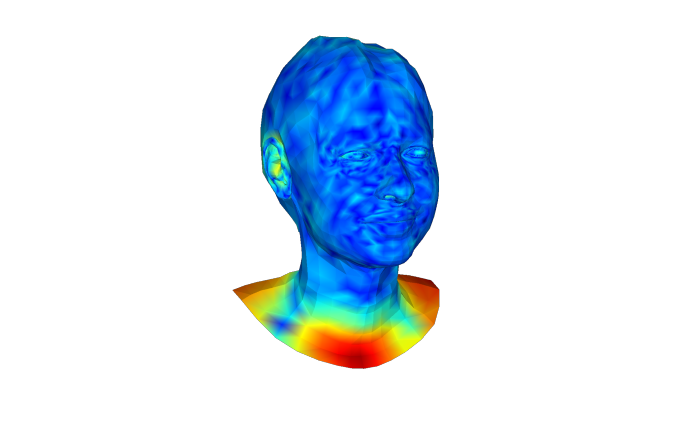}
    } 
    \subfloat
    {
        \includegraphics[trim={5.333cm 0.666cm 5.333cm 0.333cm},clip,width=.17\linewidth]{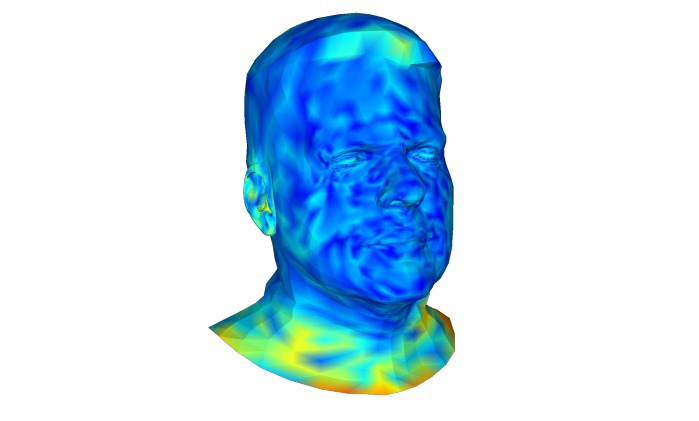}
    } 
    \subfloat
    {
        \includegraphics[trim={5cm 0.833cm 6.333cm 0.666cm},clip,width=.17\linewidth]{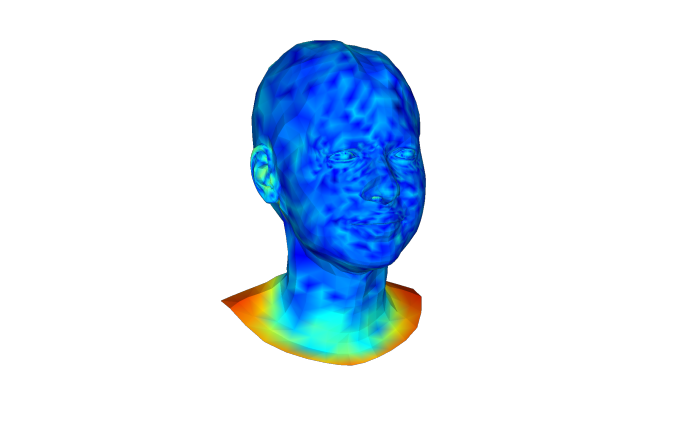}
    } 
    \subfloat
    {
        \includegraphics[trim={5.333cm 0.666cm 5.333cm 0.333cm},clip,width=.17\linewidth]{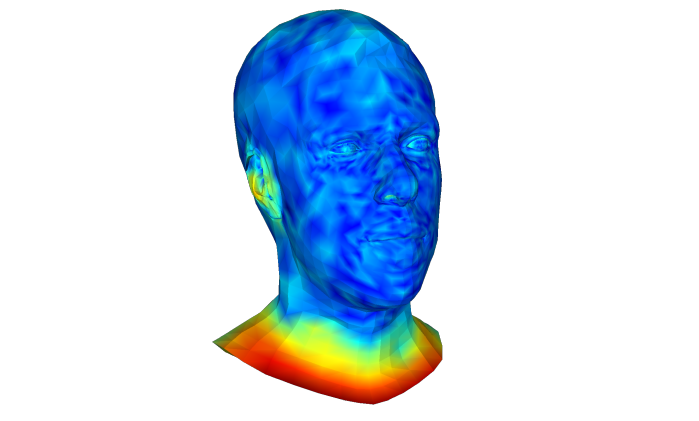}
    }
    \vspace{-8pt}
    \caption{The top row visualizes the one-sided Chamfer distance to the target scan \emph{without} vertex displacement, and the bottom row illustrates the improved alignment \emph{with} vertex displacements. Incorporating vertex displacements enhances the geometric accuracy of the head model, particularly in the hair region.}
    \label{fig:results3}
\end{figure}
%
%%%%%%%%%%%%%%%%%%%%%%%%%%%%%%%%%%%%%%%%%%%%%%%%%%%%%%%%%%%%%%%%%%%%%%%%%%%%%%%%%%%%%%%%%%%%%%
%
\par
%
%%%%%%%%%%%%%%%%%%%%%%%%%%%%%%%%%%%%%%%%%%%%%%%%%%%%%%%%%%%%%%%%%%%%%%%%%%%%%%%%%%%%%%%%%%%%%%
%
\paragraph{Shape Fitting} 
%
%%%%%%%%%%%%%%%%%%%%%%%%%%%%%%%%%%%%%%%%%%%%%%%%%%%%%%%%%%%%%%%%%%%%%%%%%%%%%%%%%%%%%%%%%%%%%%
%
Shape fitting involves a masked one-sided Chamfer loss $\mathcal{L}_\mathrm{Chamf}^\rightharpoonup$ between the FLAME mesh and the target scan, weighted by $\alpha_\mathrm{Chamf} = 1000.0$. The Chamfer loss is expressed as:
%
%%%%%%%%%%%%%%%%%%%%%%%%%%%%%%%%%%%%%%%%%%%%%%%%%%%%%%%%%%%%%%%%%%%%%%%%%%%%%%%%%%%%%%%%%%%%%%
%
\begin{equation}\label{eqn:chamfer_one}
    \mathcal{L}_\mathrm{Chamf}^\rightharpoonup(\mathbf{P}_H', \mathbf{P}_\mathrm{scan}) = \frac{1}{N_H} \sum_{\mathbf{p}_H' \in \mathbf{P}_H'} \min_{\mathbf{p}_\mathrm{scan} \in \mathbf{P}_\mathrm{scan}} \lvert\vert \mathbf{p}_H' - \mathbf{p}_\mathrm{scan}\rvert\rvert_2^2
\end{equation}
%
%%%%%%%%%%%%%%%%%%%%%%%%%%%%%%%%%%%%%%%%%%%%%%%%%%%%%%%%%%%%%%%%%%%%%%%%%%%%%%%%%%%%%%%%%%%%%%
%
where $\mathbf{P}_H'$ are $N_H$ uniformly sampled points on the FLAME mesh and $\mathbf{P}_\mathrm{scan}$ are points on the target scan. We also include a 3D landmark loss $\mathcal{L}_\mathrm{Lmk3D}$ as defined in Eq.~\ref{eqn:lmk3d}, weighted by $\alpha_\mathrm{Lmk3D} = 0.1$. The fitting process follows a \textit{freeze and refine} strategy (covered in the main paper) to optimize shape parameters $\boldsymbol{\beta}$, global translation $\boldsymbol{\theta}_\mathrm{trans}$, and global rotation $\boldsymbol{\theta}_\mathrm{rot}$, conducted in three steps of $1{,}000$ iterations each, with an initial learning rate of $0.01$. An early stopping mechanism halts optimization if no significant improvement is observed. The target scan must depict a neutral facial expression, as only shape parameters $\boldsymbol{\beta}$ are optimized. To prevent overfitting, we add a regularization term, weighted by $\alpha_{reg} = 0.000005$, which consists solely of the shape parameters $\boldsymbol{\beta}$, effectively pulling them toward zero.
%
%%%%%%%%%%%%%%%%%%%%%%%%%%%%%%%%%%%%%%%%%%%%%%%%%%%%%%%%%%%%%%%%%%%%%%%%%%%%%%%%%%%%%%%%%%%%%%
%
\par
%
%%%%%%%%%%%%%%%%%%%%%%%%%%%%%%%%%%%%%%%%%%%%%%%%%%%%%%%%%%%%%%%%%%%%%%%%%%%%%%%%%%%%%%%%%%%%%%
%
\paragraph{Per-Vertex Displacements} 
%
%%%%%%%%%%%%%%%%%%%%%%%%%%%%%%%%%%%%%%%%%%%%%%%%%%%%%%%%%%%%%%%%%%%%%%%%%%%%%%%%%%%%%%%%%%%%%%
%
For per-vertex displacement optimization, we use the Chamfer loss $\mathcal{L}_\mathrm{Chamf}^\rightharpoonup$ (defined in Eq.~\ref{eqn:chamfer_one}) with weight $\alpha_\mathrm{Chamf} = 100.0$, but without masking. Instead, regional fitting weights (described in the main paper) are applied to adjust the gradient contributions for specific areas. Importantly, the Chamfer loss is not computed directly between mesh vertices; instead, points are uniformly sampled on the mesh surfaces. This approach avoids pulling vertices directly into one another and ensures a smooth, regular mesh topology.
The optimization runs for $1{,}000$ iterations with an initial learning rate of $0.00002$, and early stopping interrupts the process if no significant progress is made. The improved accuracy of the personalized head avatar after incorporating vertex displacements is illustrated in Fig.~\ref{fig:results3}.
%
%%%%%%%%%%%%%%%%%%%%%%%%%%%%%%%%%%%%%%%%%%%%%%%%%%%%%%%%%%%%%%%%%%%%%%%%%%%%%%%%%%%%%%%%%%%%%%
%
\subsection{Motion Optimization and Expression Tracking}
\begin{figure}[t]
    \centering
    \subfloat[FAN $68$]
    {
        \includegraphics[trim={15cm 7.5cm 11cm 2cm},clip,width=.23\linewidth]{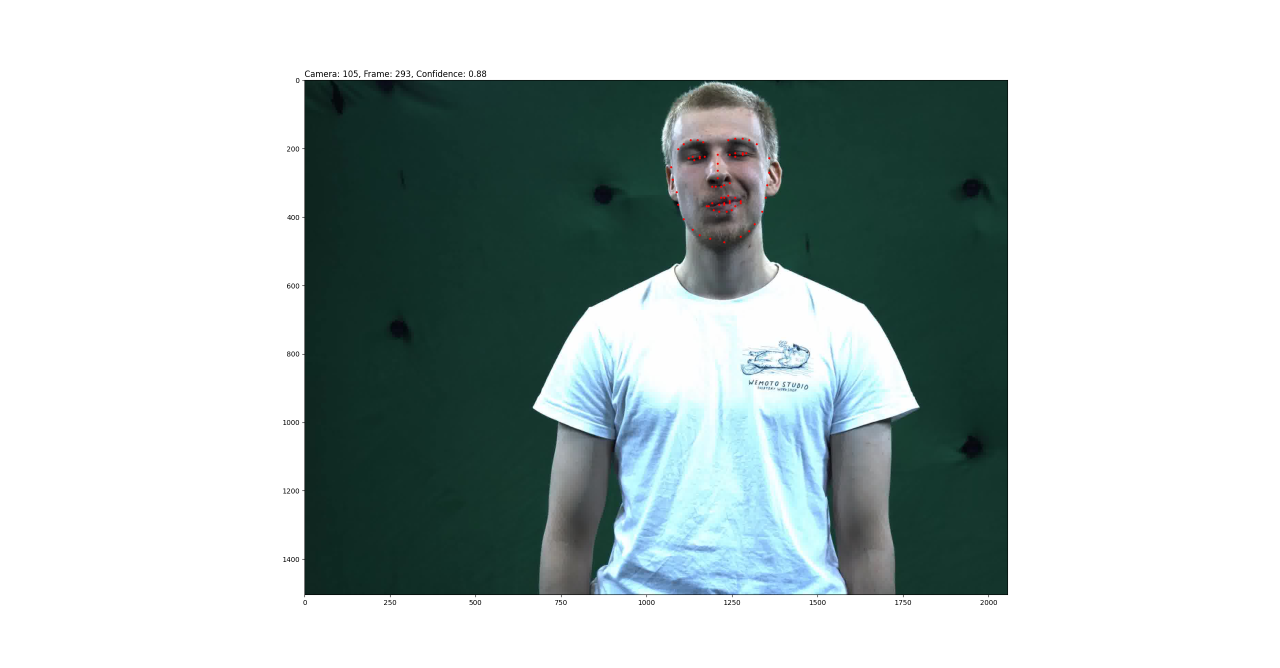}
    } 
    \subfloat[FAN $51$]
    {
        \includegraphics[trim={15cm 7.5cm 11cm 2cm},clip,width=.23\linewidth]{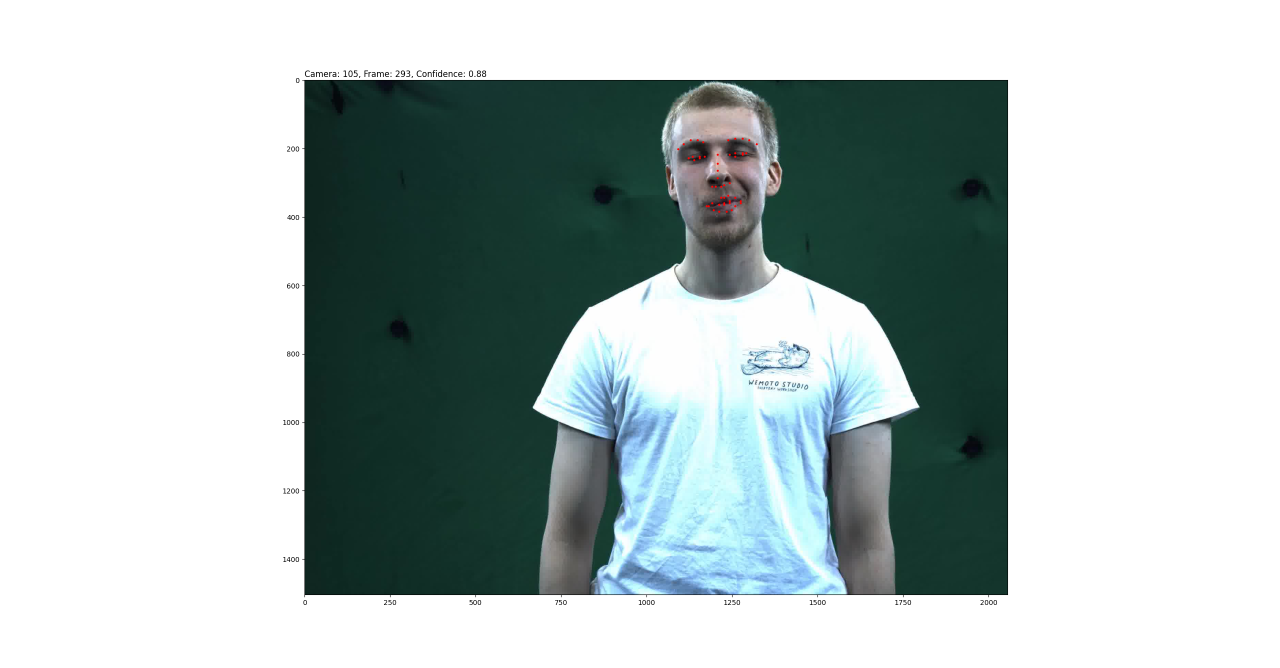}
    } 
    \subfloat[MP 468]
    {
        \includegraphics[trim={15cm 7.5cm 11cm 2cm},clip,width=.23\linewidth]{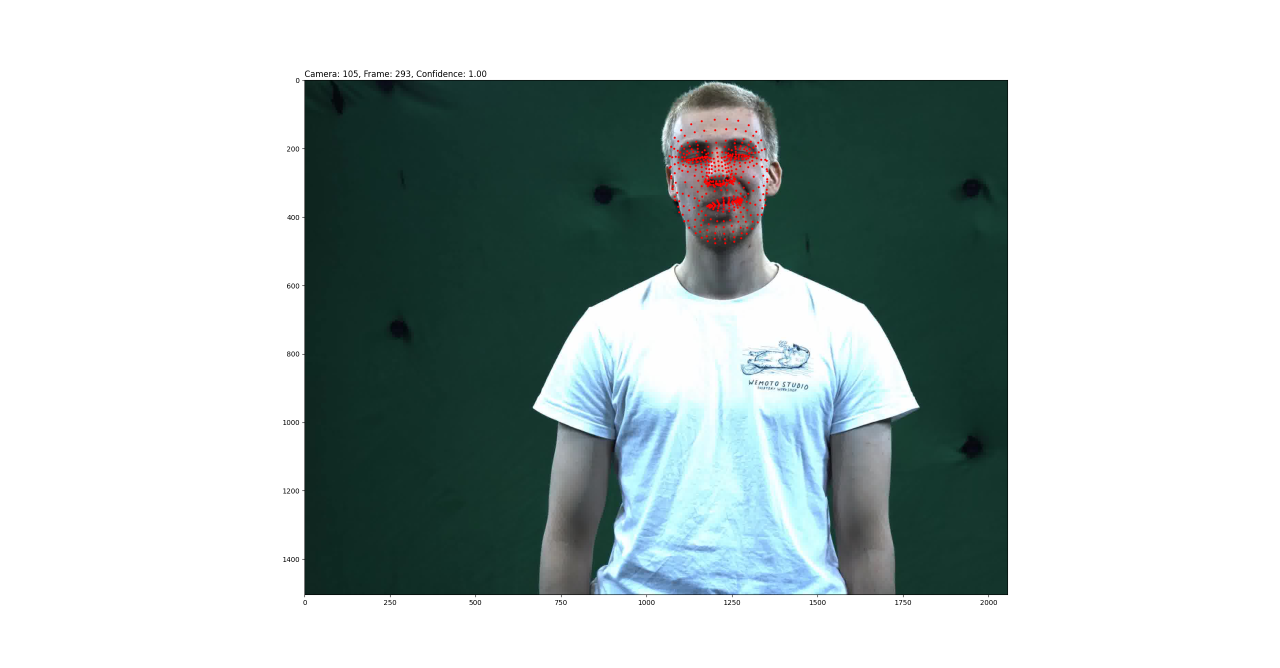}
    } 
    \subfloat[MP 105]
    {
        \includegraphics[trim={15cm 7.5cm 11cm 2cm},clip,width=.23\linewidth]{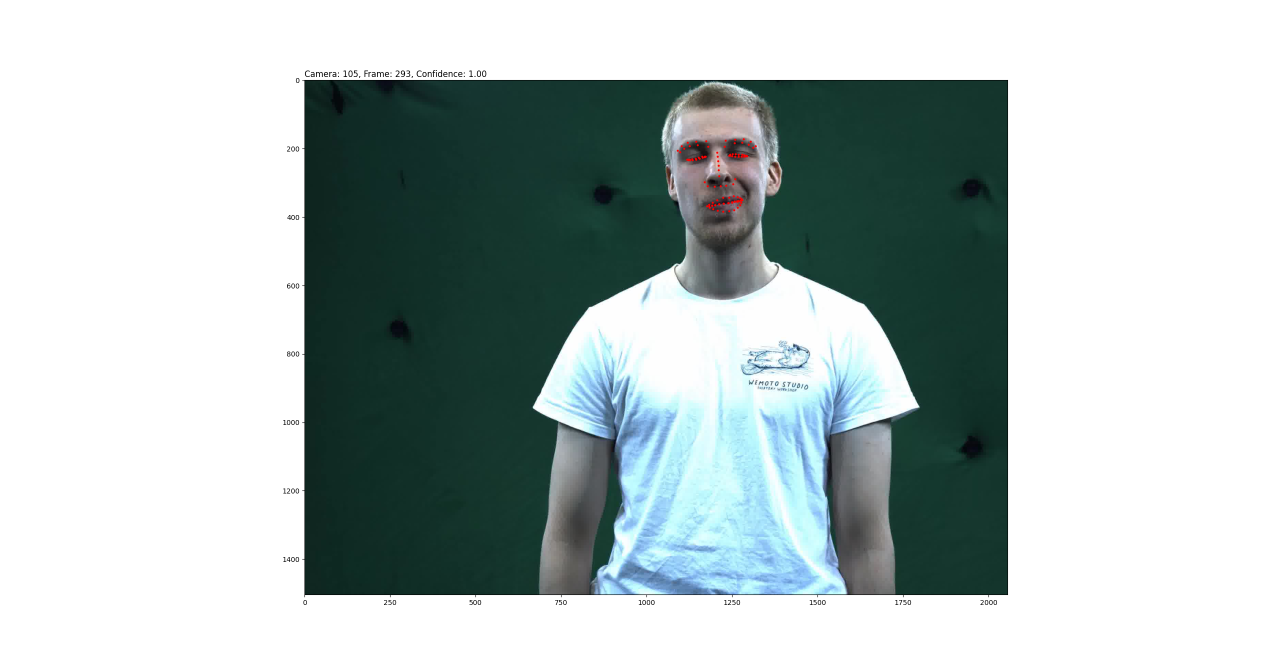}
    }
    \vspace{-8pt}
    \caption{Visualization of 2D Face Landmark Predictions: 2D face landmarks detected using the FAN framework \citep{bulat2017far} and Mediapipe (MP) \citep{mediapipe}.}
    \label{fig:landmarks}
\end{figure}

%
%%%%%%%%%%%%%%%%%%%%%%%%%%%%%%%%%%%%%%%%%%%%%%%%%%%%%%%%%%%%%%%%%%%%%%%%%%%%%%%%%%%%%%%%%%%%%%
%
\paragraph{Camera Selection and 2D Landmark Detection}
%
%%%%%%%%%%%%%%%%%%%%%%%%%%%%%%%%%%%%%%%%%%%%%%%%%%%%%%%%%%%%%%%%%%%%%%%%%%%%%%%%%%%%%%%%%%%%%%
%
For each frame, we select cameras where the face is clearly visible, as described in the main paper. From each selected camera, we detect $468$ Mediapipe \citep{mediapipe} landmarks and $68$ FAN \citep{bulat2017far} landmarks. However, we only use $51$ FAN landmarks and $105$ Mediapipe landmarks that correspond to FLAME landmark embeddings. Fig.~\ref{fig:landmarks} shows examples of the 2D landmark predictions. Notably, FAN landmarks include confidence values, which are used to weight their influence during motion optimization and expression tracking. The landmark detection process is highly parallelized, with each camera processed independently on a dedicated GPU or CPU. The final predicted landmarks are then used in motion optimization and expression tracking.
%
%%%%%%%%%%%%%%%%%%%%%%%%%%%%%%%%%%%%%%%%%%%%%%%%%%%%%%%%%%%%%%%%%%%%%%%%%%%%%%%%%%%%%%%%%%%%%%
%
%
%%%%%%%%%%%%%%%%%%%%%%%%%%%%%%%%%%%%%%%%%%%%%%%%%%%%%%%%%%%%%%%%%%%%%%%%%%%%%%%%%%%%%%%%%%%%%%
%
\paragraph{Motion Optimization}
%
%%%%%%%%%%%%%%%%%%%%%%%%%%%%%%%%%%%%%%%%%%%%%%%%%%%%%%%%%%%%%%%%%%%%%%%%%%%%%%%%%%%%%%%%%%%%%%
%
The optimization process utilizes a batch size of $1200$, an initial learning rate of $0.0001$, and runs for $1000$ iterations, with an early stopping mechanism to terminate the process when no significant improvement is observed. The pipeline is implemented in PyTorch \citep{NEURIPS2019_9015} and uses a standard gradient descent optimization approach using the Adam optimizer \citep{KingBa15} to refine skeletal pose parameters $\boldsymbol{\theta}$.
%
%%%%%%%%%%%%%%%%%%%%%%%%%%%%%%%%%%%%%%%%%%%%%%%%%%%%%%%%%%%%%%%%%%%%%%%%%%%%%%%%%%%%%%%%%%%%%%
%
\par
%
%%%%%%%%%%%%%%%%%%%%%%%%%%%%%%%%%%%%%%%%%%%%%%%%%%%%%%%%%%%%%%%%%%%%%%%%%%%%%%%%%%%%%%%%%%%%%%
%
Motion optimization employs a bidirectional Chamfer loss $\mathcal{L}_\mathrm{Chamf}^\rightleftharpoons$ with a weight of $\alpha_\mathrm{Chamf} = 1.5$, minimizing the distance between the posed expressive template mesh and per-frame surface reconstructions \citep{neus2}. The loss is defined as:
%
%%%%%%%%%%%%%%%%%%%%%%%%%%%%%%%%%%%%%%%%%%%%%%%%%%%%%%%%%%%%%%%%%%%%%%%%%%%%%%%%%%%%%%%%%%%%%%
%
\begin{equation}
    \mathcal{L}_\mathrm{Chamf}^\rightleftharpoons(\mathbf{P}_1, \mathbf{P}_2) =  \mathcal{L}_\mathrm{Chamf}^\rightharpoonup(\mathbf{P}_1, \mathbf{P}_2) +  \mathcal{L}_\mathrm{Chamf}^\rightharpoonup(\mathbf{P}_2, \mathbf{P}_1).
\end{equation}
%
%%%%%%%%%%%%%%%%%%%%%%%%%%%%%%%%%%%%%%%%%%%%%%%%%%%%%%%%%%%%%%%%%%%%%%%%%%%%%%%%%%%%%%%%%%%%%%
%
Additionally, a 2D landmark mean squared error (MSE) loss $\mathcal{L}_\mathrm{Lmk2D}$ with a weight of $\alpha_\mathrm{Lmk2D} = 100$ is utilized:
%
%%%%%%%%%%%%%%%%%%%%%%%%%%%%%%%%%%%%%%%%%%%%%%%%%%%%%%%%%%%%%%%%%%%%%%%%%%%%%%%%%%%%%%%%%%%%%%
%
\begin{equation}
    \mathcal{L}_\mathrm{Lmk2D}(\mathbf{V}, \mathbf{L}_\mathcal{V}) = \sqrt{\sum_{i=1}^{N_\mathrm{lmk}} \lvert\lvert \pi_\mathcal{V}(g_i(\mathbf{V})) - \mathbf{l}_i}\rvert\rvert_2^2.
\end{equation}
%
%%%%%%%%%%%%%%%%%%%%%%%%%%%%%%%%%%%%%%%%%%%%%%%%%%%%%%%%%%%%%%%%%%%%%%%%%%%%%%%%%%%%%%%%%%%%%%
%
Here, $\mathbf{V}$ represents the vertices of the expressive deformable template mesh, and $\mathbf{L}_\mathcal{V}$ denotes the predicted 2D landmarks for a given camera viewpoint $\mathcal{V}$. Based on these inputs, the 3D face landmarks on the FLAME mesh are computed, projected to 2D ($\pi_\mathcal{V}$), and compared with the corresponding 2D landmarks $\mathbf{l}_i$. 
%
%%%%%%%%%%%%%%%%%%%%%%%%%%%%%%%%%%%%%%%%%%%%%%%%%%%%%%%%%%%%%%%%%%%%%%%%%%%%%%%%%%%%%%%%%%%%%%
%
\par
%
%%%%%%%%%%%%%%%%%%%%%%%%%%%%%%%%%%%%%%%%%%%%%%%%%%%%%%%%%%%%%%%%%%%%%%%%%%%%%%%%%%%%%%%%%%%%%%
%
To ensure temporal smoothness, a total variation regularization, $\mathcal{L}_\mathrm{TV}$, with a weight of $\alpha_\mathrm{TV} = 10{,}000$, is applied. This loss minimizes the difference between neighboring poses to reduce temporal jittering:
%
%%%%%%%%%%%%%%%%%%%%%%%%%%%%%%%%%%%%%%%%%%%%%%%%%%%%%%%%%%%%%%%%%%%%%%%%%%%%%%%%%%%%%%%%%%%%%%
%
\begin{equation}
    \mathcal{L}_\mathrm{TV}(\boldsymbol{\theta}^\triangleright) = \frac{1}{N_\theta} \sum_{t=0}^{N-2} \lvert \boldsymbol{\theta}_t - \boldsymbol{\theta}_{t+1}\rvert.
\end{equation}
%
%%%%%%%%%%%%%%%%%%%%%%%%%%%%%%%%%%%%%%%%%%%%%%%%%%%%%%%%%%%%%%%%%%%%%%%%%%%%%%%%%%%%%%%%%%%%%%
%
Here, $\boldsymbol{\theta}^\triangleright$ represents the entire motion sequence being optimized, while $\boldsymbol{\theta}_t$ corresponds to individual poses for frame $t$.
%
%%%%%%%%%%%%%%%%%%%%%%%%%%%%%%%%%%%%%%%%%%%%%%%%%%%%%%%%%%%%%%%%%%%%%%%%%%%%%%%%%%%%%%%%%%%%%%
%
\paragraph{Expression Tracking}
%
%%%%%%%%%%%%%%%%%%%%%%%%%%%%%%%%%%%%%%%%%%%%%%%%%%%%%%%%%%%%%%%%%%%%%%%%%%%%%%%%%%%%%%%%%%%%%%
%
Expression tracking builds on the structure of motion optimization, extending it to include the optimization of both skeletal pose parameters $\boldsymbol{\theta}$ and facial expression parameters $\boldsymbol{\psi}$. Similar to motion optimization, the process uses a batch size of $1200$, an initial learning rate of $0.0001$, and runs for $1000$ iterations, with early stopping to terminate when no significant improvement is observed. The pipeline is implemented in PyTorch \citep{NEURIPS2019_9015} and employs gradient descent with the Adam optimizer \citep{KingBa15} to refine both $\boldsymbol{\theta}$ and $\boldsymbol{\psi}$. Notably, our approach incorporates an eyeblink post-processing step (detailed in Sec.~\ref{sec:supp_eyeblink}) that enhances eyelid parameters $\boldsymbol{\epsilon}$ to accurately capture eye blinks.
%
%%%%%%%%%%%%%%%%%%%%%%%%%%%%%%%%%%%%%%%%%%%%%%%%%%%%%%%%%%%%%%%%%%%%%%%%%%%%%%%%%%%%%%%%%%%%%%
%
\subsection{Disentangled 3D Gaussian Appearance}
\begin{figure}[t]
    \centering
    \subfloat[$\mathcal{T}_p^\mathrm{head}$]
    {
        \rotatebox[origin=c]{90}{\includegraphics[trim={0cm 0cm 0cm 0cm},clip,width=.23\linewidth]{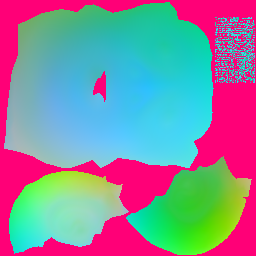}}
    } 
    \subfloat[$\mathcal{T}_n^\mathrm{head}$]
    {
        \rotatebox[origin=c]{90}{\includegraphics[trim={0cm 0cm 0cm 0cm},clip,width=.23\linewidth]{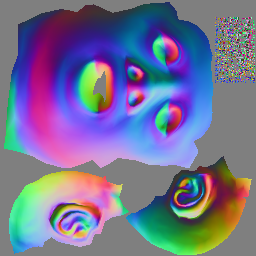}}
    } 
    \subfloat[$\mathcal{T}_p^\mathrm{body}$]
    {
        \rotatebox[origin=c]{90}{\includegraphics[trim={0cm 0cm 0cm 0cm},clip,width=.23\linewidth]{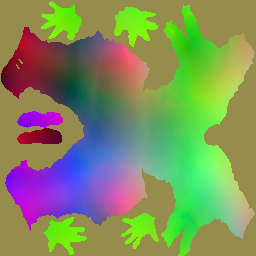}}
    } 
    \subfloat[$\mathcal{T}_n^\mathrm{body}$]
    {
        \rotatebox[origin=c]{90}{\includegraphics[trim={0cm 0cm 0cm 0cm},clip,width=.23\linewidth]{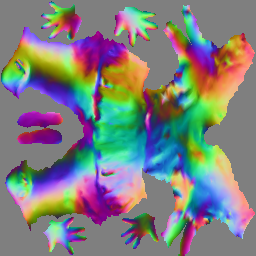}}
    } 
    \vspace{-8pt}
    \caption{
    Visualization of the normal and positional motion-aware textures for the head and body. The textures are derived from two separate UV mappings of our expressive template mesh. The "noise" in the top left corner of the head textures corresponds to the eyeball triangles.
    }
    \label{fig:textures}
\end{figure}

%
%%%%%%%%%%%%%%%%%%%%%%%%%%%%%%%%%%%%%%%%%%%%%%%%%%%%%%%%%%%%%%%%%%%%%%%%%%%%%%%%%%%%%%%%%%%%%%
%
The final disentangled 3D Gaussian appearance layer $\boldsymbol{\Phi}_\mathrm{app}$ is implemented in PyTorch \citep{NEURIPS2019_9015} and trained using the Adam optimizer \citep{KingBa15} to optimize the weights of the Gaussian parameter prediction networks $\phi_\mathrm{geo}^{(\cdot)}$ and $\phi_\mathrm{app}^{(\cdot)}$. Training is conducted on a single NVIDIA A$100$/A$40$ GPU. For the learned deformable character $C$, we leverage the results obtained during the training of DDC (see Sec.~\ref{sec:ddc_impl}).
Unlike the standard approach of training EVA on every $x$-th frame, we analyze the motion and select diverse poses that cover a broad range of actions. This strategy enhances appearance diversity in the training data and avoids the recurrence of a default standing pose in between actions. EVA is trained on cropped images with a resolution of $600\times800$ pixels, effectively equivalent to training on 2K resolution. For the body motion-aware texture, we use a size of $256\times256$, while the face motion-aware texture is set to $128\times128$ (see Fig.~\ref{fig:textures}).
We train the model with a batch size of $1$, an initial learning rate of $0.0001$, and optimize parameters over $4{,}000{,}000$ iterations. Four cameras are excluded during training and reserved for validation purposes. Consistent with the ASH framework \citep{Pang_2024_CVPR}, we incorporate $15{,}000$ warmup iterations to initialize the prediction networks.
%
%%%%%%%%%%%%%%%%%%%%%%%%%%%%%%%%%%%%%%%%%%%%%%%%%%%%%%%%%%%%%%%%%%%%%%%%%%%%%%%%%%%%%%%%%%%%%%
%
%
%%%%%%%%%%%%%%%%%%%%%%%%%%%%%%%%%%%%%%%%%%%%%%%%%%%%%%%%%%%%%%%%%%%%%%%%%%%%%%%%%%%%%%%%%%%%%%
%
We employ a pixel-wise L$1$ loss $\mathcal{L}_1$ with weight $\alpha_1 = 0.9$, a structural similarity index measure (SSIM) loss $\mathcal{L}_\mathrm{ssim}$ with weight $\alpha_\mathrm{ssim} = 0.1$, and an IDMRF loss \citep{wang2018image} $\mathcal{L}_\mathrm{IDMRF}$ with weight  $\alpha_\mathrm{IDMRF} = 0.01$. This combination ensures a balance between pixel-level accuracy (L$1$), perceptual quality (SSIM), and fine-grained texture details (IDMRF).
To enhance performance during training, we precompute motion-aware textures for the appearance model. This preprocessing step accelerates the training process. Notably, precomputing textures for $10{,}000$ frames requires approximately $50$ seconds. Furthermore, we adopt a random background augmentation covered in Sec.~\ref{sec:randomback_supp}.
%
%%%%%%%%%%%%%%%%%%%%%%%%%%%%%%%%%%%%%%%%%%%%%%%%%%%%%%%%%%%%%%%%%%%%%%%%%%%%%%%%%%%%%%%%%%%%%%
%
\section{Head Stitching} \label{sec:supp_head_stitching} 
%
%%%%%%%%%%%%%%%%%%%%%%%%%%%%%%%%%%%%%%%%%%%%%%%%%%%%%%%%%%%%%%%%%%%%%%%%%%%%%%%%%%%%%%%%%%%%%%
%
In this section, we introduce the stitching operator ($\bowtie$), which seamlessly combines our personalized head avatar $H$ and the learned deformable character $C$ to construct the final expressive template:
%
%%%%%%%%%%%%%%%%%%%%%%%%%%%%%%%%%%%%%%%%%%%%%%%%%%%%%%%%%%%%%%%%%%%%%%%%%%%%%%%%%%%%%%%%%%%%%%
%
\begin{equation}
    \boldsymbol{\Phi}_\mathrm{mesh}(\boldsymbol{\theta}^\triangleright, \boldsymbol{\psi}) = C(\boldsymbol{\theta}^\triangleright) \bowtie H(\boldsymbol{\psi}).
\end{equation}
%
%%%%%%%%%%%%%%%%%%%%%%%%%%%%%%%%%%%%%%%%%%%%%%%%%%%%%%%%%%%%%%%%%%%%%%%%%%%%%%%%%%%%%%%%%%%%%%
%
The primary objective of the stitching process is to connect the body vertices $\mathbf{V}_C$ and head vertices $\mathbf{V}_H$ into a single, watertight mesh. 
This is achieved by slicing both meshes along a defined cutting plane, aligning the resulting boundary vertices, and creating topological correspondences to generate new edges that form a seamless connection. Importantly, the entire stitching process is completely automated.
%
%%%%%%%%%%%%%%%%%%%%%%%%%%%%%%%%%%%%%%%%%%%%%%%%%%%%%%%%%%%%%%%%%%%%%%%%%%%%%%%%%%%%%%%%%%%%%%
%
\paragraph{Cutting Plane}
%
%%%%%%%%%%%%%%%%%%%%%%%%%%%%%%%%%%%%%%%%%%%%%%%%%%%%%%%%%%%%%%%%%%%%%%%%%%%%%%%%%%%%%%%%%%%%%%
%
To prepare for stitching, a cutting plane is defined to slice through both the posed body mesh generated by $C$ (which already contains a rough head geometry) and the head mesh created by $H$. The global position $\boldsymbol{\theta}_\mathrm{trans}$ and rotation $\boldsymbol{\theta}_\mathrm{rot}$ of the head avatar have already been optimized, ensuring that the two meshes are aligned in 3D space. The cutting plane is defined by marking vertices on the FLAME mesh, computing their centroid as the mean of their positions, and deriving the plane's normal vector using singular value decomposition (SVD) from the marked vertices. The slicing operation produces two non-watertight meshes, each with a single boundary loop at the neck, consisting of boundary vertices for the body $\mathbf{V}_C^\partial$ and for the head $\mathbf{V}_H^\partial$. The main challenge is connecting these vertices with newly generated triangles to create \emph{one} watertight mesh with final vertices $\mathbf{V}$.
%
%%%%%%%%%%%%%%%%%%%%%%%%%%%%%%%%%%%%%%%%%%%%%%%%%%%%%%%%%%%%%%%%%%%%%%%%%%%%%%%%%%%%%%%%%%%%%%
%
\paragraph{Boundary Alignment and Correspondence}
%
%%%%%%%%%%%%%%%%%%%%%%%%%%%%%%%%%%%%%%%%%%%%%%%%%%%%%%%%%%%%%%%%%%%%%%%%%%%%%%%%%%%%%%%%%%%%%%%
%
The stitching process begins by establishing a correspondence between the boundary vertices of the head mesh $\mathbf{V}_H^\partial$ and the body mesh $\mathbf{V}_C^\partial$. These boundary loops are represented as:
\begin{equation}
    \mathbf{V}_H^\partial = \left( \mathbf{v}_{H,i} \right)_{i=1}^{N_H^\partial}, \quad \mathbf{V}_C^\partial = \left( \mathbf{v}_{C,i} \right)_{i=1}^{N_C^\partial},
\end{equation}
where $\mathbf{v}_{(\cdot),i} \in \mathbb{R}^3$ are individual vertices, and $N_H^\partial$, $N_C^\partial$ denote the number of boundary vertices for the head and body, respectively.
A topological correspondence is established iteratively. For each head boundary vertex $\mathbf{v}_{H,i}$, the nearest body boundary vertex $\mathbf{v}_{C,j}$ is found using the Euclidean distance:
%
%%%%%%%%%%%%%%%%%%%%%%%%%%%%%%%%%%%%%%%%%%%%%%%%%%%%%%%%%%%%%%%%%%%%%%%%%%%%%%%%%%%%%%%%%%%%%%
%
\begin{equation}
    \mathbf{v}_{C,j} = \arg\underset{\mathbf{v} \in \mathbf{V}_C^\partial}{\min} \| \mathbf{v}_{H,i} - \mathbf{v} \|.
\end{equation}
%
%%%%%%%%%%%%%%%%%%%%%%%%%%%%%%%%%%%%%%%%%%%%%%%%%%%%%%%%%%%%%%%%%%%%%%%%%%%%%%%%%%%%%%%%%%%%%%
%
\paragraph{Mesh Stitching}
%
%%%%%%%%%%%%%%%%%%%%%%%%%%%%%%%%%%%%%%%%%%%%%%%%%%%%%%%%%%%%%%%%%%%%%%%%%%%%%%%%%%%%%%%%%%%%%%
%
With the correspondences established, new edges are created to connect the boundary vertices, forming triangular faces that seamlessly bridge the head and body meshes. The process iteratively connects vertices in a circular manner, starting with any head vertex and linking it to the nearest body vertex. Each head vertex $\mathbf{v}_{H,i}$ is then connected to its corresponding nearest body vertex $\mathbf{v}_{C,j}$ and all intermediate body vertices up to the last connected one. This approach ensures no body vertices are skipped, resulting in a consistent and watertight stitching that creates a seamless connection between the meshes.
Finally, we generate new triangular faces from the created edges, resulting in a unified, watertight mesh $\mathbf{V}$. This final mesh retains the expressiveness of the personalized head avatar $H$ and the motion-dependent dynamics of the learned character deformation model $C$.
%
%%%%%%%%%%%%%%%%%%%%%%%%%%%%%%%%%%%%%%%%%%%%%%%%%%%%%%%%%%%%%%%%%%%%%%%%%%%%%%%%%%%%%%%%%%%%%%
%
\section{Eyelid Postprocessing}\label{sec:supp_eyeblink}
%
%%%%%%%%%%%%%%%%%%%%%%%%%%%%%%%%%%%%%%%%%%%%%%%%%%%%%%%%%%%%%%%%%%%%%%%%%%%%%%%%%%%%%%%%%%%%%%
%
To accurately track eye blinks and closed eyes, we incorporate an automatic post-processing step that enhances the eyelid blendshapes $\boldsymbol{\epsilon}$. This process analyzes the spatial relationship between the upper and lower eye landmarks across an entire sequence. We scale and modify $\boldsymbol{\epsilon}$ based on these distances. 
%
%%%%%%%%%%%%%%%%%%%%%%%%%%%%%%%%%%%%%%%%%%%%%%%%%%%%%%%%%%%%%%%%%%%%%%%%%%%%%%%%%%%%%%%%%%%%%%
%
\par
%
%%%%%%%%%%%%%%%%%%%%%%%%%%%%%%%%%%%%%%%%%%%%%%%%%%%%%%%%%%%%%%%%%%%%%%%%%%%%%%%%%%%%%%%%%%%%%%
%
Let $\mathbf{D} \in \mathbb{R}^{2\times N_f}$ represent the landmark differences for $N_f$ frames, a matrix that contains the positional differences between the upper and lower eyelids across all frames for the left and right eyes. The mean difference vector is denoted as $\boldsymbol{\mu} \in \mathbb{R}^2$, representing the average openness of the left and right eyes. 
The differences are then centered around this mean:
%
%%%%%%%%%%%%%%%%%%%%%%%%%%%%%%%%%%%%%%%%%%%%%%%%%%%%%%%%%%%%%%%%%%%%%%%%%%%%%%%%%%%%%%%%%%%%%%
%
\begin{equation}
    \mathbf{D} \leftarrow -(\mathbf{D} - \boldsymbol{\mu}).
\end{equation}
%
%%%%%%%%%%%%%%%%%%%%%%%%%%%%%%%%%%%%%%%%%%%%%%%%%%%%%%%%%%%%%%%%%%%%%%%%%%%%%%%%%%%%%%%%%%%%%%
%
For consistent scaling, the elements of $\mathbf{D}$ are normalized for positive values $ \boldsymbol{D}_\mathrm{pos}$ and negative values $\boldsymbol{D}_\mathrm{neg}$ separately:
%
%%%%%%%%%%%%%%%%%%%%%%%%%%%%%%%%%%%%%%%%%%%%%%%%%%%%%%%%%%%%%%%%%%%%%%%%%%%%%%%%%%%%%%%%%%%%%%
%
\begin{equation}
    \boldsymbol{D}_\mathrm{norm} =
    \begin{cases} 
        \dfrac{\boldsymbol{D}_\mathrm{pos} - \min \boldsymbol{D}_\mathrm{pos}}{\max \boldsymbol{D}_\mathrm{pos} - \min \boldsymbol{D}_\mathrm{pos}} \\[10pt]
        \dfrac{\boldsymbol{D}_\mathrm{neg} - \min \boldsymbol{D}_\mathrm{neg}}{\max \boldsymbol{D}_\mathrm{neg} - \min \boldsymbol{D}_\mathrm{neg}}.
    \end{cases}
\end{equation}
%
%%%%%%%%%%%%%%%%%%%%%%%%%%%%%%%%%%%%%%%%%%%%%%%%%%%%%%%%%%%%%%%%%%%%%%%%%%%%%%%%%%%%%%%%%%%%%%
%
In this context, positive values of $\mathbf{D}_\mathrm{norm}$ indicate a closing of the eyes, while negative values indicate an opening. To account for non-optimal landmark detection that may not capture complete closure of the eyes, we empirically choose a threshold $\zeta  =0.75$ and set values greater than this threshold to one, indicating complete eye closure.
Given the eyelid parameters for a single frame $\boldsymbol{\epsilon}_t \in \mathbb{R}^2$, we can define the adjustment $\boldsymbol{\delta}_t  \in \mathbb{R}^2$ as:
%
%%%%%%%%%%%%%%%%%%%%%%%%%%%%%%%%%%%%%%%%%%%%%%%%%%%%%%%%%%%%%%%%%%%%%%%%%%%%%%%%%%%%%%%%%%%%%%
%
\begin{equation}
    \boldsymbol{\delta}_t = (\mathbf{1} - \boldsymbol{\epsilon}_t) \cdot \mathbf{D}_{\mathrm{norm},t}.
\end{equation}
%
%%%%%%%%%%%%%%%%%%%%%%%%%%%%%%%%%%%%%%%%%%%%%%%%%%%%%%%%%%%%%%%%%%%%%%%%%%%%%%%%%%%%%%%%%%%%%%
%
Here, $\mathbf{D}_{\mathrm{norm},t} \in \mathbb{R}^2$ denotes the normalized spatial difference for the left and right eye for frame $t$.
%
%%%%%%%%%%%%%%%%%%%%%%%%%%%%%%%%%%%%%%%%%%%%%%%%%%%%%%%%%%%%%%%%%%%%%%%%%%%%%%%%%%%%%%%%%%%%%%
%
\par
%
%%%%%%%%%%%%%%%%%%%%%%%%%%%%%%%%%%%%%%%%%%%%%%%%%%%%%%%%%%%%%%%%%%%%%%%%%%%%%%%%%%%%%%%%%%%%%%
%
Since open eyes are generally well-handled by 2D landmark detectors, and we assume the eyes are open on average, we downscale the opening adjustment by updating $\boldsymbol{\delta}$: 
%
%%%%%%%%%%%%%%%%%%%%%%%%%%%%%%%%%%%%%%%%%%%%%%%%%%%%%%%%%%%%%%%%%%%%%%%%%%%%%%%%%%%%%%%%%%%%%%
%
\begin{align}
    \boldsymbol{\delta}_t^\mathrm{left} &\leftarrow \boldsymbol{\delta}_t^\mathrm{left} \cdot \omega \text{,\ \ \ if \ \  } \boldsymbol{\delta}_t^\mathrm{left} < 0, \\
    \boldsymbol{\delta}_t^\mathrm{right} &\leftarrow  \boldsymbol{\delta}_t^\mathrm{right} \cdot \omega  \text{,\ \ \ if \ \  } \boldsymbol{\delta}_t^\mathrm{right} < 0.
\end{align}
%
%%%%%%%%%%%%%%%%%%%%%%%%%%%%%%%%%%%%%%%%%%%%%%%%%%%%%%%%%%%%%%%%%%%%%%%%%%%%%%%%%%%%%%%%%%%%%%
%
Here, $\boldsymbol{\delta}_t^{(\cdot)}$ is the adjustment of the left and right eye for frame $t$.
Finally, the updated eyelid parameters are given by:
%
%%%%%%%%%%%%%%%%%%%%%%%%%%%%%%%%%%%%%%%%%%%%%%%%%%%%%%%%%%%%%%%%%%%%%%%%%%%%%%%%%%%%%%%%%%%%%%
%
\begin{equation}
    \boldsymbol{\epsilon}_t \leftarrow \boldsymbol{\epsilon}_t + \boldsymbol{\delta}_t.
\end{equation} 
%
%%%%%%%%%%%%%%%%%%%%%%%%%%%%%%%%%%%%%%%%%%%%%%%%%%%%%%%%%%%%%%%%%%%%%%%%%%%%%%%%%%%%%%%%%%%%%%
%
This process refines the eyelid parameters to accommodate potential inaccuracies in landmark detection, ensuring fully closed eyes during blinking. The threshold $\zeta$ and scaling value $\omega$ for the eyelid opening are empirically chosen hyperparameters.
%
%%%%%%%%%%%%%%%%%%%%%%%%%%%%%%%%%%%%%%%%%%%%%%%%%%%%%%%%%%%%%%%%%%%%%%%%%%%%%%%%%%%%%%%%%%%%%%
%
\section{Runtime Analysis} \label{sec:supp_runtime}
%
%%%%%%%%%%%%%%%%%%%%%%%%%%%%%%%%%%%%%%%%%%%%%%%%%%%%%%%%%%%%%%%%%%%%%%%%%%%%%%%%%%%%%%%%%%%%%%
%
\begin{table}[t]
    \centering
    \setlength{\tabcolsep}{4pt} 
    \renewcommand{\arraystretch}{1} 
    \caption{Separate runtime analysis of the geometry layer and appearance layer. The low-quality (LQ), medium-quality (MQ), and high-quality (HQ) settings correspond to the appearance models $64/128$, $128/256$, and $256/512$, respectively. The method $x/y$ denotes the appearance model, where $x \times x$ is the texture resolution for the face and $y \times y$ for the body.}
    \vspace{-8pt}
    \label{tab:runtime}
    \begin{tabular}{l|c|cccccc}
        \toprule
         & \textbf{Geometry} & Tex & Pred & Tra & Ren & \textbf{Appearance} \\
        \midrule
        LQ & $24.4\text{\tiny ms}$ ($41\text{\tiny FPS}$) & $10\text{\tiny ms}$ & $9.4\text{\tiny ms}$ & $0.6\text{\tiny ms}$ & $2.1\text{\tiny ms}$ & $22.1\text{\tiny ms}$ ($45\text{\tiny FPS}$)\\
        MQ & $24.4\text{\tiny ms}$ ($41\text{\tiny FPS}$) & $10\text{\tiny ms}$ & $9.7\text{\tiny ms}$ & $6.8\text{\tiny ms}$ & $2.0\text{\tiny ms}$ & $28.5\text{\tiny ms}$ ($35\text{\tiny FPS}$) \\
        HQ & $24.4\text{\tiny ms}$ ($41\text{\tiny FPS}$) & $30\text{\tiny ms}$ & $9.8\text{\tiny ms}$ & $39.1\text{\tiny ms}$ & $2.8\text{\tiny ms}$ & $81.7\text{\tiny ms}$ ($12\text{\tiny FPS}$) \\
        \bottomrule
    \end{tabular}
    \vspace{-0pt}
\end{table}
%
%%%%%%%%%%%%%%%%%%%%%%%%%%%%%%%%%%%%%%%%%%%%%%%%%%%%%%%%%%%%%%%%%%%%%%%%%%%%%%%%%%%%%%%%%%%%%%
%
Tab.~\ref{tab:runtime} provides a detailed runtime analysis of the geometry layer $\boldsymbol{\Phi}_\mathrm{mesh}$ and the appearance layer $\boldsymbol{\Phi}_\mathrm{app}$ of EVA. This includes computations of motion-dependent character deformations using $\phi_{\mathrm{eg}}$ and $\phi_{\mathrm{delta}}$, as well as the separately executed appearance layer running the disentangled Gaussian predictors $\phi_\mathrm{geo}^{(\cdot)}$, $\phi_\mathrm{app}^{(\cdot)}$. The analysis considers a setup where the two layers are run on individual A$40$ GPUs, with the final rendered images having a resolution of $2056 \times 1504$ pixels. We define real-time performance as exceeding $25$ frames per second (fps).
To motivate the choice of texture size, we compare three configurations of appearance models with varying 2D texture embedding sizes: low quality (LQ), medium quality (MQ), and high quality (HQ). These configurations differ in the resolution of the motion-aware textures $\mathcal{T}_n^{(\cdot), \triangleright}$ and $\mathcal{T}_p^{(\cdot), \triangleright}$, which in turn determine the number of Gaussians used to model the human. We define the texture sizes for the low-, medium-, and high-quality models as $64/128$, $128/256$, and $256/512$, respectively, where $x/y$ indicates a texture resolution of $x \times x$ for face textures ($\mathcal{T}_n^{\mathrm{face},\triangleright}$, $\mathcal{T}_p^{\mathrm{face},\triangleright}$) and $y \times y$ for body textures ($\mathcal{T}_n^{\mathrm{body},\triangleright}$, $\mathcal{T}_p^{\mathrm{body},\triangleright}$).
Furthermore, we split the runtime analysis of the appearance layer into separate steps to give more insights:
%
%%%%%%%%%%%%%%%%%%%%%%%%%%%%%%%%%%%%%%%%%%%%%%%%%%%%%%%%%%%%%%%%%%%%%%%%%%%%%%%%%%%%%%%%%%%%%%
%
\begin{enumerate}
    \item Tex: Computing motion-aware textures $\mathcal{T}_n^{(\cdot), \triangleright}$ and $\mathcal{T}_p^{(\cdot), \triangleright}$ from the expressive deformable mesh generated by the geometry layer $\boldsymbol{\Phi}_\mathrm{mesh}$.
    \item Pred: Performing a feed-forward pass of the disentangled canonical Gaussian parameter predictors $\phi_\mathrm{geo}^{(\cdot)}$ and $\phi_\mathrm{app}^{(\cdot)}$.
    \item Tra: Computing posed and deformed 3D Gaussians for the head $\{\mathcal{G}^\mathrm{head}_i\}_{N_h}$ and the body $\{\mathcal{G}^\mathrm{body}_i\}_{N_b}$ through UV mapping and dual quaternion skinning.
    \item Ren: Tile-based rasterization of the 3D Gaussians $\{\mathcal{G}^\mathrm{head}_i\}_{N_h}$ and $\{\mathcal{G}^\mathrm{body}_i\}_{N_b}$ into a final 2D image with a resolution of $2056 \times 1504$ pixels.
\end{enumerate}
%
%%%%%%%%%%%%%%%%%%%%%%%%%%%%%%%%%%%%%%%%%%%%%%%%%%%%%%%%%%%%%%%%%%%%%%%%%%%%%%%%%%%%%%%%%%%%%%
%
\par 
%
%%%%%%%%%%%%%%%%%%%%%%%%%%%%%%%%%%%%%%%%%%%%%%%%%%%%%%%%%%%%%%%%%%%%%%%%%%%%%%%%%%%%%%%%%%%%%%
%
It is important to note that increasing the number of Gaussians through larger 2D embeddings does not make Gaussian parameter prediction and rendering the bottleneck. Instead, computing motion-aware textures and calculating the posed and deformed Gaussians account for a significant portion of the total runtime. For the high-quality model (HQ), generating the input 2D textures and transforming the 2D output features back into a posed and deformed 3D Gaussian model consumes approximately $85$\% of the total runtime.
%
%%%%%%%%%%%%%%%%%%%%%%%%%%%%%%%%%%%%%%%%%%%%%%%%%%%%%%%%%%%%%%%%%%%%%%%%%%%%%%%%%%%%%%%%%%%%%%
%
\section{Dataset}\label{sec:dataset_supp}
%
%%%%%%%%%%%%%%%%%%%%%%%%%%%%%%%%%%%%%%%%%%%%%%%%%%%%%%%%%%%%%%%%%%%%%%%%%%%%%%%%%%%%%%%%%%%%%%
%
\begin{figure}[t]
    \centering
    \subfloat[$S1$]
    {
        \includegraphics[trim={9.75cm 0.75cm 9.75cm 2.25cm},clip,width=.233\linewidth]{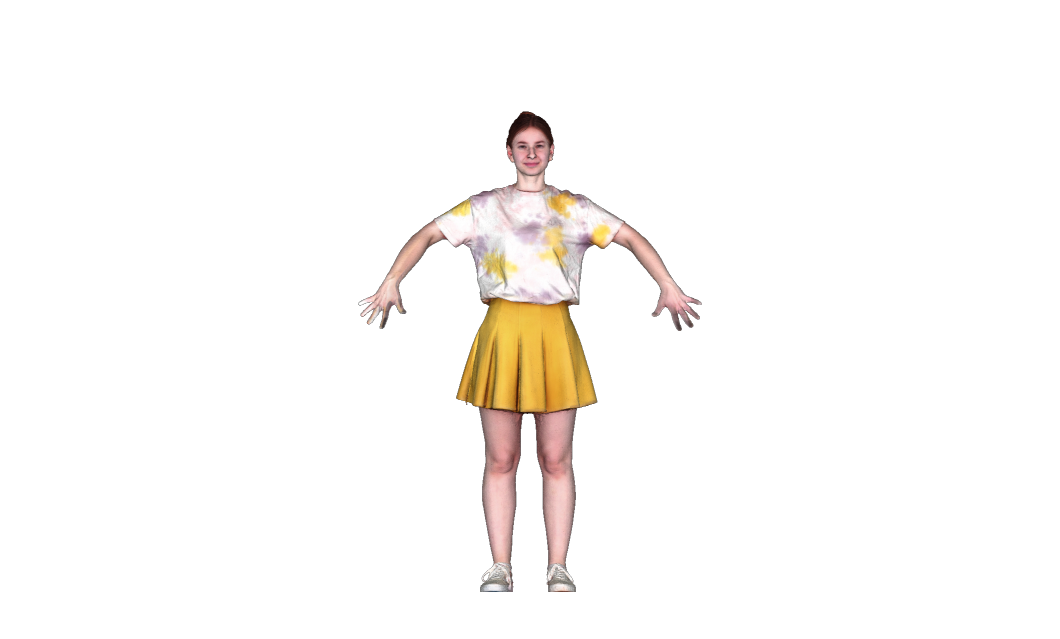}
    } 
    \subfloat[$S2$]
    {
        \includegraphics[trim={9.75cm 0.75cm 9.75cm 1.5cm},clip,width=.233\linewidth]{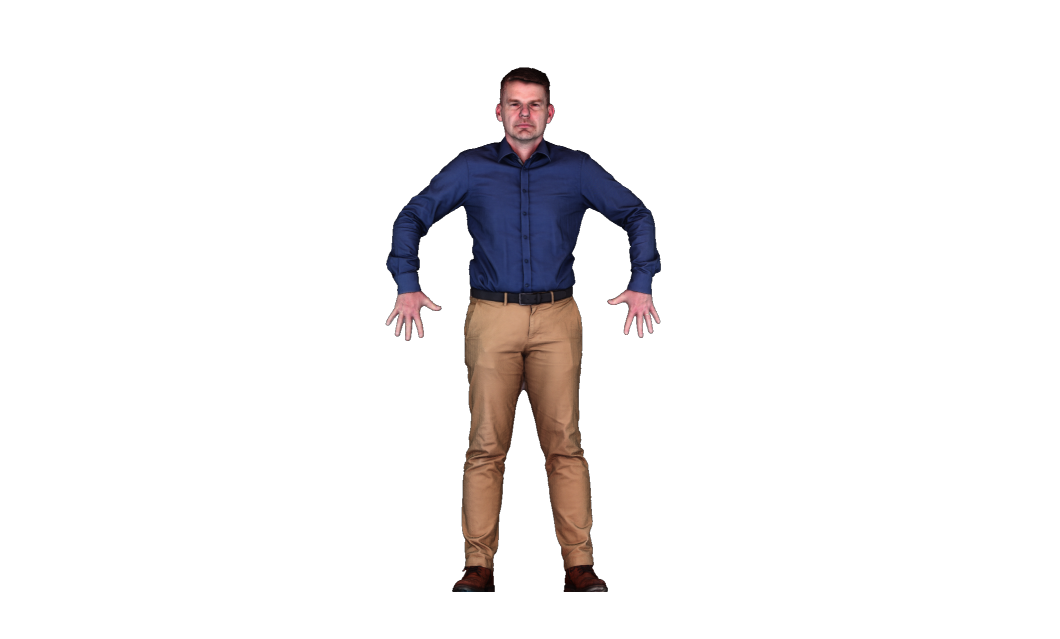}
    }
    \subfloat[$S3$]
    {
        \includegraphics[trim={9.75cm 0.75cm 9.75cm 2.15cm},clip,width=.233\linewidth]{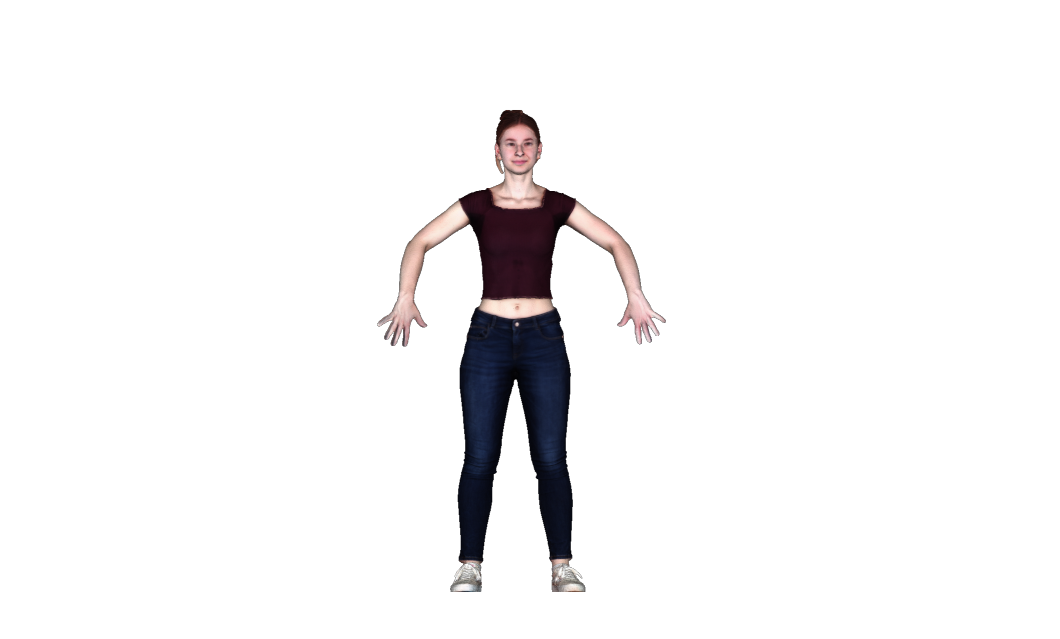}
    }
    \subfloat[$S4$]
    {
        \includegraphics[trim={9.75cm 0.75cm 9.75cm 1.15cm},clip,width=.233\linewidth]{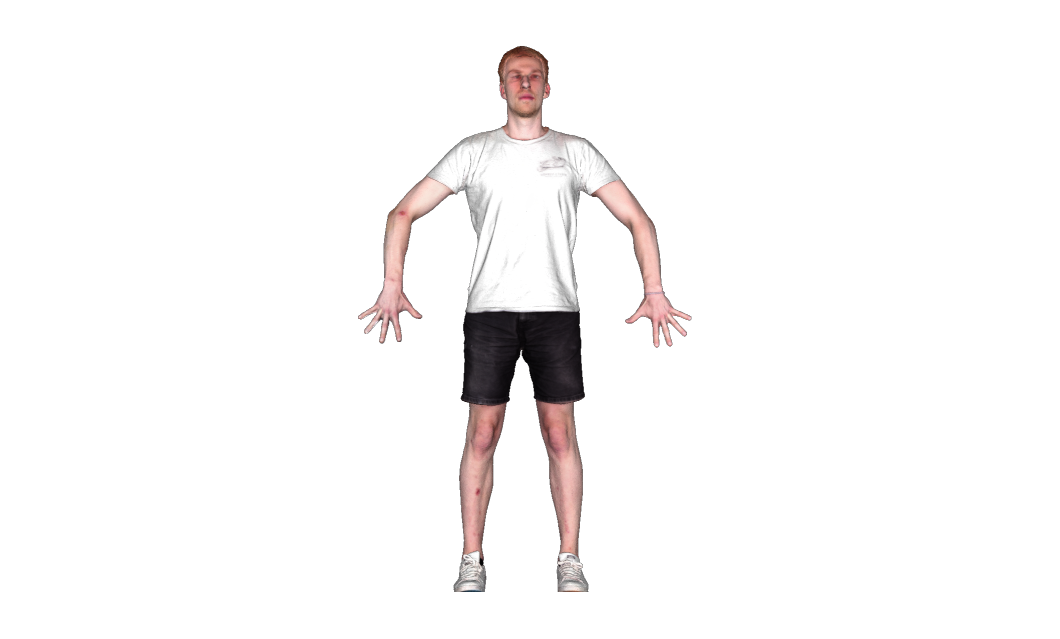}
    }
    \vspace{-8pt}
    \caption{Textured high-resolution 3D scans of all subjects in our dataset. Notably, subject $S2$ is provided by \citet{shetty2024holoported}.}
    \label{fig:subjectsoverview}
\end{figure}
%
%%%%%%%%%%%%%%%%%%%%%%%%%%%%%%%%%%%%%%%%%%%%%%%%%%%%%%%%%%%%%%%%%%%%%%%%%%%%%%%%%%%%%%%%%%%%%%
%
\begin{figure}[t]
    \centering
    \subfloat
    {
        \includegraphics[width=.19\linewidth]{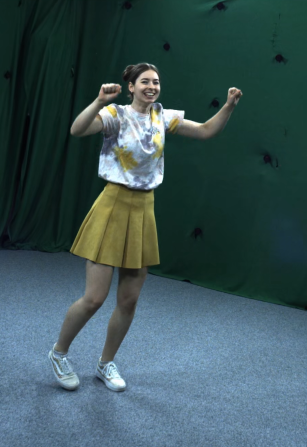}
    } 
    \hspace*{-0.7em}
    \subfloat
    {
        \includegraphics[width=.19\linewidth]{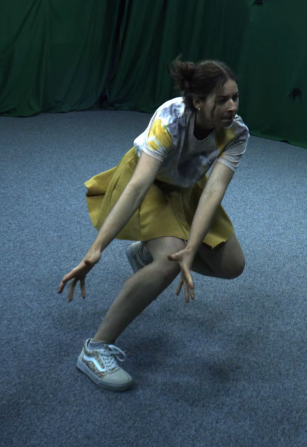}
    }
    \hspace*{-0.7em}
    \subfloat
    {
        \includegraphics[width=.19\linewidth]{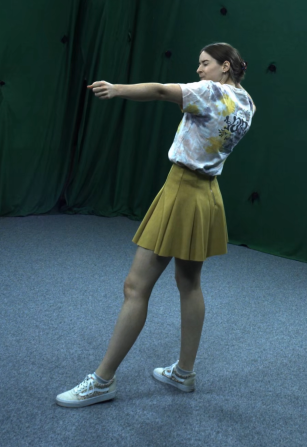}
    }
    \hspace*{-0.7em}
    \subfloat
    {
        \includegraphics[width=.19\linewidth]{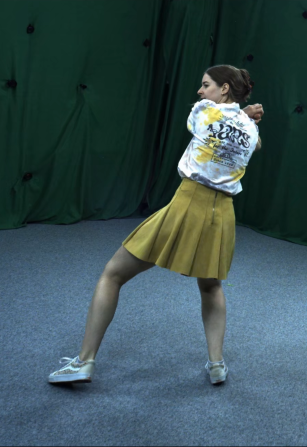}
    }
    \hspace*{-0.7em}
    \subfloat
    {
        \includegraphics[width=.19\linewidth]{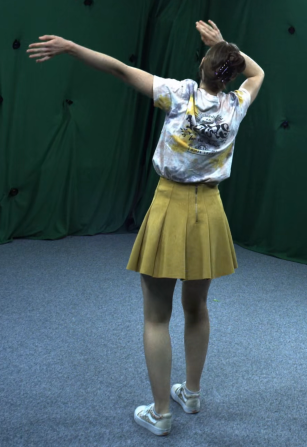}
    }
    \\
    \vspace*{-0.82em}
    \subfloat
    {
        \stackunder[5pt]{\includegraphics[trim={0cm 1cm 0cm 0.5cm},clip,width=.19\linewidth]{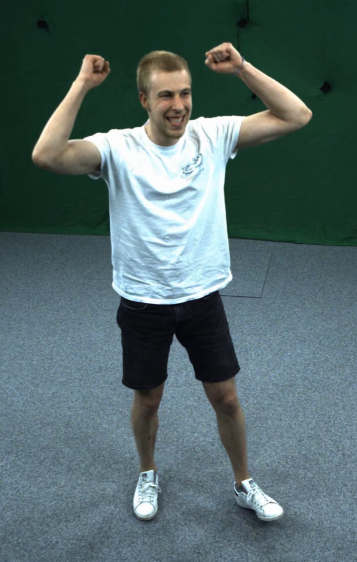}}{\small{(a) Celebrate}}
    } 
    \hspace*{-0.79em}
    \subfloat
    {
        \stackunder[5pt]{\includegraphics[trim={0cm 0.89cm 0cm 0.5cm},clip,width=.19\linewidth]{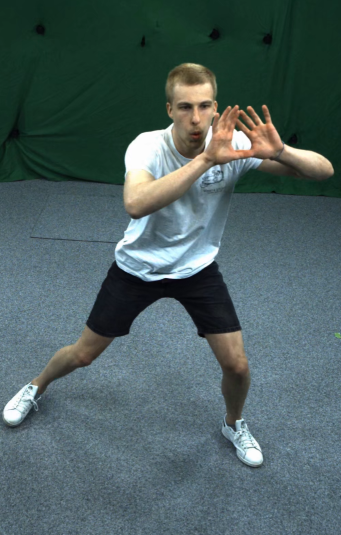}}{\small{(b) Goalkeeper}}
    }
    \hspace*{-0.78em}
    \subfloat
    {
        \stackunder[5pt]{\includegraphics[trim={0cm 1cm 0cm 0.5cm},clip,width=.19\linewidth]{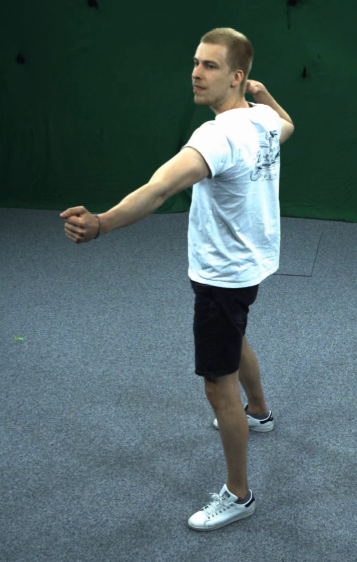}}{\small{(c) Archer}}
    }
    \hspace*{-0.7em}
    \subfloat
    {
        \stackunder[5pt]{\includegraphics[trim={0cm 1cm 0cm 0.5cm},clip,width=.19\linewidth]{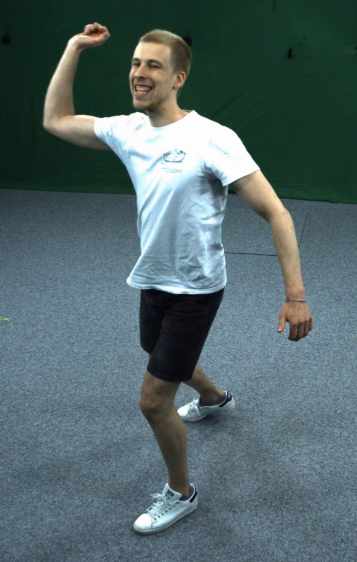}}{\small{(d) Baseball}}
    }
    \hspace*{-0.7em}
    \subfloat
    {
        \stackunder[5pt]{\includegraphics[trim={0cm 1cm 0cm 0.5cm},clip,width=.19\linewidth]{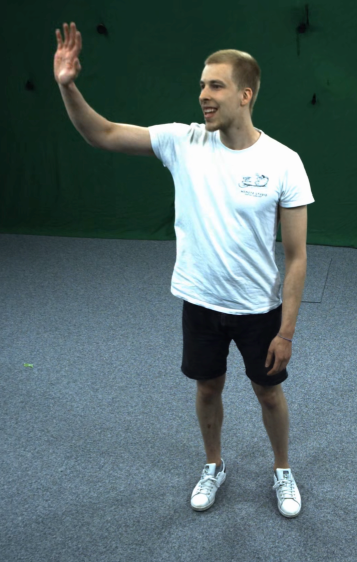}}{\small{(e) Wave}}
    }
    \vspace{-8pt}
    \caption{Cropped frames showcasing different actions performed by actors: celebrating, acting as a goalkeeper, shooting as an archer, playing baseball, and waving.}
    \label{fig:mvrecording}
\end{figure}
%
%%%%%%%%%%%%%%%%%%%%%%%%%%%%%%%%%%%%%%%%%%%%%%%%%%%%%%%%%%%%%%%%%%%%%%%%%%%%%%%%%%%%%%%%%%%%%%
%
Our dataset contains four subjects, each with garments that vary in fit, including both loose- and tight-fitting styles (see Fig.~\ref{fig:subjectsoverview}). Loose clothing, such as the skirt worn by $S1$, presents significant challenges for accurate modeling. As a result, this setting is particularly valuable for evaluating our model's ability to represent non-rigid deformations. We captured $S1$, $S3$, and $S4$ specifically for this project, while $S2$ is kindly provided by \citet{shetty2024holoported}. Each subject's constant geometry and texture are reconstructed from multi-view images using multi-view stereo reconstruction \citep{agisoft}. We employ a commercial 3D scanner \citep{treedys}.
For each subject, we capture multiple training ($27{,}000$ frames) and testing ($8{,}000$ frames) sequences in our multi-camera green-screen studio, equipped with $\approx 120$ synced $25$ fps cameras with $4112\times3008$ resolution. To ensure the authenticity of the data, actors are specifically instructed to incorporate strong facial expressions into their performances throughout the sequences. Fig.~\ref{fig:mvrecording} depicts example frames from the training data. Each frame is annotated with 3D skeletal poses \citep{thecaptury} that we further refine, FLAME parameters \citep{FLAME:SiggraphAsia2017}, foreground segmentations \citep{BGMv2, kirillov2023segany}, and 3D reconstructions \citep{neus2}. Refer to the main paper for more details regarding the dataset.
%
%%%%%%%%%%%%%%%%%%%%%%%%%%%%%%%%%%%%%%%%%%%%%%%%%%%%%%%%%%%%%%%%%%%%%%%%%%%%%%%%%%%%%%%%%%%%%%
%
\section{More Qualitative Results}\label{sec:results_supp}
%
%%%%%%%%%%%%%%%%%%%%%%%%%%%%%%%%%%%%%%%%%%%%%%%%%%%%%%%%%%%%%%%%%%%%%%%%%%%%%%%%%%%%%%%%%%%%%%
%
\subsection{Personalized Head Avatar} 
\begin{figure}[t]
    \centering
    \hspace*{-1em}
    \subfloat
    {
        \includegraphics[trim={15cm 2cm 19cm 2cm},clip,width=.20\linewidth]{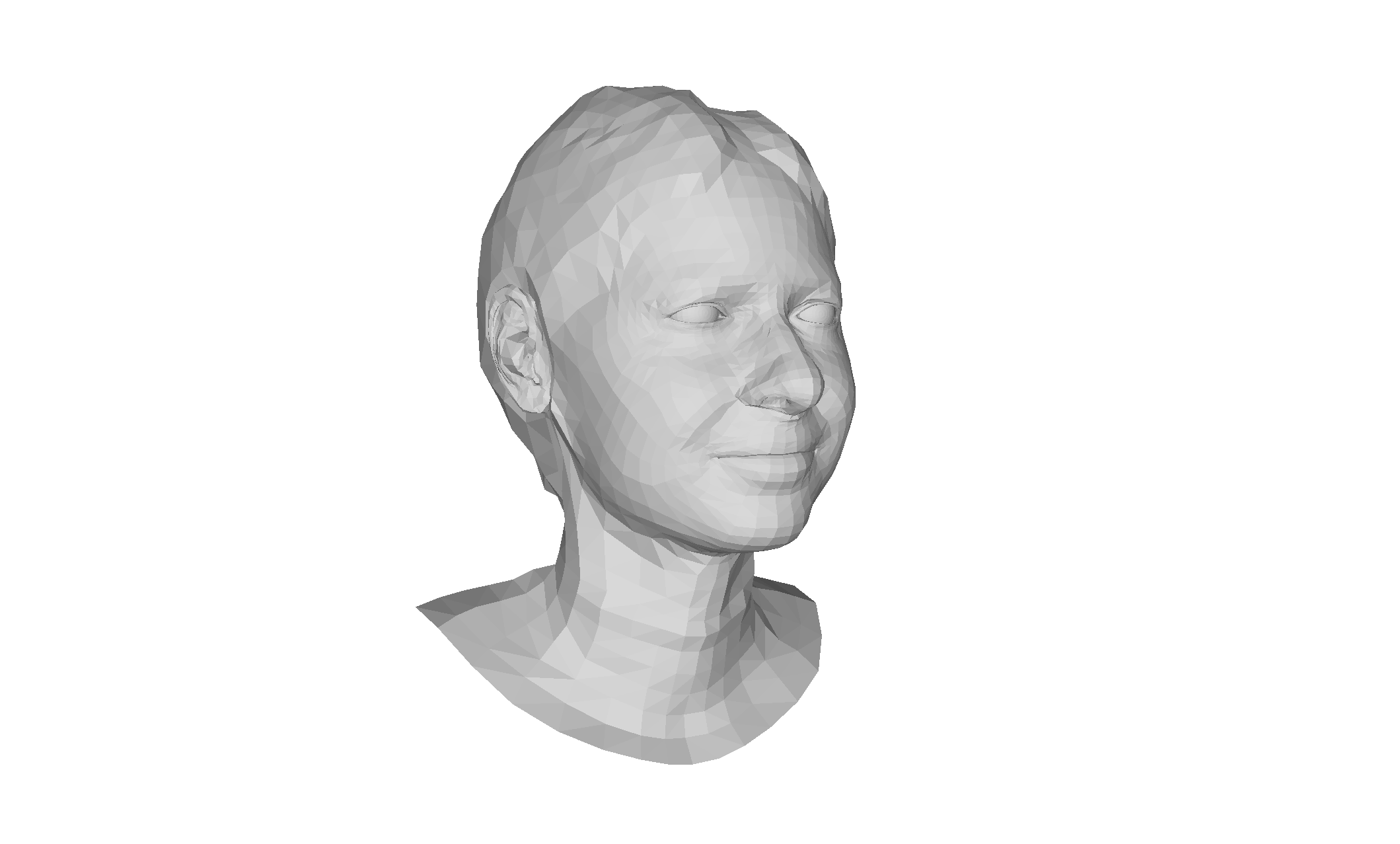}
    } 
    \subfloat
    {
        \includegraphics[trim={17cm 3cm 17cm 2cm},clip,width=.20\linewidth]{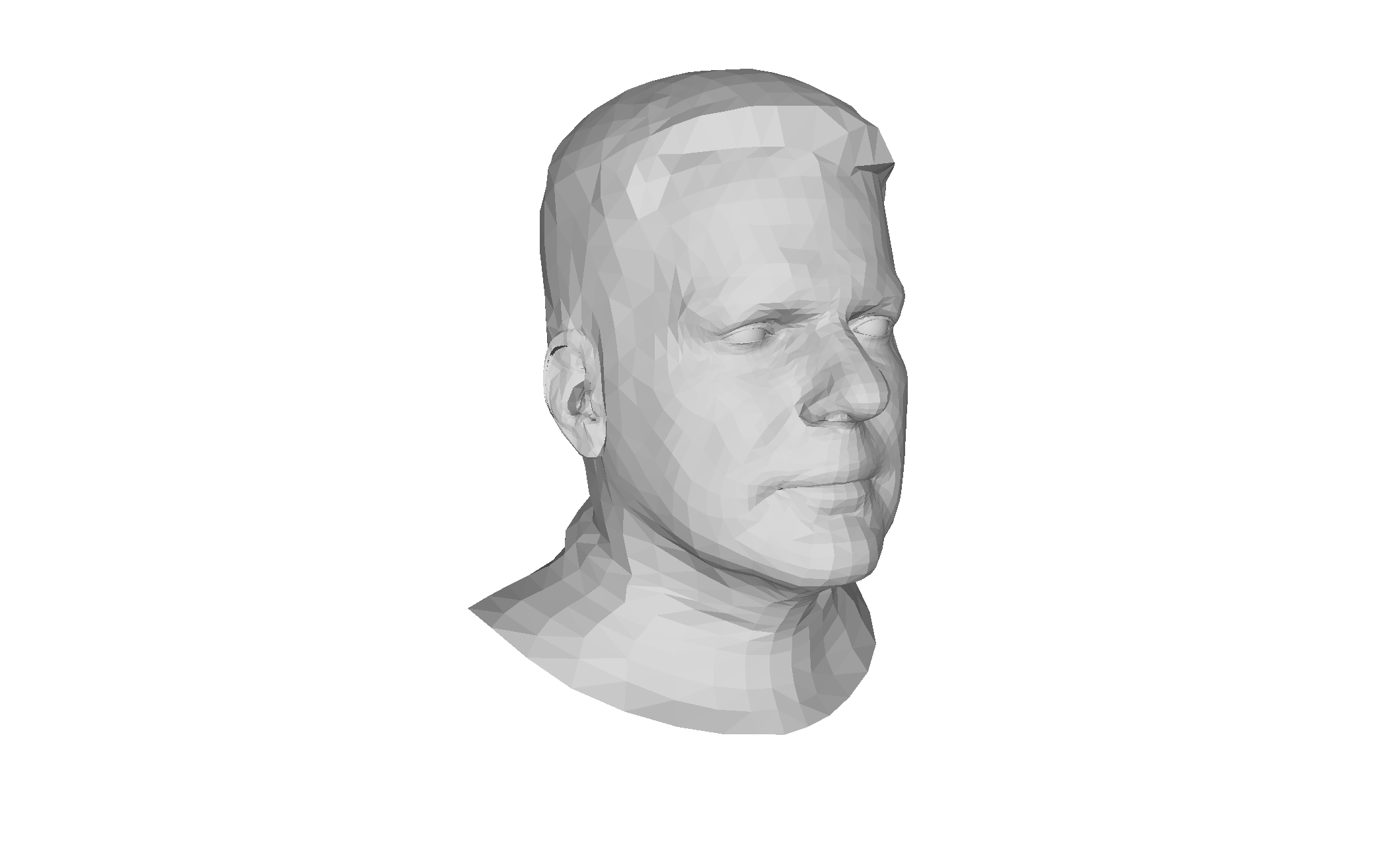}
    } 
    \subfloat
    {
        \includegraphics[trim={15cm 2cm 19cm 2cm},clip,width=.20\linewidth]{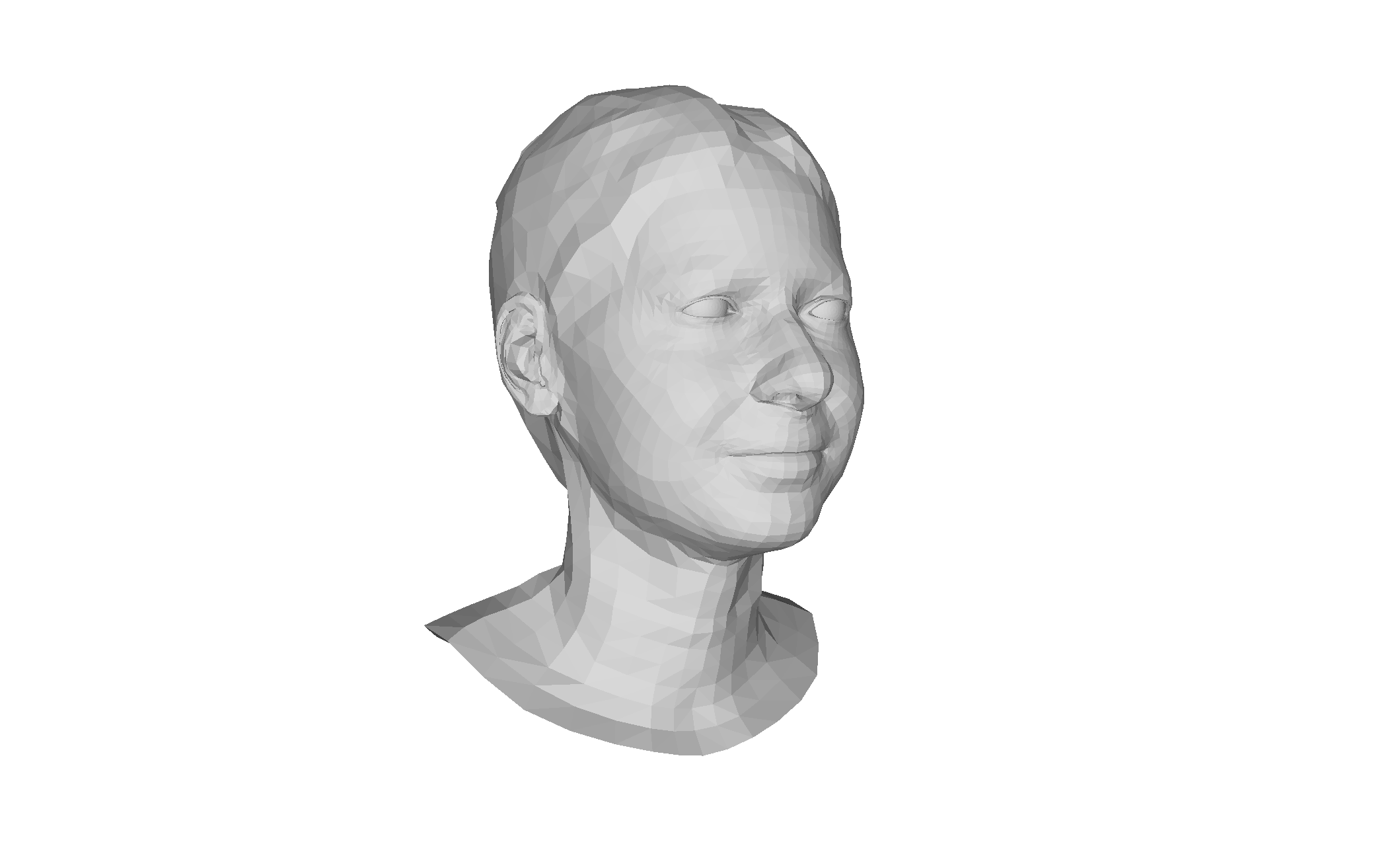}
    } 
    \subfloat
    {
        \includegraphics[trim={17cm 2.2cm 17cm 1.5cm},clip,width=.20\linewidth]{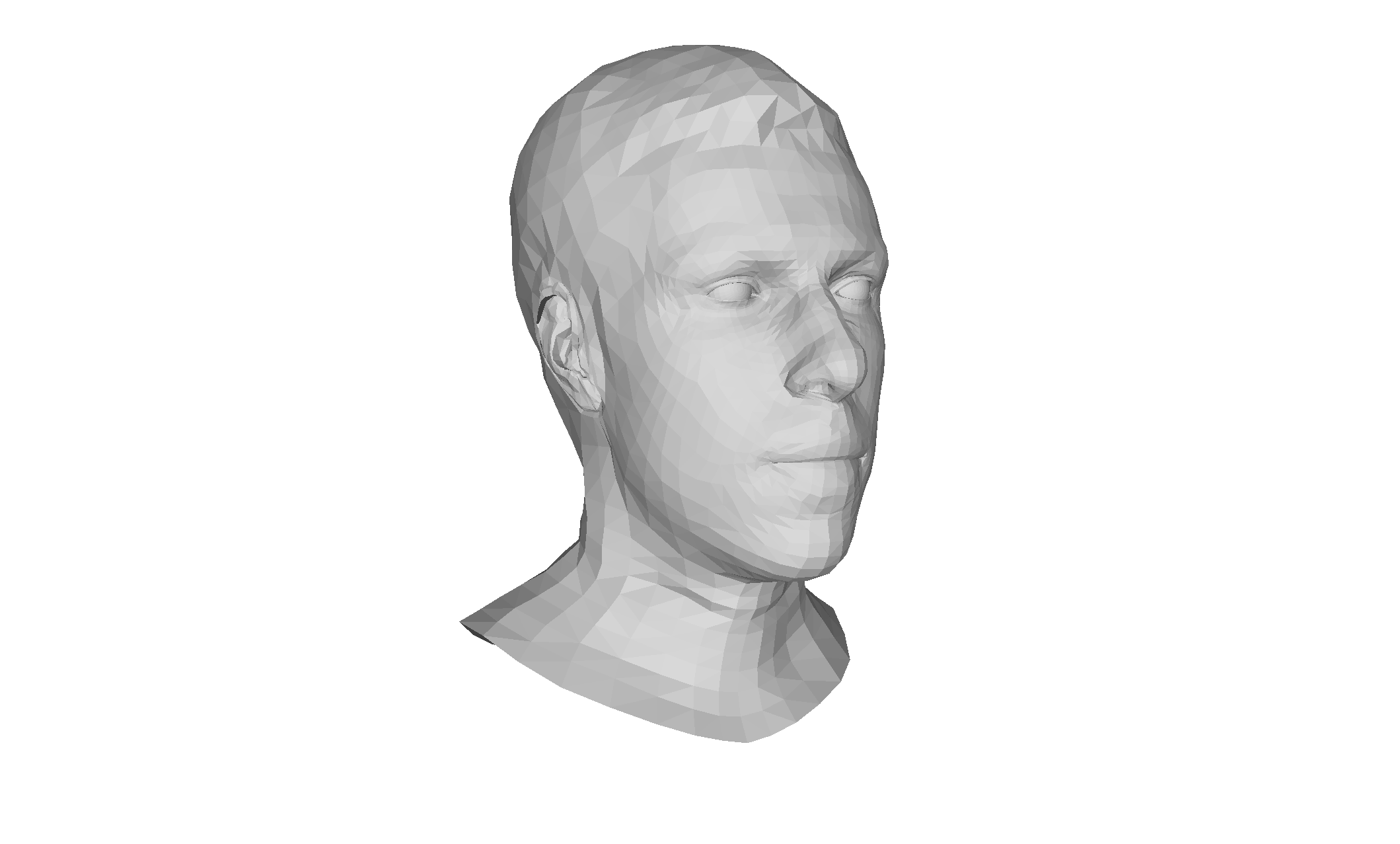}
    }
    \\
    \vspace*{-0.82em}
    \subfloat
    {
        \stackunder[5pt]{\includegraphics[trim={15cm 2cm 19cm 2cm},clip,width=.20\linewidth]{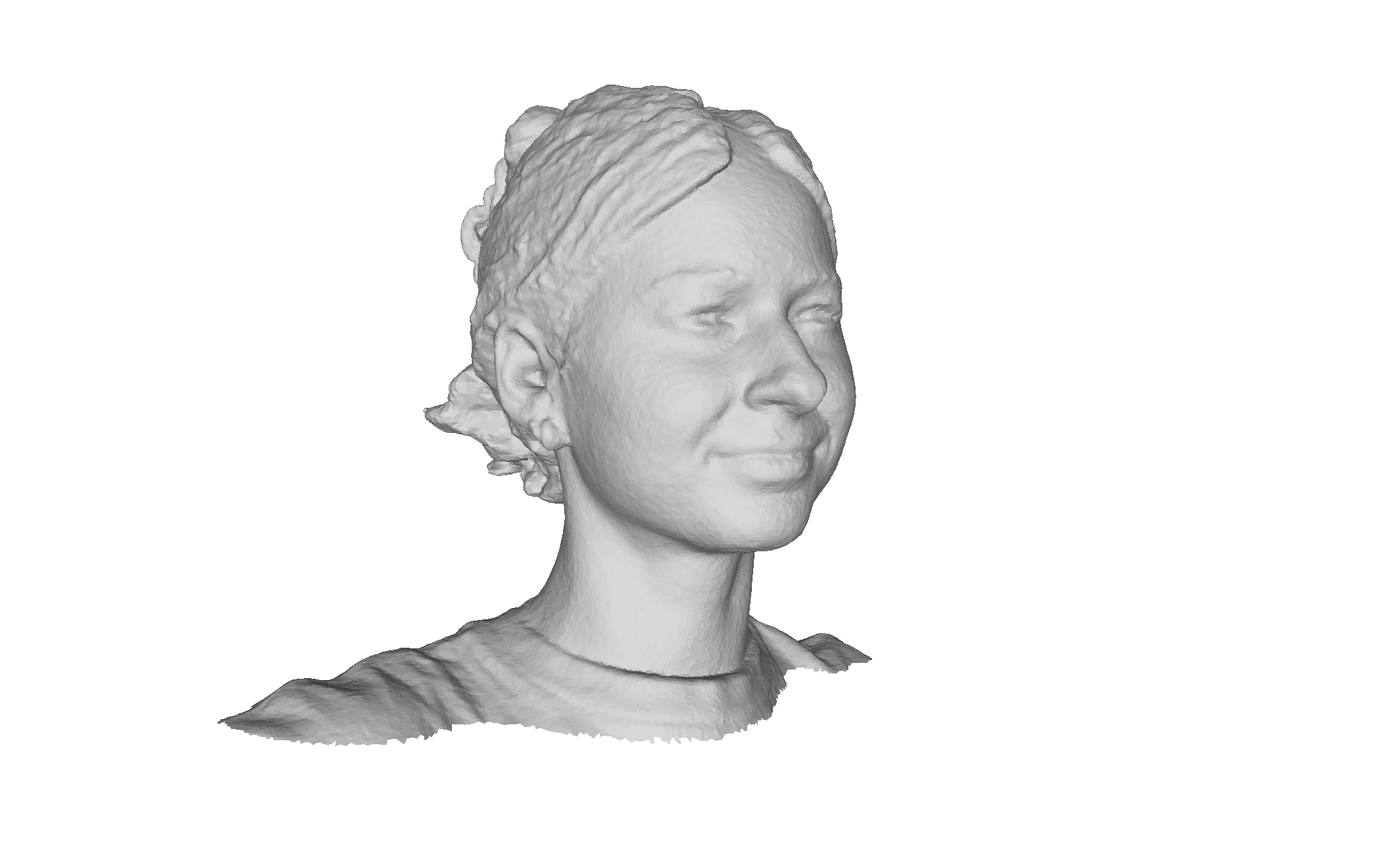}}{\small{(a) $S1$}}
    } 
    \subfloat
    {
        \stackunder[5pt]{\includegraphics[trim={17cm 3cm 17cm 2cm},clip,width=.20\linewidth]{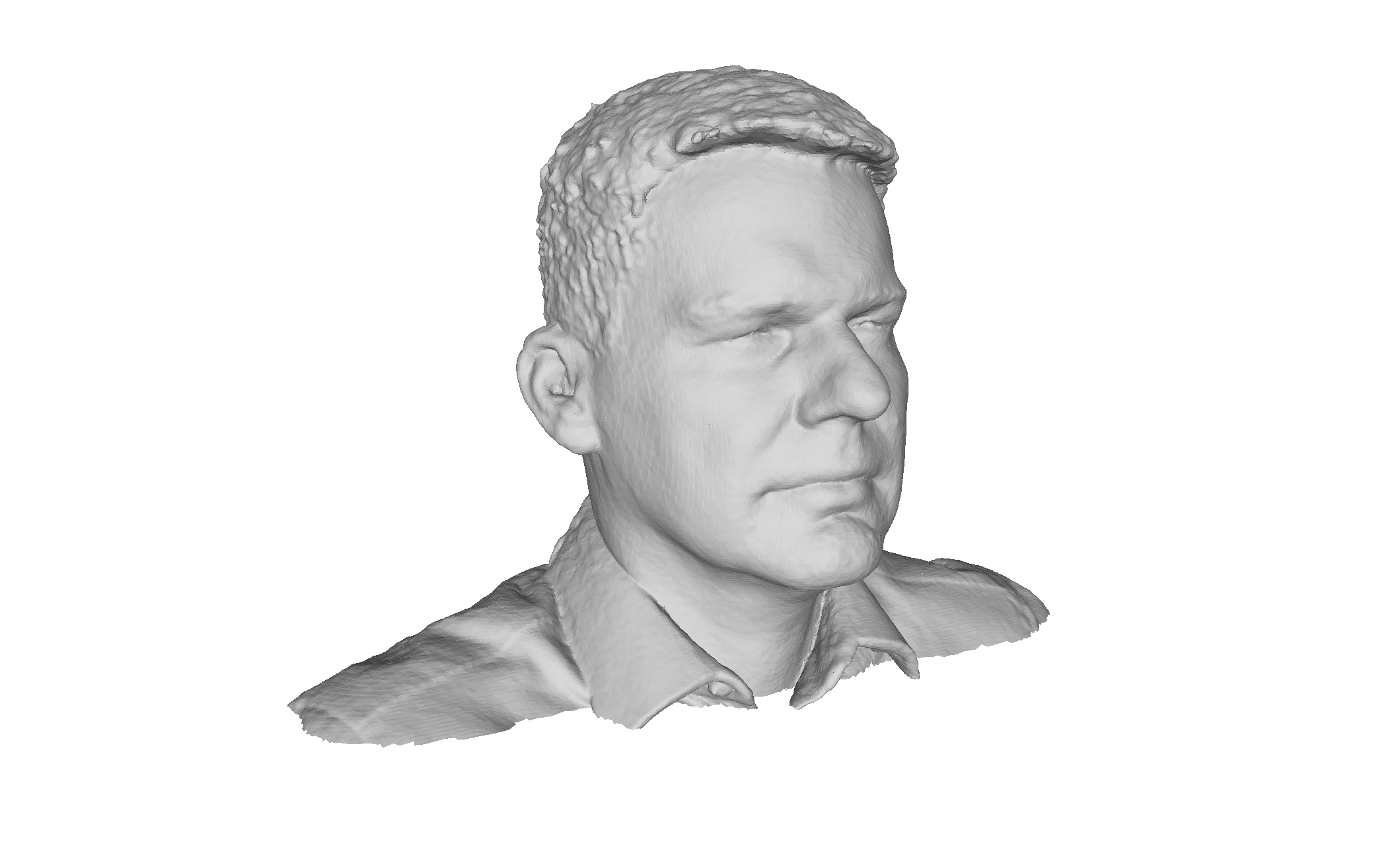}}{\small{(b) $S2$}}
    } 
    \subfloat
    {
        \stackunder[5pt]{\includegraphics[trim={15cm 2cm 19cm 2cm},clip,width=.20\linewidth]{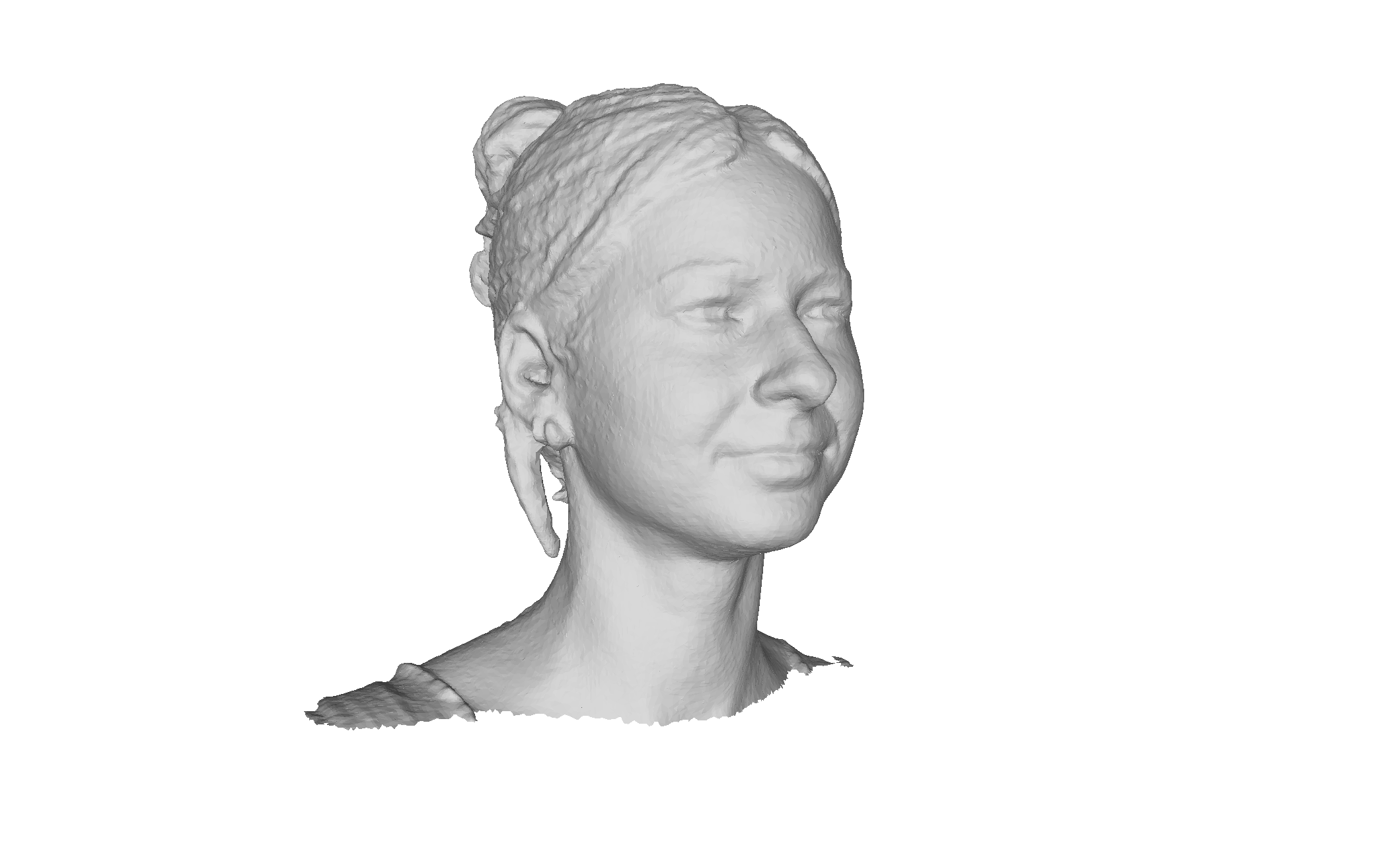}}{\small{(c) $S3$}}
    } 
    \subfloat
    {
        \stackunder[5pt]{\includegraphics[trim={17cm 2.2cm 17cm 1.5cm},clip,width=.20\linewidth]{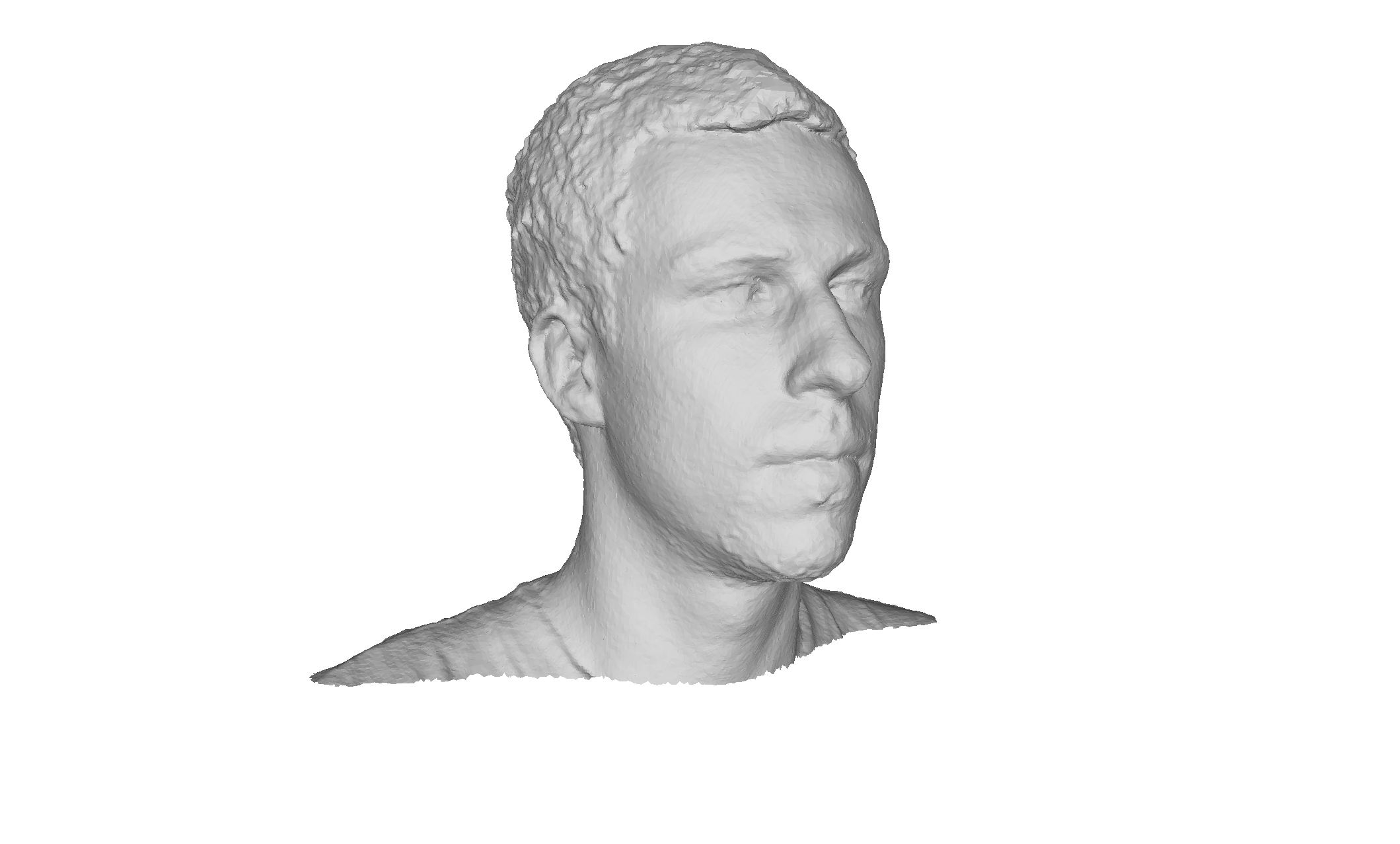}}{\small{(d) $S4$}}
    }
    \vspace{-8pt}
    \caption{Visualization of our personalized head avatar $H$ for all subjects. The top row presents the person-specific head avatars in a neutral expression, while the bottom row displays the corresponding target head scans used for fitting as a comparison.}
    \label{fig:results1}
\end{figure}

\begin{figure}[t]
    \centering
    \hspace*{-1em}
    \subfloat
    {
        \includegraphics[trim={15cm 0cm 15cm 0cm},clip,width=.19\linewidth]{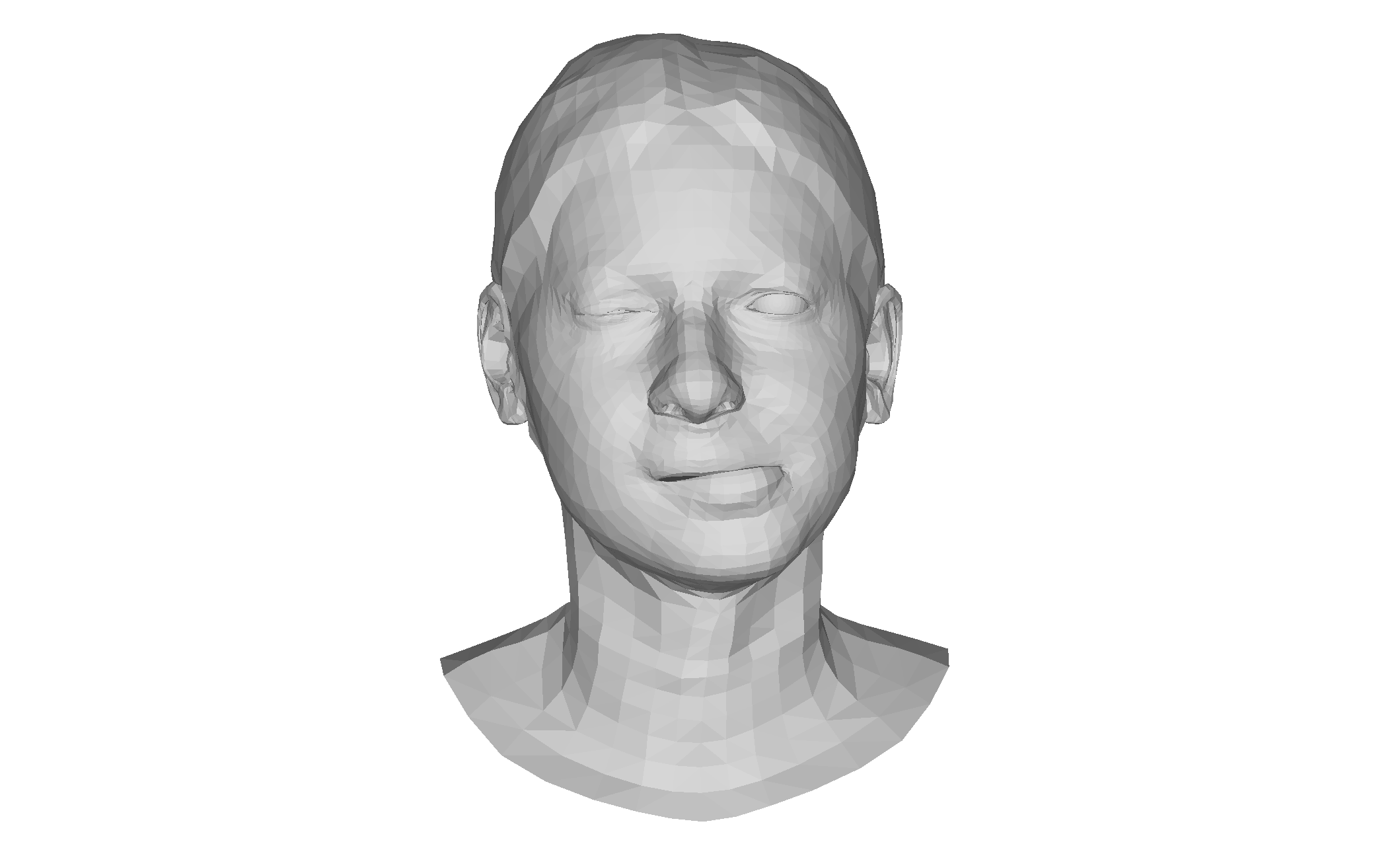}
    } 
    \subfloat
    {
        \includegraphics[trim={15cm 0cm 15cm 0cm},clip,width=.19\linewidth]{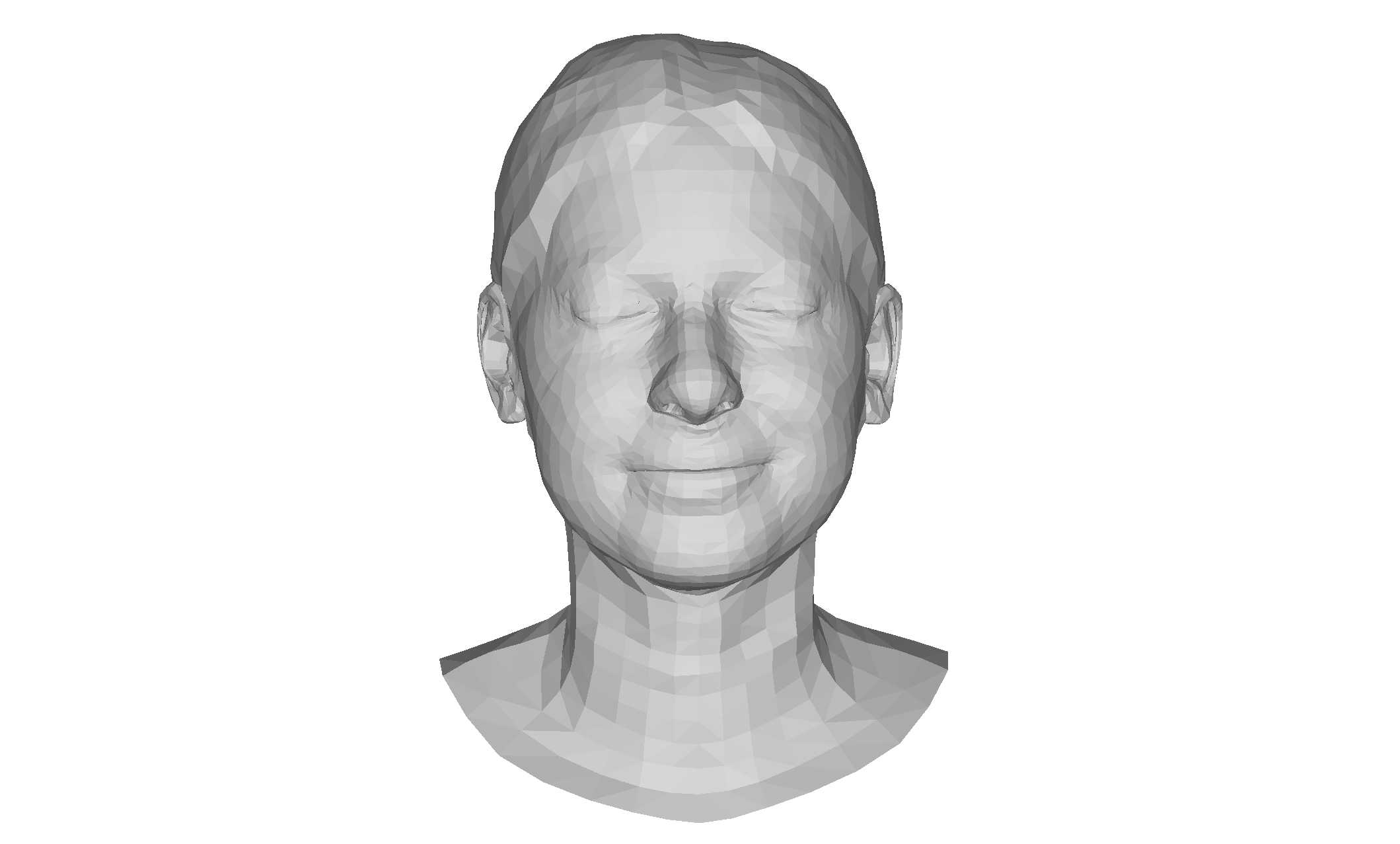}
    } 
    \subfloat
    {
        \includegraphics[trim={15cm 0cm 15cm 0cm},clip,width=.19\linewidth]{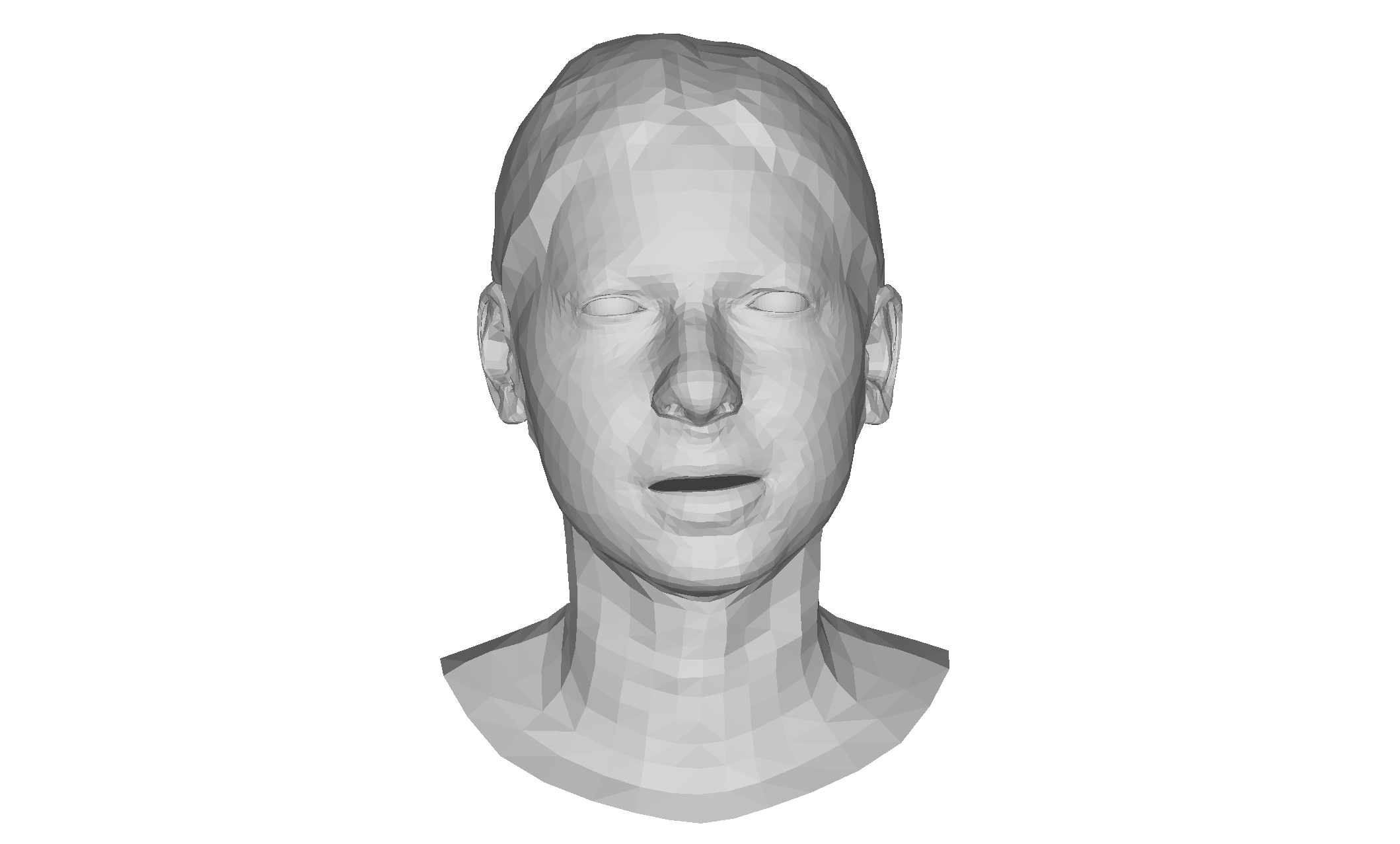}
    } 
    \subfloat
    {
        \includegraphics[trim={15cm 0cm 15cm 0cm},clip,width=.19\linewidth]{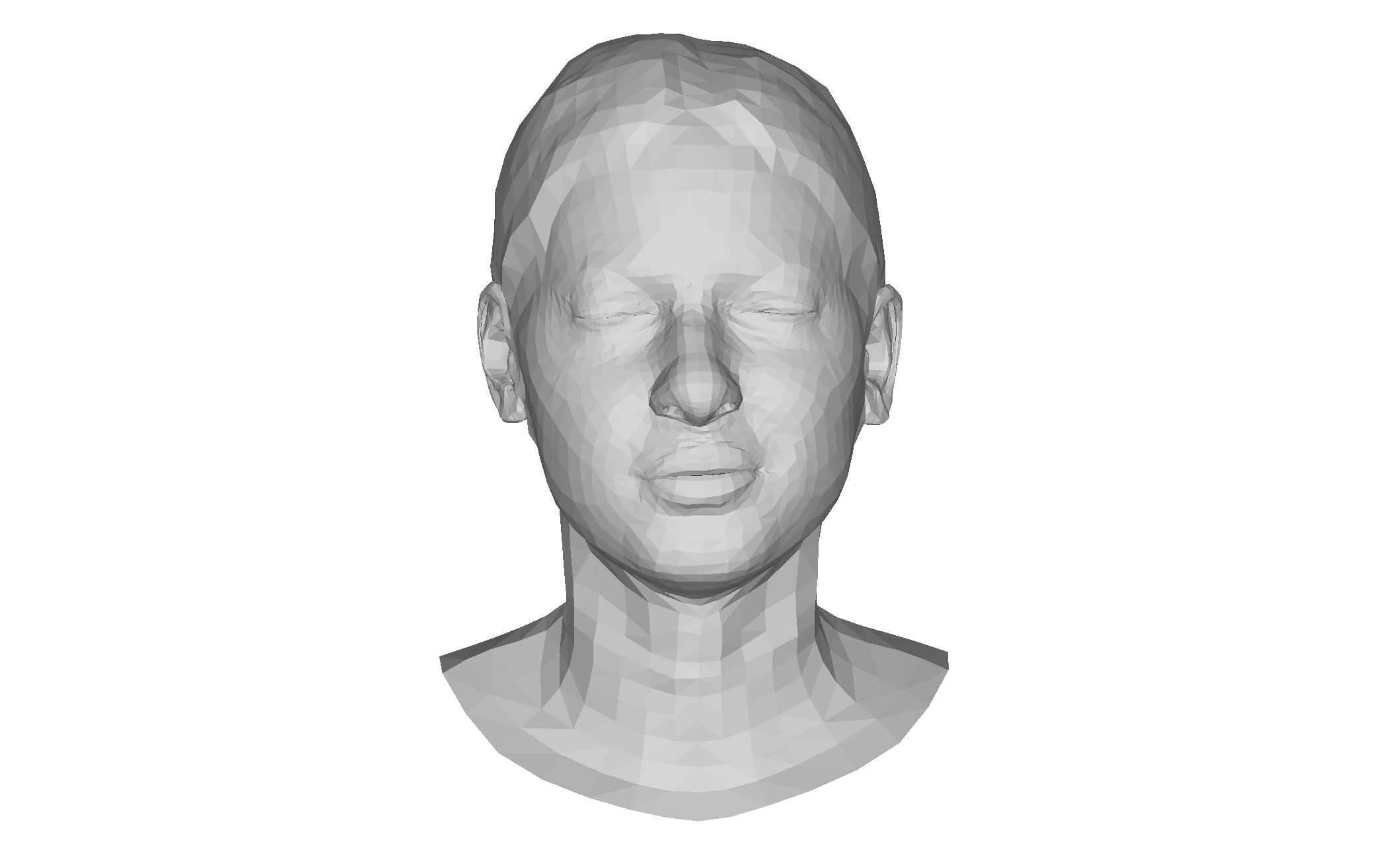}
    }
    \\
    \vspace*{-0.82em}
    \subfloat
    {
        \includegraphics[trim={15cm 0cm 15cm 0cm},clip,width=.19\linewidth]{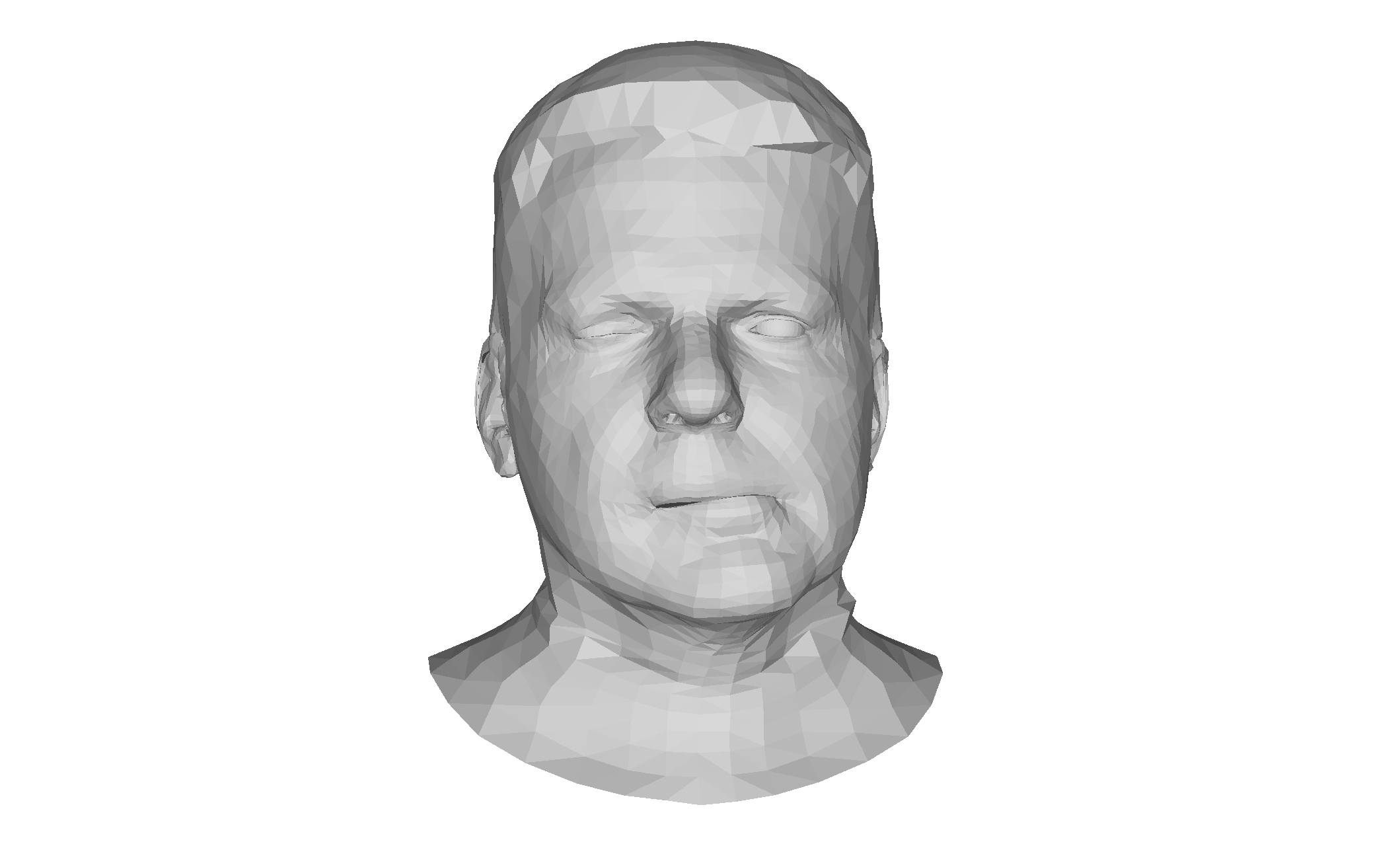}
    } 
    \subfloat
    {
        \includegraphics[trim={15cm 0cm 15cm 0cm},clip,width=.19\linewidth]{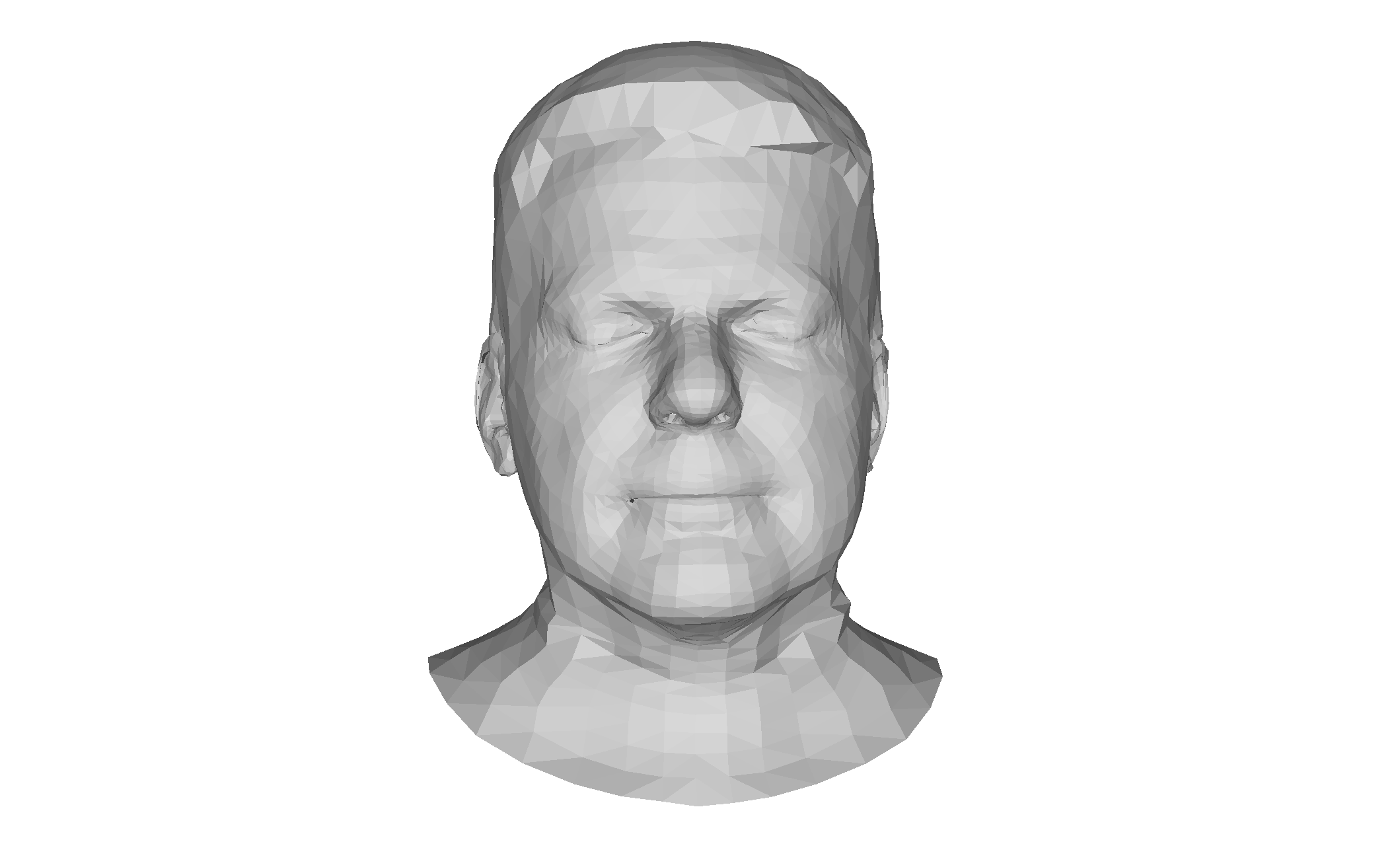}
    } 
    \subfloat
    {
        \includegraphics[trim={15cm 0cm 15cm 0cm},clip,width=.19\linewidth]{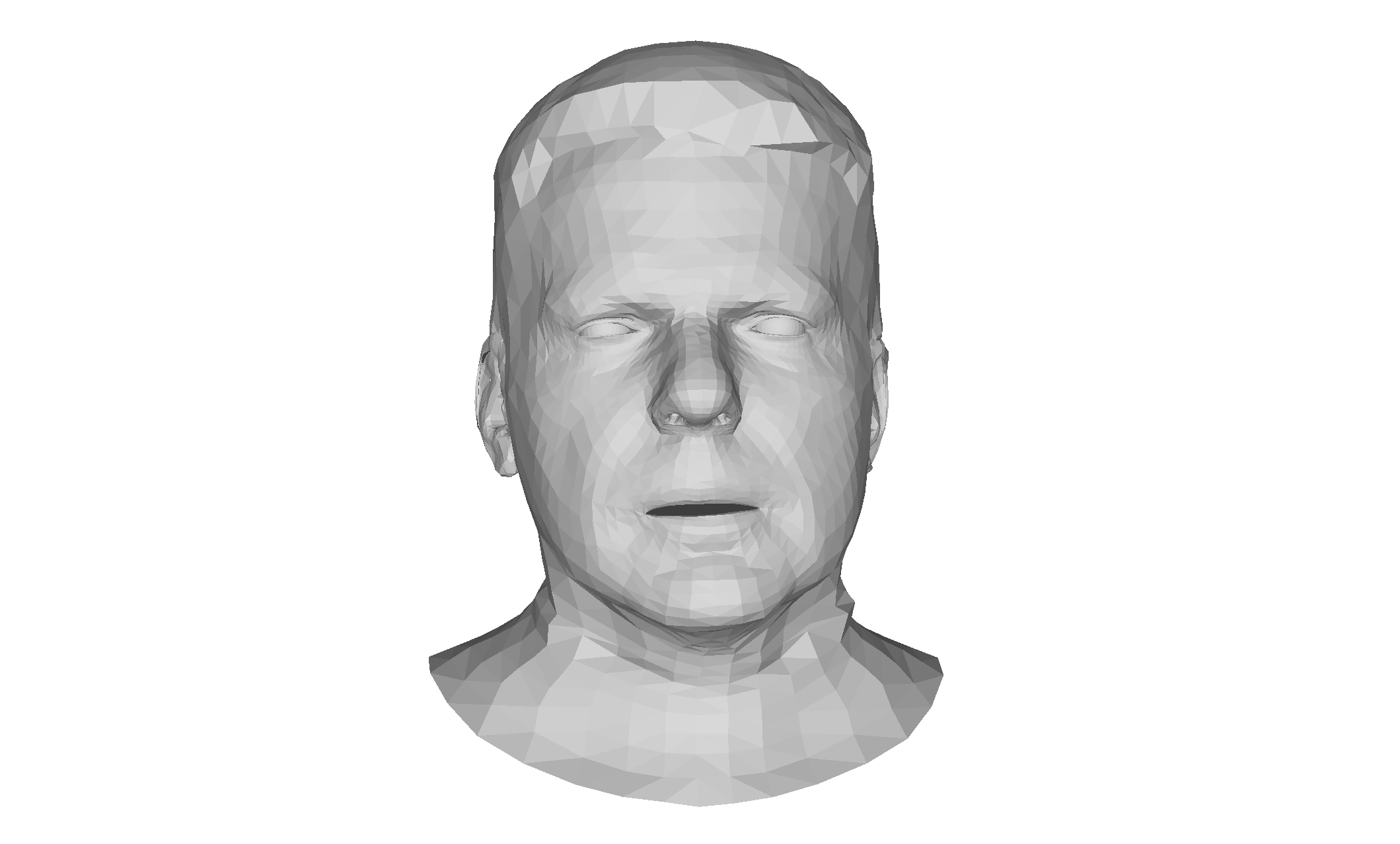}
    } 
    \subfloat
    {
        \includegraphics[trim={15cm 0cm 15cm 0cm},clip,width=.19\linewidth]{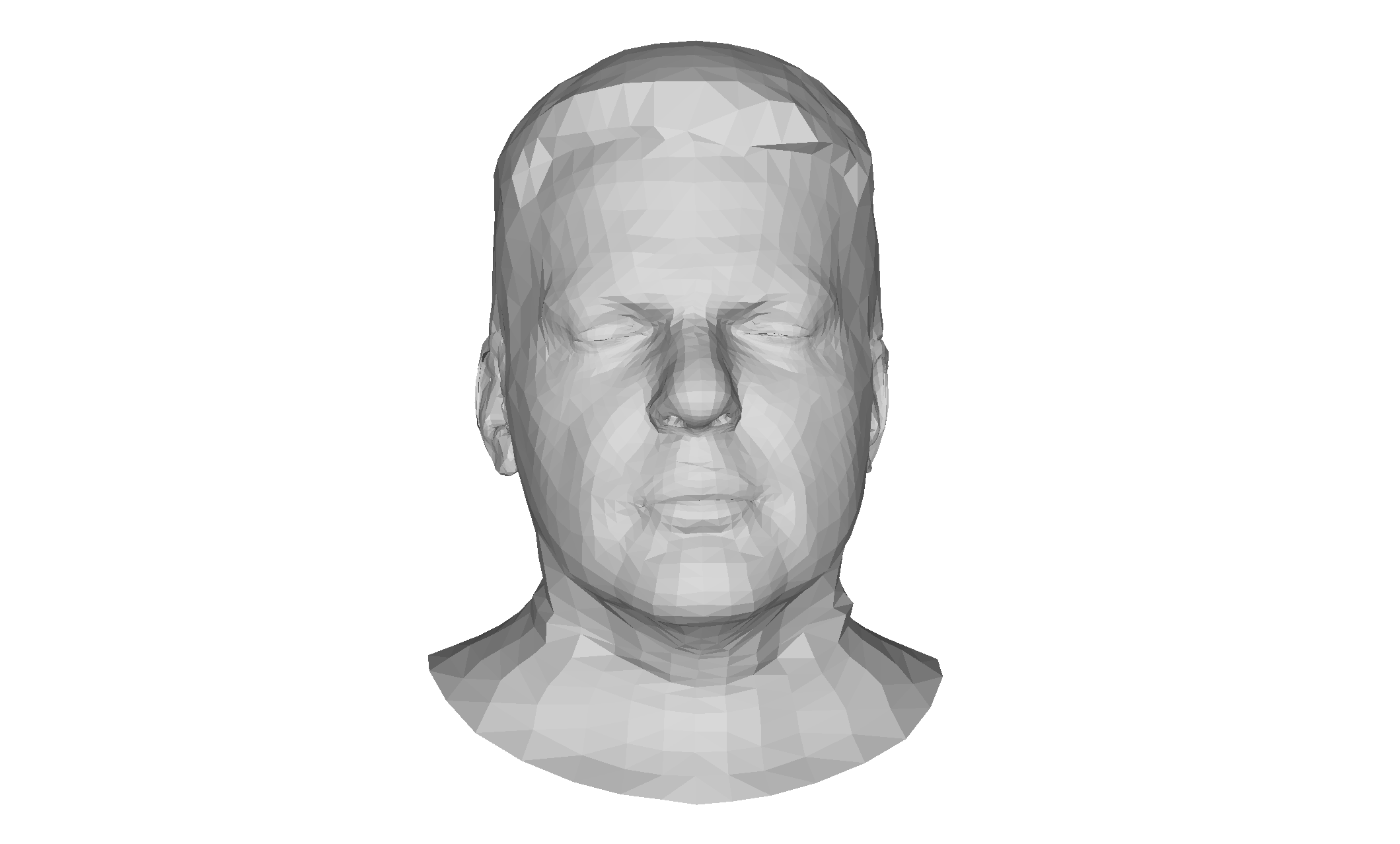}
    }
    \\
    \vspace*{-0.82em}
    \subfloat
    {
        \includegraphics[trim={15cm 0cm 15cm 0cm},clip,width=.19\linewidth]{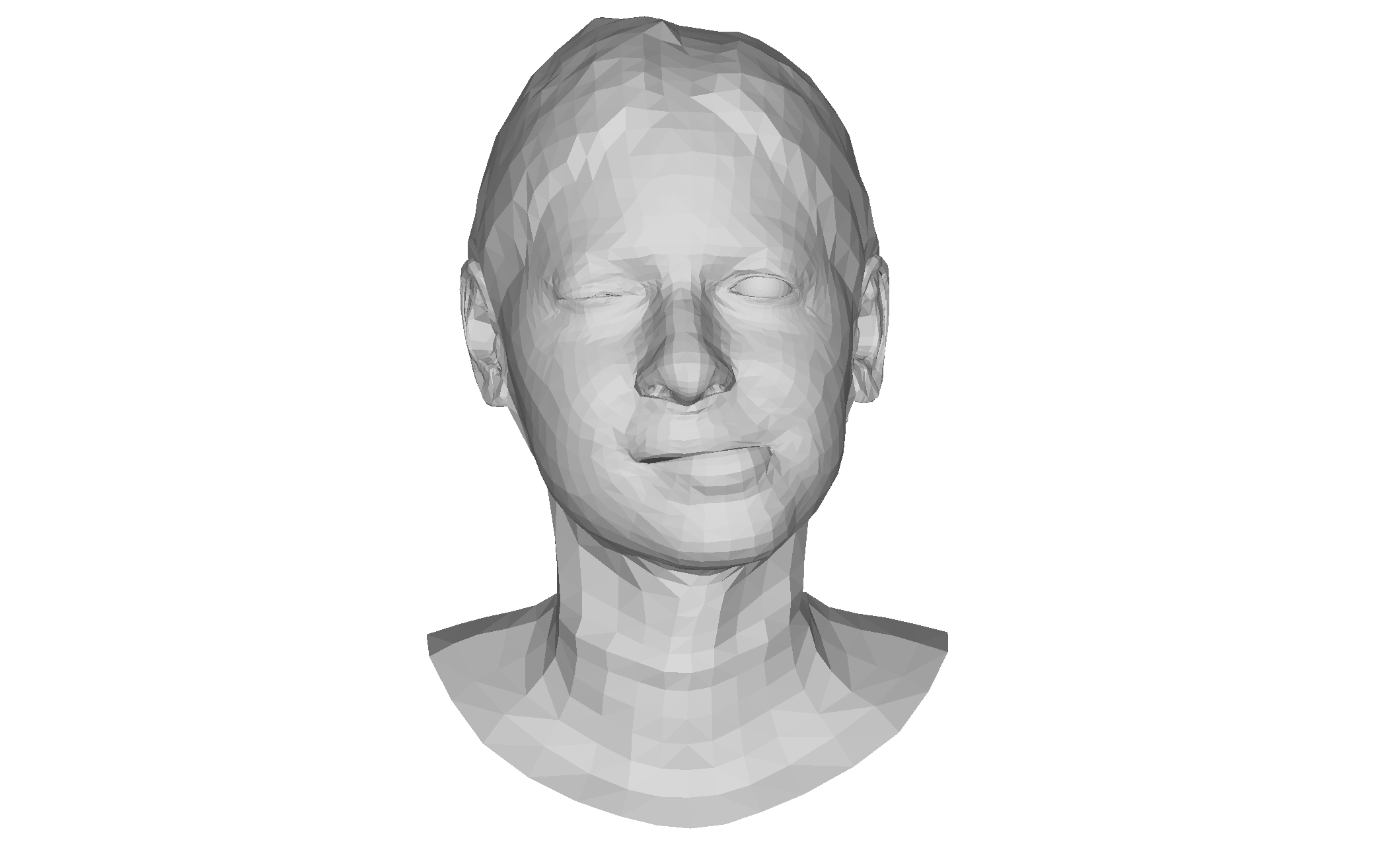}
    } 
    \subfloat
    {
        \includegraphics[trim={15cm 0cm 15cm 0cm},clip,width=.19\linewidth]{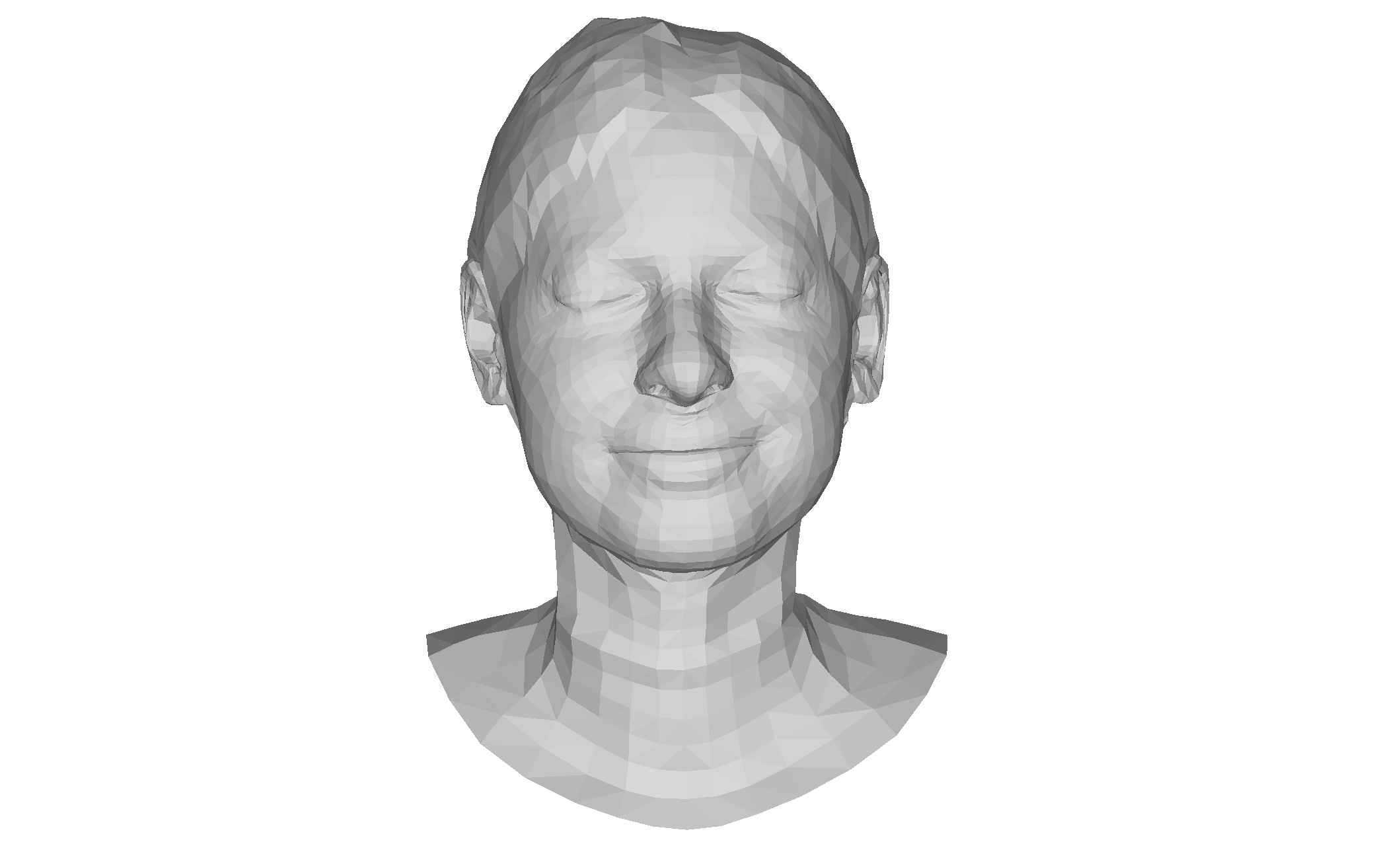}
    } 
    \subfloat
    {
        \includegraphics[trim={15cm 0cm 15cm 0cm},clip,width=.19\linewidth]{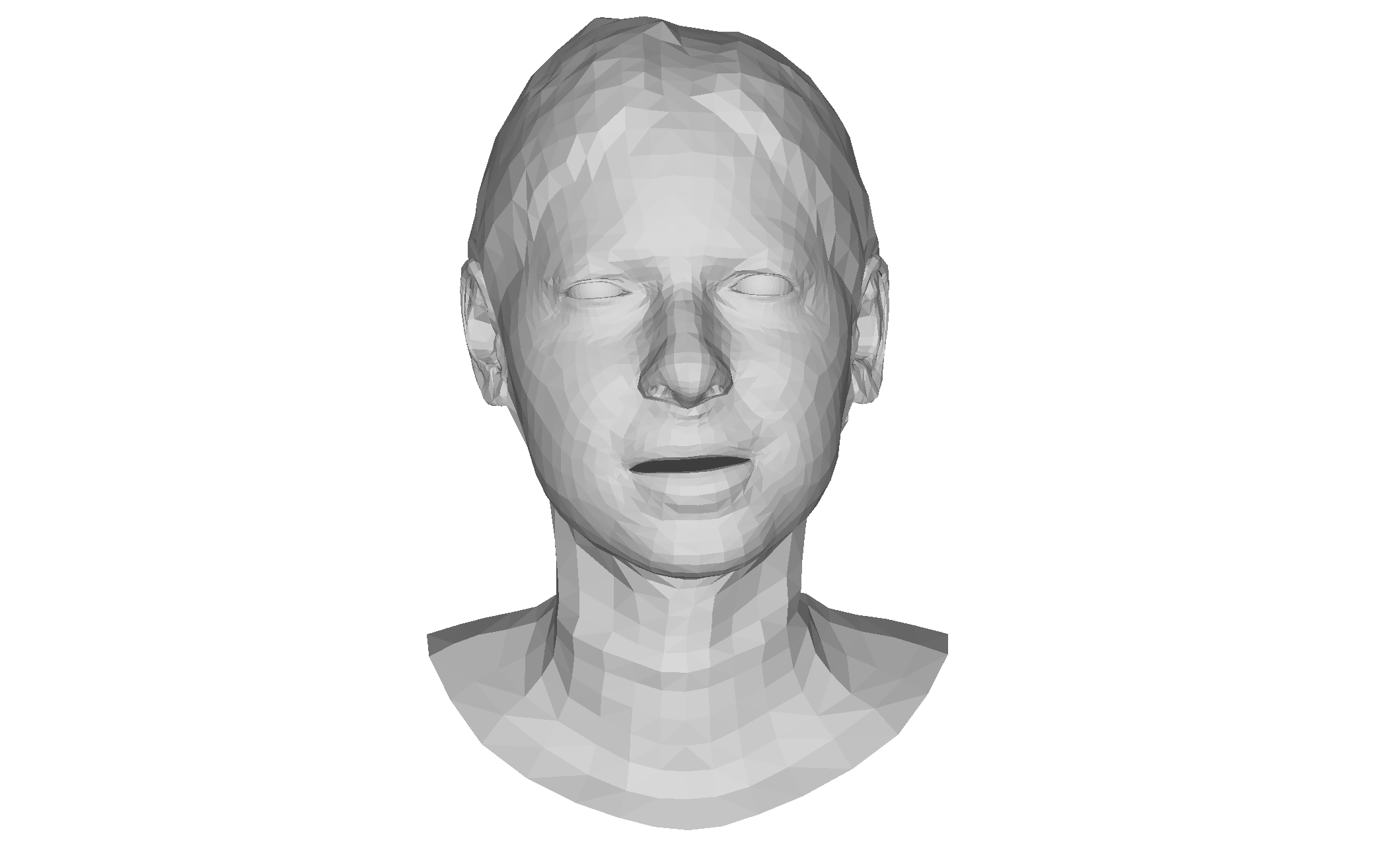}
    } 
    \subfloat
    {
        \includegraphics[trim={15cm 0cm 15cm 0cm},clip,width=.19\linewidth]{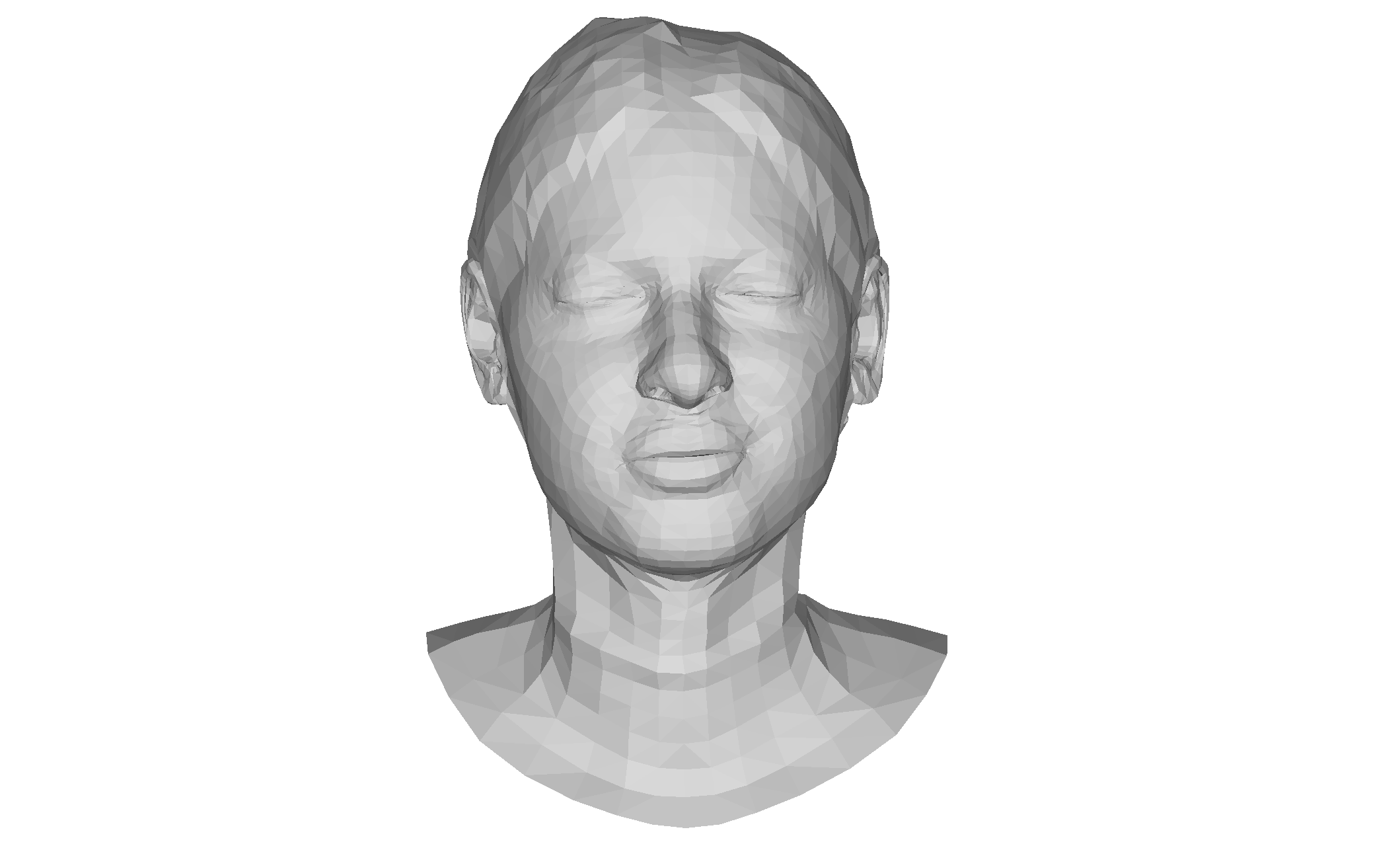}
    }
    \\
    \vspace*{-0.82em}
    \subfloat
    {
        \stackunder[5pt]{\includegraphics[trim={15cm 0cm 15cm 0cm},clip,width=.19\linewidth]{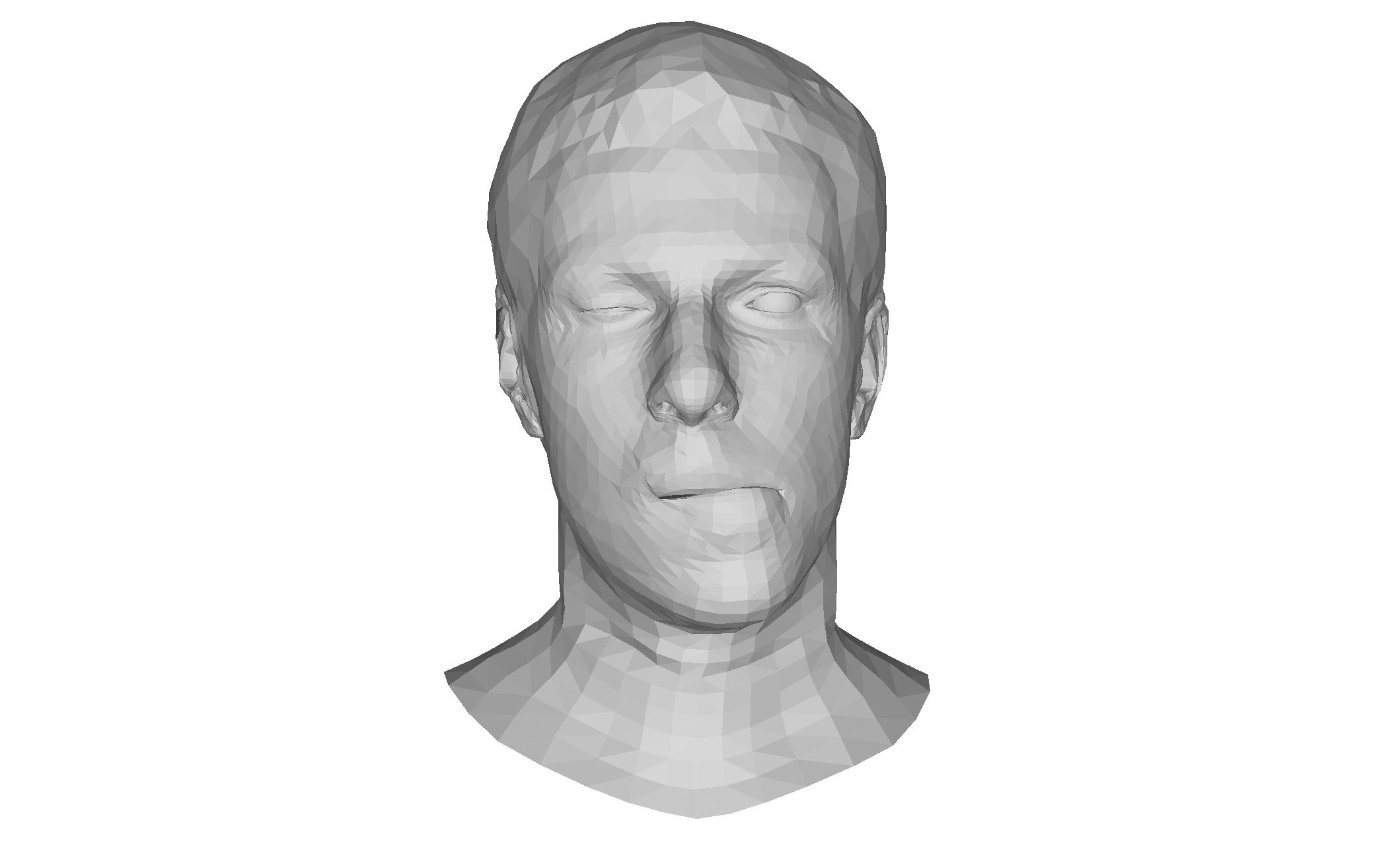}}{\small{(a)} }
    } 
    \subfloat
    {
        \stackunder[5pt]{\includegraphics[trim={15cm 0cm 15cm 0cm},clip,width=.19\linewidth]{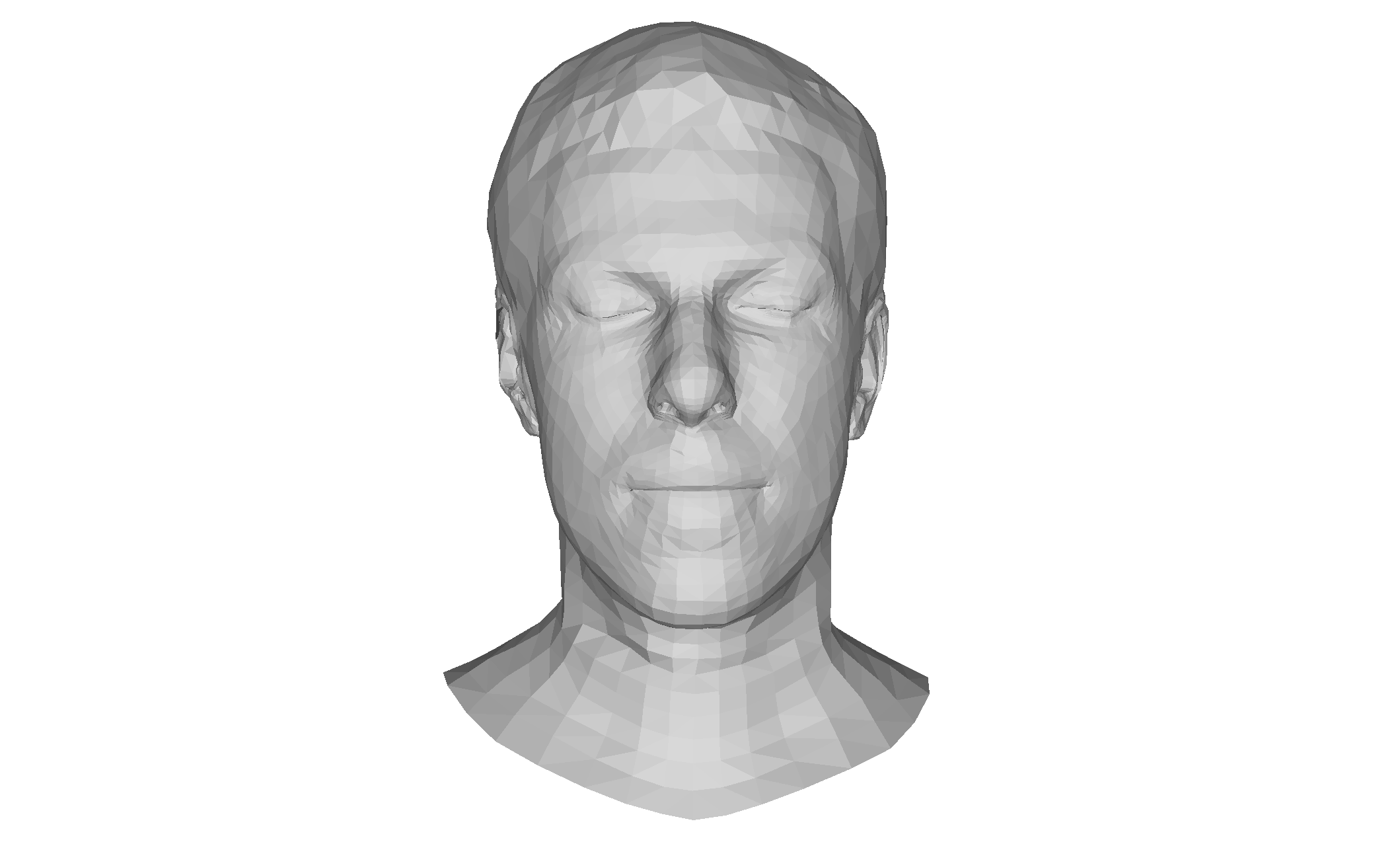}}{\small{(b)}}
    } 
    \subfloat
    {
        \stackunder[5pt]{\includegraphics[trim={15cm 0cm 15cm 0cm},clip,width=.19\linewidth]{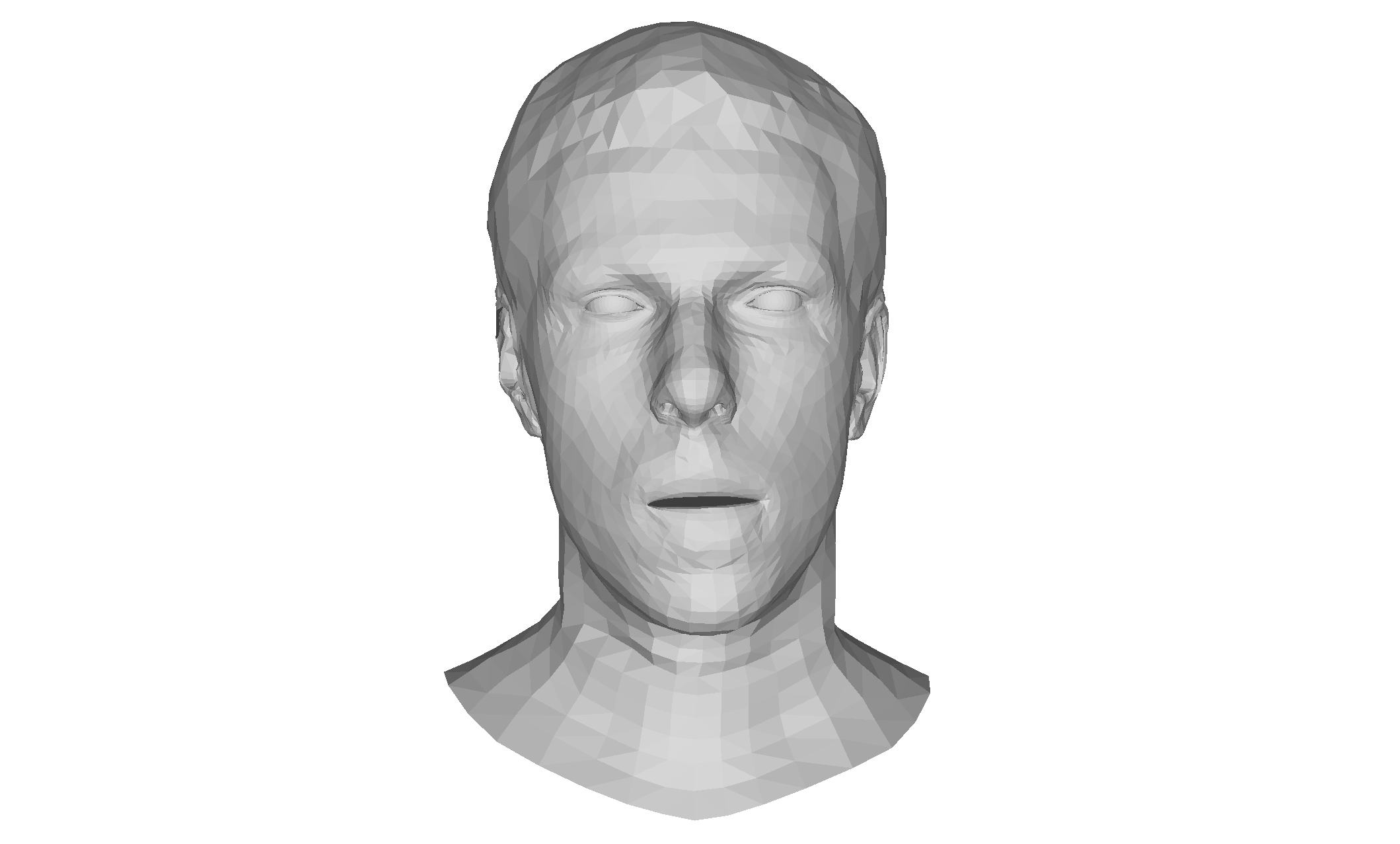}}{\small{(c)}}
    } 
    \subfloat
    {
        \stackunder[5pt]{\includegraphics[trim={15cm 0cm 15cm 0cm},clip,width=.19\linewidth]{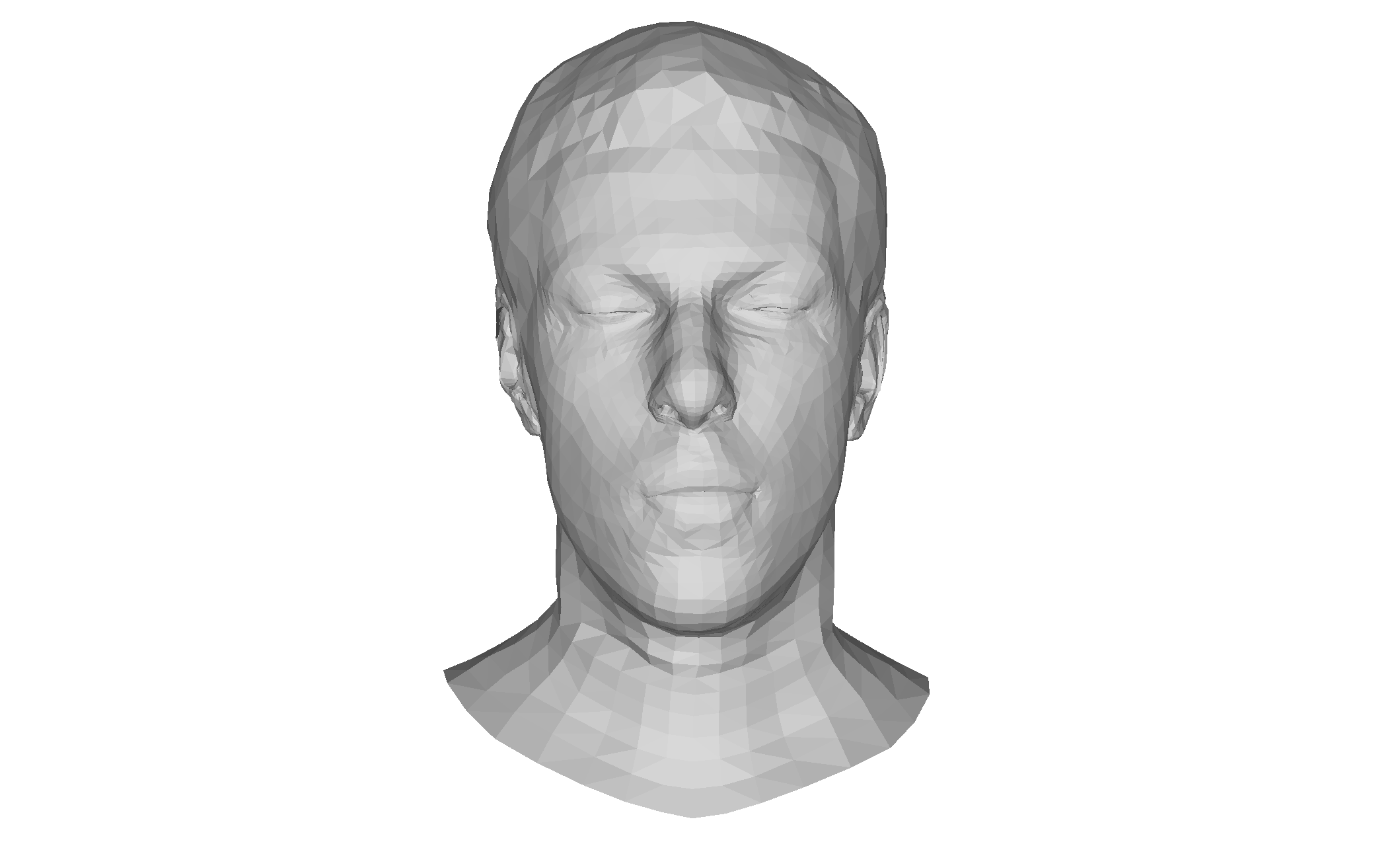}}{\small{(d)}}
    }
    \vspace{-8pt}
    \caption{Personalized head avatars $H$ for all subjects, showcasing four randomly sampled facial expressions.}
    \label{fig:results2}
\end{figure}

%
%%%%%%%%%%%%%%%%%%%%%%%%%%%%%%%%%%%%%%%%%%%%%%%%%%%%%%%%%%%%%%%%%%%%%%%%%%%%%%%%%%%%%%%%%%%%%%
%
Fig.~\ref{fig:results1} presents the finalized personalized head avatars $H$ for each subject, highlighting our pipeline's ability to accurately model subject-specific facial geometry. A notable limitation, however, is the inability to fully capture hair details. This is due to the underlying 3D morphable face model, FLAME \citep{FLAME:SiggraphAsia2017}, which does not account for hair.
Fig.~\ref{fig:results2} further demonstrates the head avatars with a variety of randomly sampled expressions, showcasing both the expressiveness of the avatars and the transferability of expressions across different subjects.
%
%%%%%%%%%%%%%%%%%%%%%%%%%%%%%%%%%%%%%%%%%%%%%%%%%%%%%%%%%%%%%%%%%%%%%%%%%%%%%%%%%%%%%%%%%%%%%%
%
%
%%%%%%%%%%%%%%%%%%%%%%%%%%%%%%%%%%%%%%%%%%%%%%%%%%%%%%%%%%%%%%%%%%%%%%%%%%%%%%%%%%%%%%%%%%%%%%
%
%
%%%%%%%%%%%%%%%%%%%%%%%%%%%%%%%%%%%%%%%%%%%%%%%%%%%%%%%%%%%%%%%%%%%%%%%%%%%%%%%%%%%%%%%%%%%%%%
%
\subsection{Learned Deformable Character}
\begin{figure}[t]
    \centering
    \includegraphics[width=0.8\linewidth]{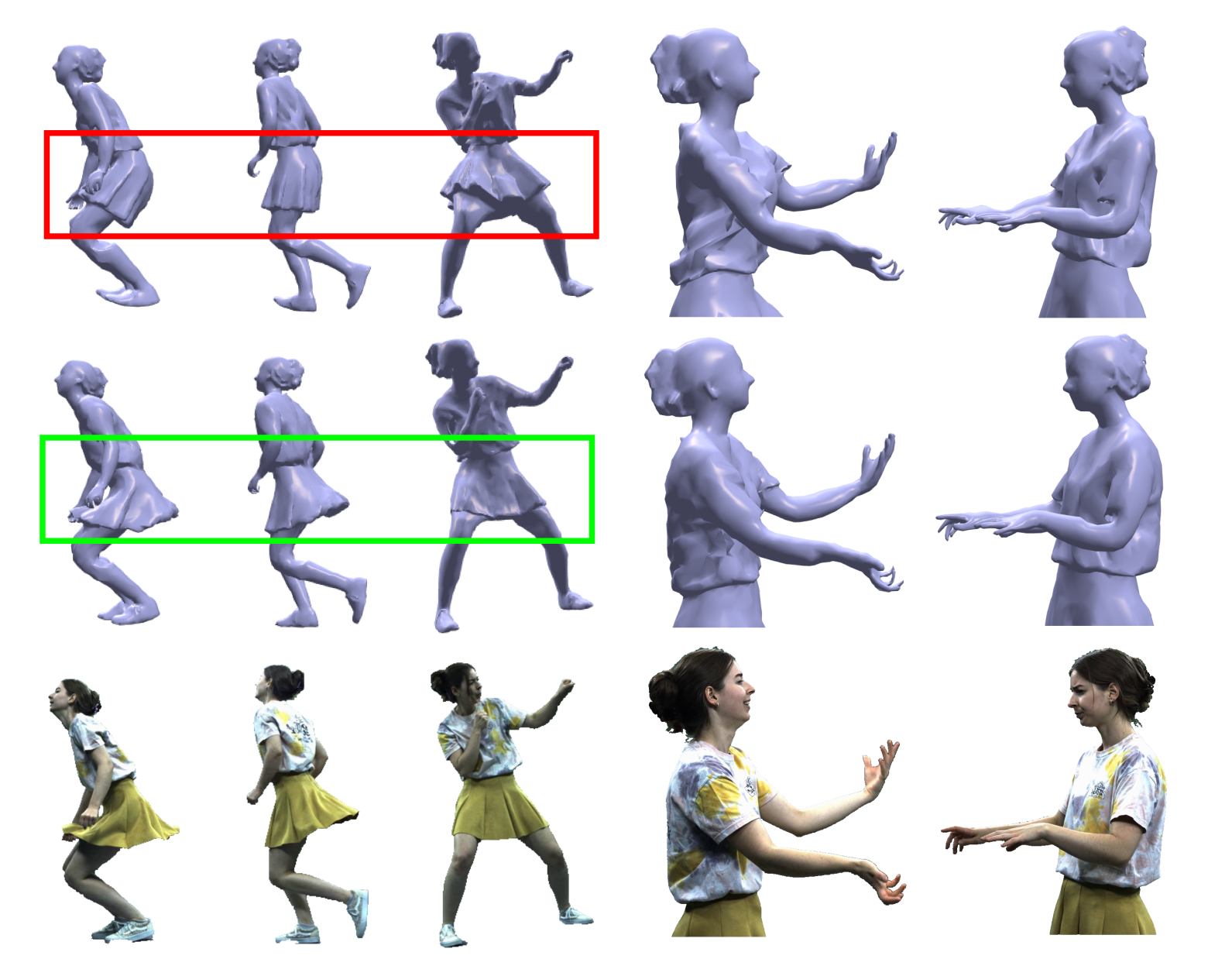}
    \vspace{-8pt}
    \caption{Leveraging DDC \citep{habermann2021DDC} as an underlying character deformation model enables precise simulation of loose clothing dynamics and effectively mitigates potential skinning artifacts, resulting in more realistic character animations.}
    \label{fig:results5}
\end{figure}

%
%%%%%%%%%%%%%%%%%%%%%%%%%%%%%%%%%%%%%%%%%%%%%%%%%%%%%%%%%%%%%%%%%%%%%%%%%%%%%%%%%%%%%%%%%%%%%%
%
Fig.~\ref{fig:results5} illustrates the capability of the learned deformable character model to predict physically plausible clothing deformations. This is particularly evident for subject $S1$, where the model excels in simulating the realistic swinging of the skirt. 
%
%%%%%%%%%%%%%%%%%%%%%%%%%%%%%%%%%%%%%%%%%%%%%%%%%%%%%%%%%%%%%%%%%%%%%%%%%%%%%%%%%%%%%%%%%%%%%%
%
%
%%%%%%%%%%%%%%%%%%%%%%%%%%%%%%%%%%%%%%%%%%%%%%%%%%%%%%%%%%%%%%%%%%%%%%%%%%%%%%%%%%%%%%%%%%%%%%
%
\subsection{Audio-driven Avatar} 
\begin{figure}[t]
    \centering
    \includegraphics[width=1\linewidth]{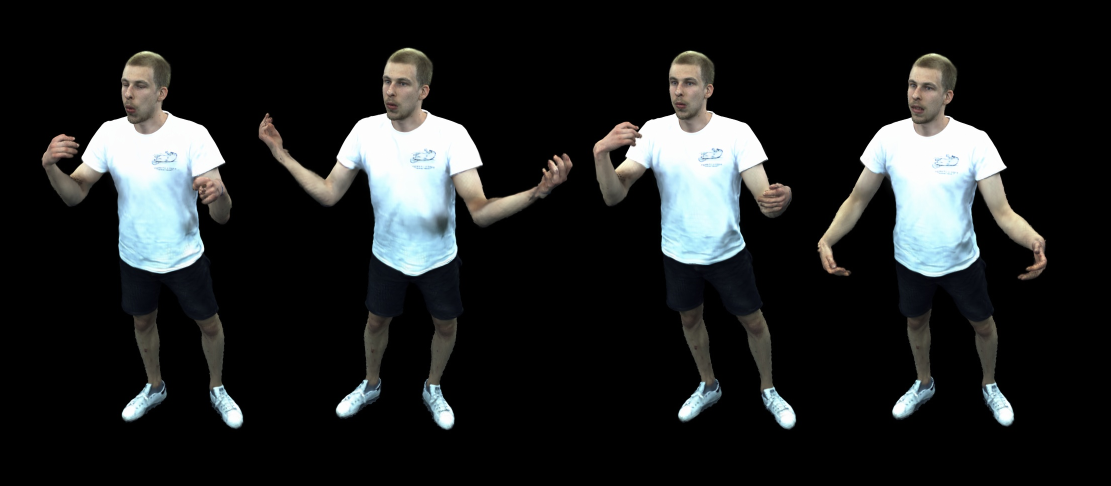}
    \vspace{-17pt}
    \caption{EVA uses skeletal motions and facial expressions generated from audio inputs \citep{mughal2024raggesture} to create realistic talking avatars, entirely driven by audio.}
    \label{fig:audiodriven}
\end{figure}

%
%%%%%%%%%%%%%%%%%%%%%%%%%%%%%%%%%%%%%%%%%%%%%%%%%%%%%%%%%%%%%%%%%%%%%%%%%%%%%%%%%%%%%%%%%%%%%%
%
Audio-driven avatars significantly enhance virtual education and sales by synthesizing motions and expressions directly from audio inputs, as shown in Fig.~\ref{fig:audiodriven}. We leverage the method from \citet{mughal2024raggesture} to generate poses and expressions, respectively.
%
%%%%%%%%%%%%%%%%%%%%%%%%%%%%%%%%%%%%%%%%%%%%%%%%%%%%%%%%%%%%%%%%%%%%%%%%%%%%%%%%%%%%%%%%%%%%%%
%
%
%%%%%%%%%%%%%%%%%%%%%%%%%%%%%%%%%%%%%%%%%%%%%%%%%%%%%%%%%%%%%%%%%%%%%%%%%%%%%%%%%%%%%%%%%%%%%%
%
\section{Comparison with Animatable Gaussians} \label{sec:ag_comp}
\begin{figure}[bth]
    \centering
    \begin{minipage}{0.02\linewidth}
        \small
        \raggedright
        \rotatebox{90}{GT\ \ \ \ \ \ \ \ \ \ \ \ } \\[6.5em]
        \rotatebox{90}{\textbf{EVA (ours)}} \\[6.5em]
        \rotatebox{90}{\ \ \ \ \ \ \ \ \ \ \ \ AG}
    \end{minipage}
    \hfill
    \begin{minipage}{0.95\linewidth}
        \centering
        \includegraphics[width=\linewidth]{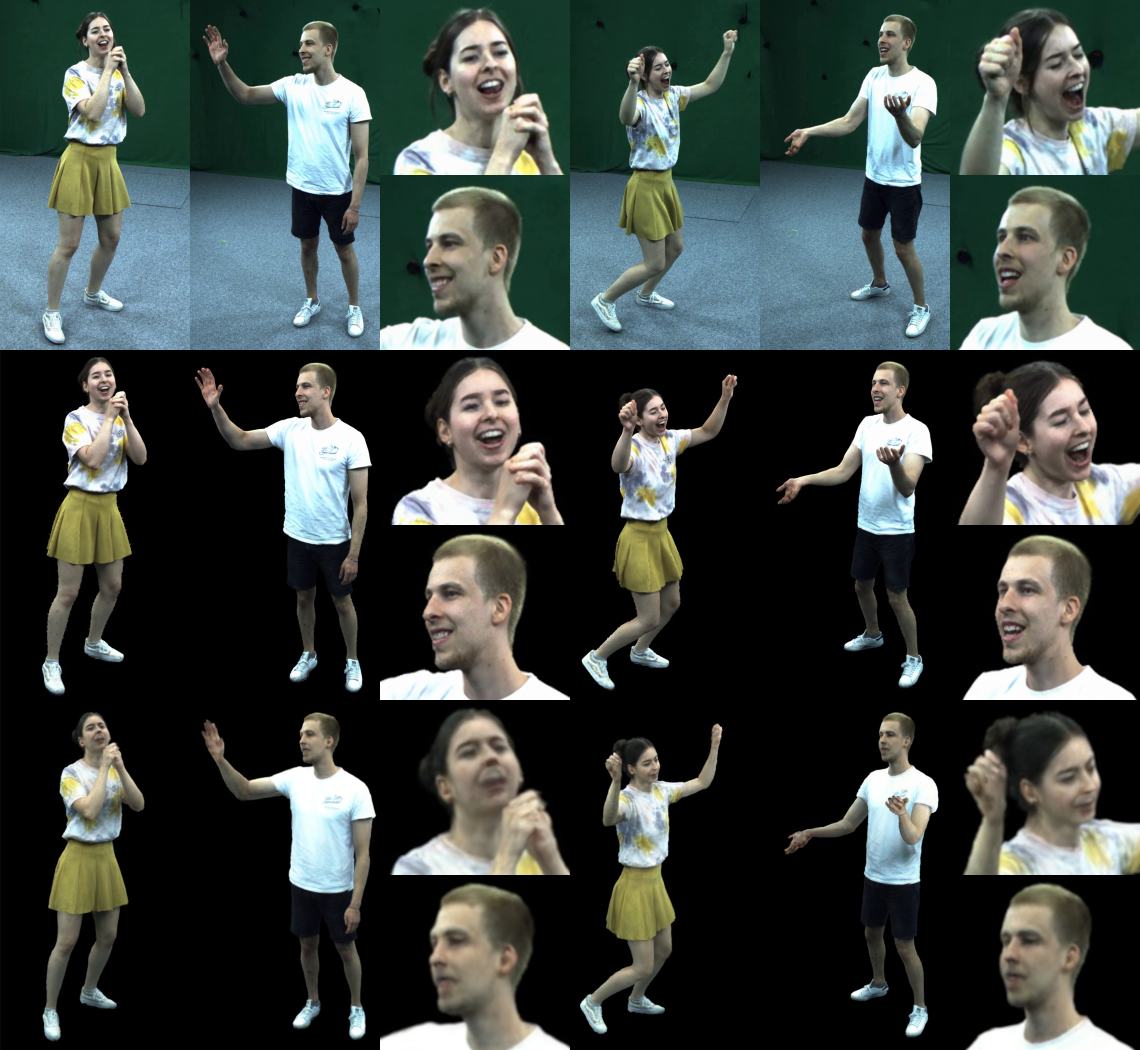}
    \end{minipage}

    \vspace{0.5em}
    \hspace*{0.02\linewidth} 
    \begin{minipage}{0.48\linewidth}
        \centering
        \small
        Novel View
    \end{minipage}
    \hfill
    \begin{minipage}{0.48\linewidth}
        \centering
        \small
        Novel View, Pose \& Expression
    \end{minipage}
    \vspace{-0.5em}
    \caption{Qualitative comparison of EVA with AnimatableGaussians (AG) \citep{li2024animatablegaussians}.}
    \label{fig:ag_comparison}
\end{figure}

We also compare our method with AnimatableGaussians (AG)~\citep{li2024animatablegaussians}, which estimates 3D Gaussians from pose-driven position map inputs. While this approach captures pose-dependent garment details, it does not offer explicit control over facial expressions and is unable to recover them at all. Additionally, its overall output tends to appear visually blurry.
As shown in Fig.~\ref{fig:ag_comparison}, our method produces \textbf{qualitatively} superior results. EVA not only reconstructs finer body details but also renders the head and hands with significantly improved sharpness and fidelity.
This advantage is further supported by \textbf{quantitative} evaluation. As presented in Tab.~\ref{tab:comparisons_AGGA}, EVA consistently outperforms AnimatableGaussians across multiple metrics and for novel view, pose, and expression synthesis (testing sequences from unseen viewpoints), as well as novel view rendering quality (training sequences from unseen viewpoints).
%
%%%%%%%%%%%%%%%%%%%%%%%%%%%%%%%%%%%%%%%%%%%%%%%%%%%%%%%%%%%%%%%%%%%%%%%%%%%%%%%%%%%%%%%%%%%%%%
%
%
%%%%%%%%%%%%%%%%%%%%%%%%%%%%%%%%%%%%%%%%%%%%%%%%%%%%%%%%%%%%%%%%%%%%%%%%%%%%%%%%%%%%%%%%%%%%%%
%
\begin{table}[t]
    \centering
    \small
    \setlength{\tabcolsep}{4pt} 
    \renewcommand{\arraystretch}{0.75} 
     \caption{Comparison results of EVA against ASH, DDC, AnimatableGaussians, and GaussianAvatars. We highlight the \textbf{best} and \underline{second-best} results.}
     \vspace{-3pt}
    \label{tab:comparisons_AGGA}
    \begin{tabular}{lcccccc}
        \toprule
        \textbf{} & \multicolumn{3}{c}{\textbf{Novel Pose/Expr.}} & \multicolumn{3}{c}{\textbf{Novel View}} \\
        \cmidrule(lr){2-4} \cmidrule(lr){5-7}
        & LPIPS $\downarrow$ & PSNR $\uparrow$ & SSIM $\uparrow$ & LPIPS $\downarrow$ & PSNR $\uparrow$ & SSIM $\uparrow$ \\
        \midrule
        ASH & 22.15 & \underline{45.78}  & \underline{98.75}  & 19.30  & 46.31  & \underline{99.09} \\
        DDC & 20.70 & 45.36  & 98.62 & 18.14  & \textbf{46.87}  & 99.05 \\
        AG & \underline{16.35} & 44.66 & 98.23  & \underline{15.90}  & 44.68 & 98.27 \\
        GA & \underline{16.41}  & 45.01 & 98.62  & \underline{13.03}  & 45.82 & 99.01\\
         \midrule
        \textbf{EVA} & \textbf{15.55}  & \textbf{46.00}  & \textbf{98.79}  & \textbf{12.55}  & \underline{46.39}  & \textbf{99.22} \\
        \bottomrule
    \end{tabular}
\end{table}
%
%%%%%%%%%%%%%%%%%%%%%%%%%%%%%%%%%%%%%%%%%%%%%%%%%%%%%%%%%%%%%%%%%%%%%%%%%%%%%%%%%%%%%%%%%%%%%%
%
\section{More Ablations}\label{sec:ablation_supp} 
%
%%%%%%%%%%%%%%%%%%%%%%%%%%%%%%%%%%%%%%%%%%%%%%%%%%%%%%%%%%%%%%%%%%%%%%%%%%%%%%%%%%%%%%%%%%%%%%
%
\subsection{Random Background Augmentation}\label{sec:randomback_supp}
\begin{figure}[t]
    \centering
    \includegraphics[width=0.8\linewidth]{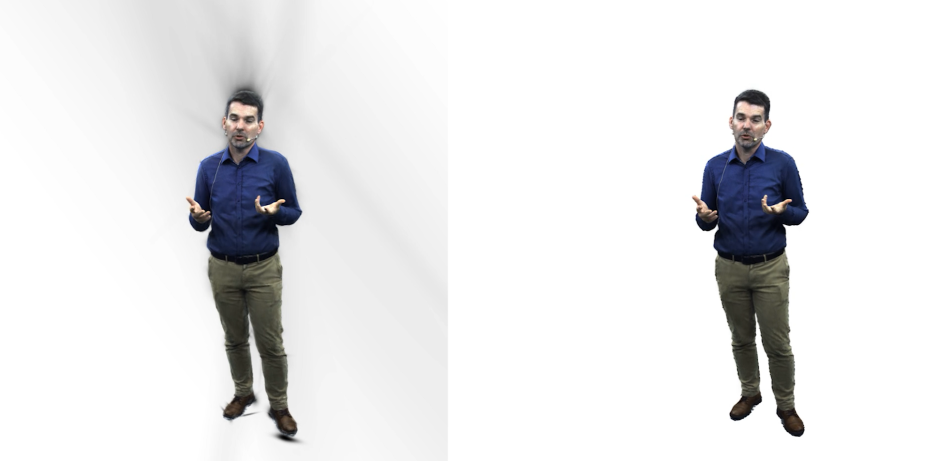}
    \\
    \vspace{-8pt}
    \subfloat[w/o rand. background]{\includegraphics[trim={0cm 0cm 0cm 40cm},clip,width=.5\linewidth]{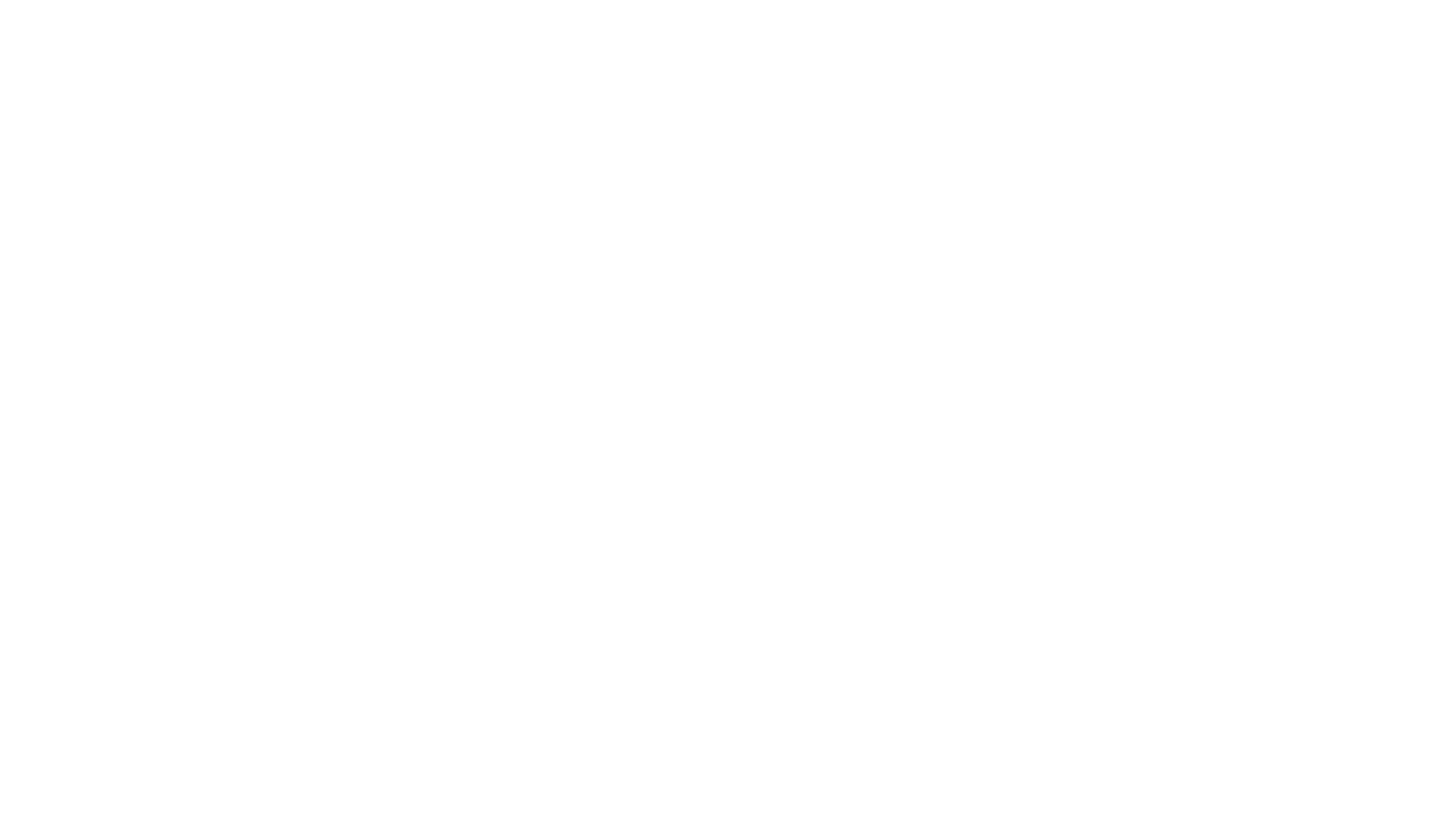}}
    \hfill
    \subfloat[\textbf{w/ rand. background}]{\includegraphics[trim={0cm 0cm 0cm 40cm},clip,width=.5\linewidth]{fig_supp/transparent.png}}
    \vspace{-8pt}
    \caption{Comparison of the appearance model trained with random backgrounds applied at each iteration versus a model trained exclusively with a fixed black background.}
    \label{fig:random_backgrounds}
\end{figure}

%
%%%%%%%%%%%%%%%%%%%%%%%%%%%%%%%%%%%%%%%%%%%%%%%%%%%%%%%%%%%%%%%%%%%%%%%%%%%%%%%%%%%%%%%%%%%%%%
%
During training of the appearance model $\boldsymbol{\Phi}_\mathrm{app}$, we noticed that solely training with a black background results in the model becoming dependent on the background. The Gaussian parameters are learned such that they require a black background. To address this issue, we adopt a random background augmentation that changes the background color for each iteration by modifying the background of the 3D Gaussian splatting rasterizer and the background color of the ground truth images. This allows the model to learn Gaussian splats that are entirely independent of the background. Additionally, this serves as a form of regularization by preventing Gaussians at the edge of the character from bleeding into the background (see Fig.~\ref{fig:random_backgrounds}).
%
%%%%%%%%%%%%%%%%%%%%%%%%%%%%%%%%%%%%%%%%%%%%%%%%%%%%%%%%%%%%%%%%%%%%%%%%%%%%%%%%%%%%%%%%%%%%%%
%
%
%%%%%%%%%%%%%%%%%%%%%%%%%%%%%%%%%%%%%%%%%%%%%%%%%%%%%%%%%%%%%%%%%%%%%%%%%%%%%%%%%%%%%%%%%%%%%%
%
\subsection{Global Position Latent}\label{sec:latent_supp}
%
%%%%%%%%%%%%%%%%%%%%%%%%%%%%%%%%%%%%%%%%%%%%%%%%%%%%%%%%%%%%%%%%%%%%%%%%%%%%%%%%%%%%%%%%%%%%%%
%
EVA's appearance predictor incorporates a global positional latent variable $\Psi$ as an additional input to its appearance networks, $\mathcal{E}_\mathrm{app}^\mathrm{body}$ and $\mathcal{E}_\mathrm{app}^\mathrm{head}$ \citep{Pang_2024_CVPR}. This latent $\Psi$ is derived by passing the character’s normalized root translation through a shallow MLP. Therefore, it is designed to capture variations in studio lighting conditions and thus serves effectively as a \emph{lighting latent}.
%
%%%%%%%%%%%%%%%%%%%%%%%%%%%%%%%%%%%%%%%%%%%%%%%%%%%%%%%%%%%%%%%%%%%%%%%%%%%%%%%%%%%%%%%%%%%%%%
%
\par
\begin{figure}[t]
    \centering
    \subfloat[Low Light]{\includegraphics[trim={25cm 17cm 25cm 10cm},clip,width=.5\linewidth]{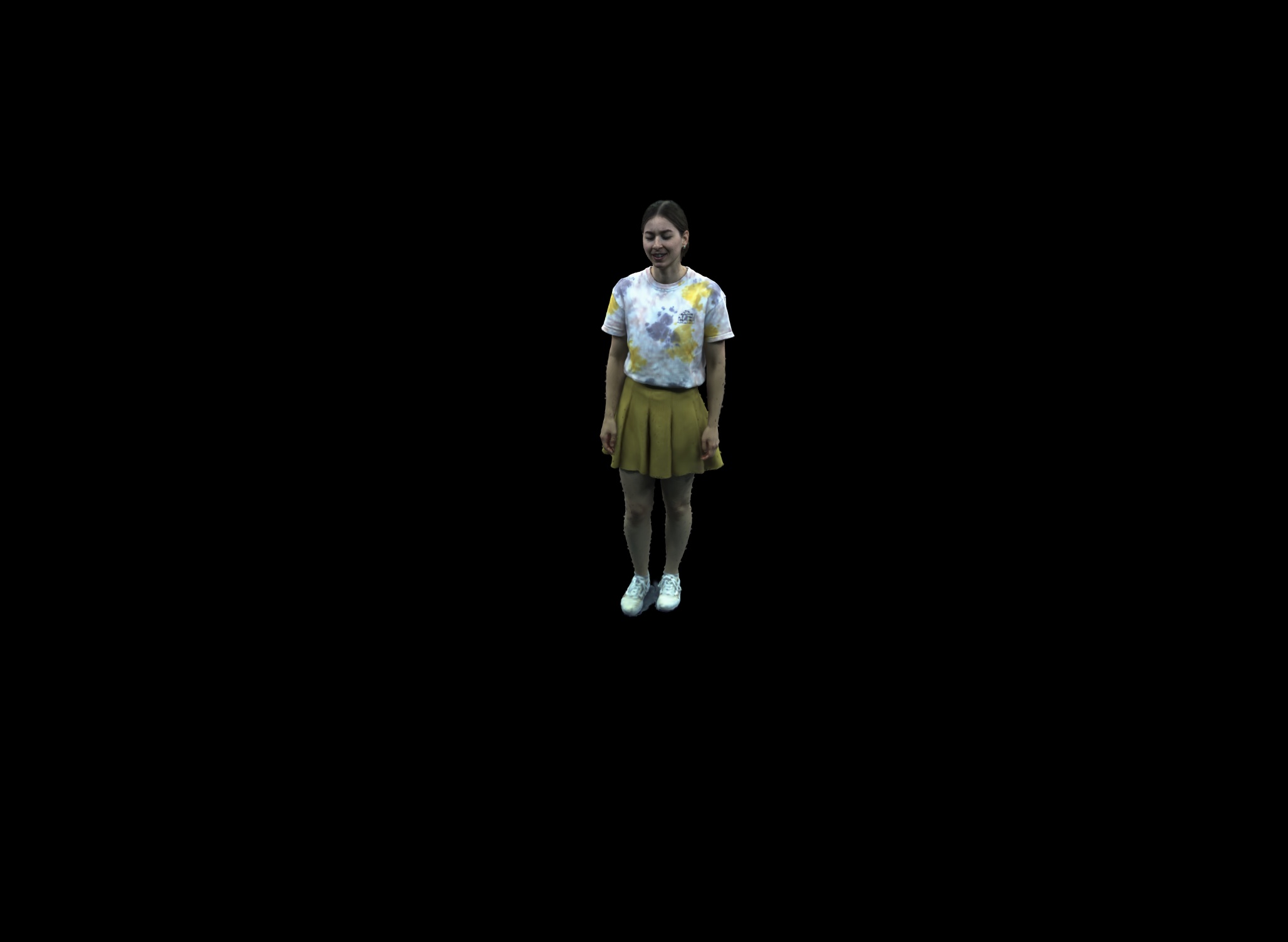}}
    \hfill
    \subfloat[Bright Light]{\includegraphics[trim={25cm 17cm 25cm 10cm},clip,width=.5\linewidth]{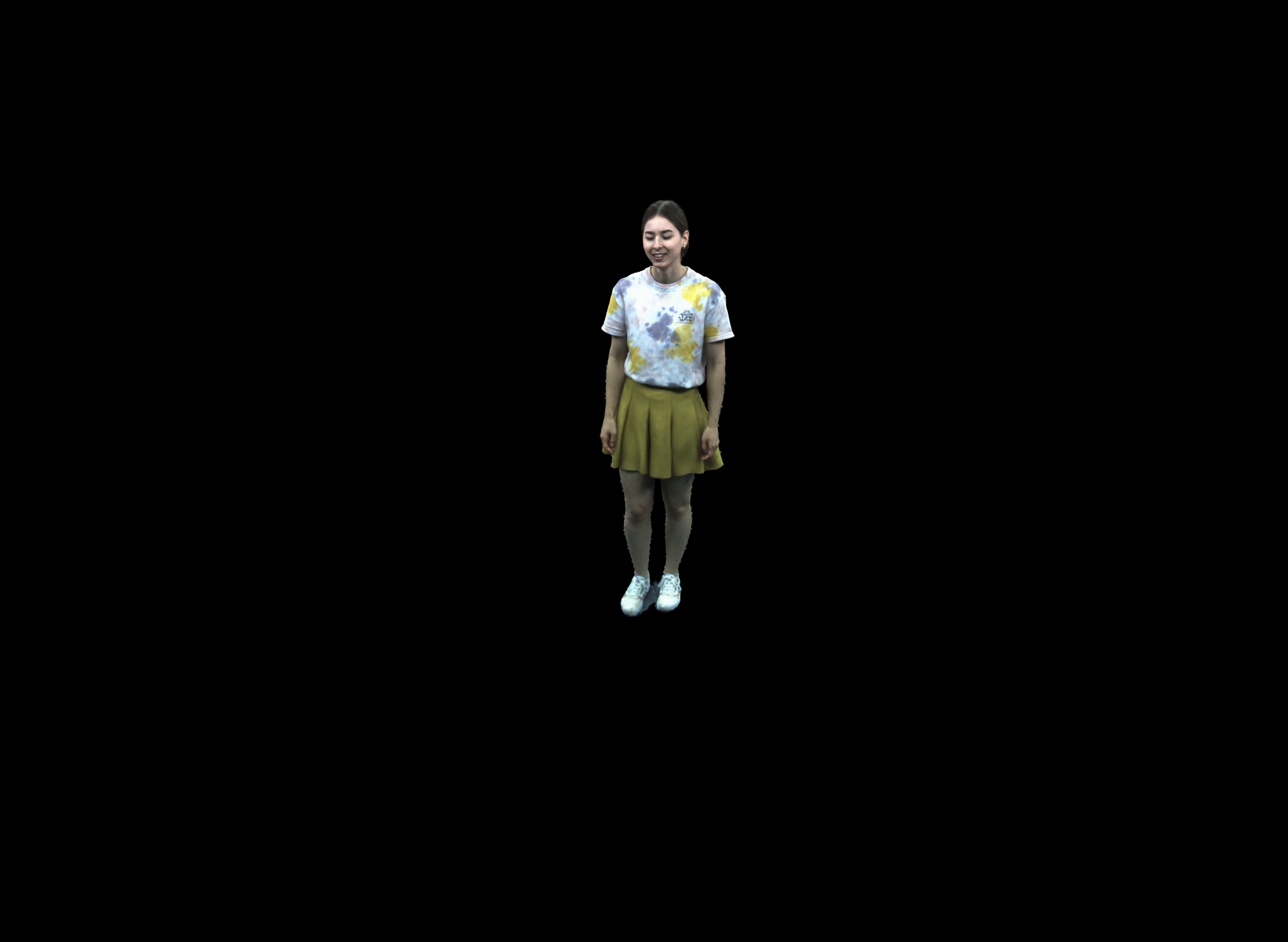}}
    \vspace{-8pt}
    \caption{Comparison of the rendered avatar under varying lighting conditions: (a) low light and (b) bright light.}
    \label{fig:latent}
\end{figure}

%
%%%%%%%%%%%%%%%%%%%%%%%%%%%%%%%%%%%%%%%%%%%%%%%%%%%%%%%%%%%%%%%%%%%%%%%%%%%%%%%%%%%%%%%%%%%%%%
%
The appearance predictor is defined as:
%
%%%%%%%%%%%%%%%%%%%%%%%%%%%%%%%%%%%%%%%%%%%%%%%%%%%%%%%%%%%%%%%%%%%%%%%%%%%%%%%%%%%%%%%%%%%%%%
%
\begin{equation}
    \phi_\mathrm{app}^{(\cdot)}(\mathcal{T}_n^{(\cdot),\triangleright}, \mathcal{T}_p^{(\cdot),\triangleright}, \Psi) = (\boldsymbol{\eta}_\mathrm{uv})_i
\end{equation}
%
%%%%%%%%%%%%%%%%%%%%%%%%%%%%%%%%%%%%%%%%%%%%%%%%%%%%%%%%%%%%%%%%%%%%%%%%%%%%%%%%%%%%%%%%%%%%%%
%
This formulation enables lighting control during inference by modifying the lighting latent, thereby granting the model a degree of relightability. Fig.~\ref{fig:latent} illustrates this relighting capability.
%
%%%%%%%%%%%%%%%%%%%%%%%%%%%%%%%%%%%%%%%%%%%%%%%%%%%%%%%%%%%%%%%%%%%%%%%%%%%%%%%%%%%%%%%%%%%%%%
%
%
%%%%%%%%%%%%%%%%%%%%%%%%%%%%%%%%%%%%%%%%%%%%%%%%%%%%%%%%%%%%%%%%%%%%%%%%%%%%%%%%%%%%%%%%%%%%%%
%
\subsection{Constant Head Gaussians}\label{sec:constgauss_supp}
%
%%%%%%%%%%%%%%%%%%%%%%%%%%%%%%%%%%%%%%%%%%%%%%%%%%%%%%%%%%%%%%%%%%%%%%%%%%%%%%%%%%%%%%%%%%%%%%
%
Importantly, EVA’s Gaussian parameters are predicted based on body pose and facial expression, enabling more expressive and adaptable appearances. This stands in contrast to prior approaches that solely optimize static 3D Gaussians attached to the FLAME mesh \citep{qian2023gaussianavatars}. This motivated us to implemented a version of EVA only optimizing constant Gaussians for the head as an ablation study. This effectively serves as our own re-implementation of GaussianAvatars \citep{qian2023gaussianavatars} since the original approach cannot faithfully track our full-body motion sequences. 
%
%%%%%%%%%%%%%%%%%%%%%%%%%%%%%%%%%%%%%%%%%%%%%%%%%%%%%%%%%%%%%%%%%%%%%%%%%%%%%%%%%%%%%%%%%%%%%%
%
\par
\begin{figure}[t]
    \centering
    \subfloat[GT]{\includegraphics[trim={0cm 0cm 0cm 0cm},clip,width=.33\linewidth]{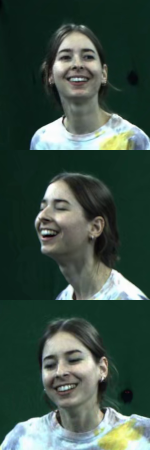}}
    \hspace{-1mm}
    \subfloat[\textbf{EVA}]{\includegraphics[trim={0cm 0cm 0cm 0cm},clip,width=.33\linewidth]{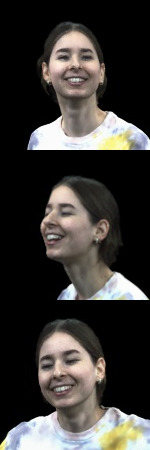}}
    \hspace{-1mm}
    \subfloat[GA]{\includegraphics[trim={0cm 0cm 0cm 0cm},clip,width=.33\linewidth]{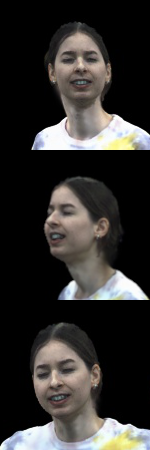}}
    \vspace{-8pt}
    \caption{Comparison between (b) our method, EVA, and (c) a reimplementation of GaussianAvatars (GA) \citep{qian2023gaussianavatars} under novel view, pose, and expression conditions. Notably, because GA optimizes only static 3D Gaussians anchored to a FLAME mesh, it fails to capture and reconstruct detailed facial expressions accurately.}
    \label{fig:ga_comparison}
\end{figure}

%
%%%%%%%%%%%%%%%%%%%%%%%%%%%%%%%%%%%%%%%%%%%%%%%%%%%%%%%%%%%%%%%%%%%%%%%%%%%%%%%%%%%%%%%%%%%%%%
%
Fig.~\ref{fig:ga_comparison} illustrates the limitations of GaussianAvatars (GA), which struggles to accurately reconstruct detailed facial expressions when relying solely on the optimization of static Gaussian parameters. This shortcoming underscores a key advantage of our method over previous approaches
This limitation is also reflected quantitatively (Tab.~\ref{tab:comparisons_AGGA}): in the novel view, pose, and expression synthesis setting, GaussianAvatars achieves a lower PSNR of $45.01$ compared to EVA’s $46.00$.
%
%%%%%%%%%%%%%%%%%%%%%%%%%%%%%%%%%%%%%%%%%%%%%%%%%%%%%%%%%%%%%%%%%%%%%%%%%%%%%%%%%%%%%%%%%%%%%%
%
%
%%%%%%%%%%%%%%%%%%%%%%%%%%%%%%%%%%%%%%%%%%%%%%%%%%%%%%%%%%%%%%%%%%%%%%%%%%%%%%%%%%%%%%%%%%%%%%
%
\section{Ethical Considerations} \label{sec:ethical}
%
%%%%%%%%%%%%%%%%%%%%%%%%%%%%%%%%%%%%%%%%%%%%%%%%%%%%%%%%%%%%%%%%%%%%%%%%%%%%%%%%%%%%%%%%%%%%%%
Our method enables new and exciting applications, which have the potential to positively impact society. 
We believe our avatars can lead to more immersive communication experiences across the globe, and we hope that such technology can bring people closer together even though they are physically separated. 
However, since our expressive avatars can produce photorealistic renderings with body motion and facial expression control, it also raises ethical concerns about identity information leakage and fake video generation.
To mitigate these risks, it is essential to explore and impose encryption design, traceability, transparency, and strict usage policies. 
We advocate for responsible deployment and strongly hope that our work is
applied in ways that positively impact society.
%

%%%%%%%%%%%%%%%%%%%%%%%%%%%%%%%%%%%%%%%%%%%%%%%%%%%%%%%%%%%%%%%%%%%%%%%%%%%%%%%%%%%%%%%%%%%%%%

\end{document}